\documentclass{article}

\usepackage{biorxiv} 

\usepackage[utf8]{inputenc} 
\usepackage[T1]{fontenc}    
\usepackage{hyperref}       
\usepackage{url}            
\usepackage{booktabs}       
\usepackage{amsfonts}       
\usepackage{nicefrac}       
\usepackage{microtype}      
\usepackage{lipsum}		
\usepackage{lineno}
\usepackage{algorithm}
\usepackage{arevmath}
\usepackage{amsmath}
\usepackage{amsthm}
\usepackage{graphicx}

\usepackage[noend]{algpseudocode}
\usepackage{caption}
\title{Visualizing Data Velocity using DSNE}

\author{
  Songting Shi
    \\
  Department of Scientific and Engineering Computing\\ 
  School of Mathematical Sciences\\
  Peking University\\
  Beijing 300071, P. R. China \\
  \texttt{songtingstone@gmail.com} \\
}

\begin{document}
\maketitle

\begin{abstract}
We present a new technique called "DSNE" which  learns the velocity embeddings of low dimensional map points when given the  high-dimensional data points with its velocities. The technique is a variation of Stochastic Neighbor Embedding,  which uses  the Euclidean distance  on the  unit sphere between the unit-length velocity of the point  and the unit-length direction from the point to its near neighbors to define similarities, and try to match the two kinds of similarities in the high dimension space and low dimension space to find the velocity embeddings on the low dimension space. DSNE can help to visualize how the data points move in the high dimension space  by presenting the movements  in two or three dimensions space. It is helpful for understanding the mechanism of cell differentiation and embryo development. 

\end{abstract}


\section{Introduction}
Visualization of high-dimensional data movement is an import problem in many different domains. Currently, in the biological science, we can compute the velocity of the mature mRNAs by RNA velocity techniques ( \cite{RNA-velocity, dynamical-RNA-velocity} ), and visualizing how the cell transit from one cell type to other cell type, which is very important for the cell differentiation  and  embryo development. 
 \cite{dynamical-RNA-velocity} promote a method to represent the velocity of the high dimensional data points on the the low dimensional map, where the velocity embeddings are modeled by  an intuitive  probability average of  the directions from the point to its $K$  nearest neighbors, which basically captures the direction of movements. We now give a more rigorous and mathematical  description of this  idea, and form a optimization problem to learn  the direction of the velocity on the low-dimensional map by  keeping  the sphere Euclidean distance invariant up to scalar, where the sphere Euclidean distance is defined between the unit-length velocity of the data point and the unit-length direction from the point to its $K$ nearest neighbors, this is finished by mimicking the Stochastic Neighbor Embedding(\cite{SNE}).

\section{Directional Stochastic Neighbor Embedding}
Similar the Stochastic Neighbor Embedding(SNE), the Directional Stochastic Neighbor Embedding of the velocity starts by converting the high-dimensional Euclidean distance between the velocity with unit length and  the unit-length direction from the point to its the near neighbors into conditional probabilities that represent similarities. The similarity of the  point $i$ with velocity $v_i$ and  the direction  from datapoint $x_i$ to datapoint $x_j$  is the conditional probability, $p_{j|i}$, that $v_i$ would  coincide with the direction from  $x_i$ to datapoint $x_j$ in proportion to their probability density under a Gaussion centered at $0$  with the distance $ || \frac {x_j - x_i} {||x_i - x_i||} - \frac{v_i} {||v_i||}||^2$ on the unit sphere. For nearby directions, $p_{j|i}$ is very high, whereas for opposite direction, $p_{j|i}$ will be almost infinitesimal (for reasonable values of the variance of the Gaussion, $\sigma_{x ,i}$). Mathematically, the conditional probability $p_{j|i}$ is given by 
\begin{equation}
\label{eq:conditonal-probability-p}
p_{j|i} = \frac{1} {Z_{x, i}} \exp( - \beta_{x, i} || \frac {x_j - x_i} {||x_i - x_i||} - \frac{v_i} {||v_i||}||^2 ) 
\end{equation}
where $\beta_{x, i} = \frac 1 {2 \sigma_{x ,i}^2}$ is the inverse of the Gaussion variance  and $Z_{x, i}= 1+  \sum_{j \in neighbors \; of \;  i} \exp( - \beta_{x, i} || \frac {x_j - x_i} {||x_i - x_i||} - \frac{v_i} {||v_i||}||^2 ) $ is the normalization factor. The $1$ in $Z_{x,i}$ accounts for the pseudo-point $x_i + tv_i$ which is generated by moving point $x_i$ along the velocity direction $v_i$ with time $t$. We include  the $p_{i|i}$ by set the $p_{i|i} = \frac{1} {Z_{x, i} } \exp( - \beta_{x, i} || \frac {(x_i +tv_i)- x_i} {||(x_i +tv_i) - x_i||} - \frac{v_i} {||v_i||}||^2 ) =  \frac{1} {Z_{x, i} }$. 

Define the cosine distance $\tilde \cos_{x, ij} := \langle \frac {x_j - x_i} {||x_i - x_i||},  \frac{v_i} {||v_i||} \rangle$ where $\langle x, y \rangle := x^Ty$ is the inner product of vector $x$ and $y$, we can simplify the conditional probability into 
\begin{equation}
\label{eq:conditonal-probability-p-cos}
p_{j|i} = \frac{1} { Z_{x, i} } \exp( - 2 \beta_{x, i} ( 1- \tilde \cos_{x, ij})  ) 
\end{equation}
where $Z_{x, i} =  1+ \sum_{j \;  \in\;  neighbors \; of \;  i}   \exp( - 2 \beta_{x, i} ( 1- \tilde \cos_{x, ij})  ) $. 
 Note that the popular dimensional-reduction techniques, e.g., t-SNE (\cite{t-SNE}), UMAP (\cite{UMAP}), they mainly focus on the preservation of  the local organization structure, which implies that the velocity direction  are only preserved on the local structure, so we choose the neighbors of $i$ by finding its K near neighbors under the Euclidean measure $||x_j - x_i||$ and also including  the pseudo-point $x_i + tv_i$ as stated before. 
For the low-dimensional conterparts $y_j$ and $y_i$ with the low-dimensional velocity $w_i$, it is possible to compute a similar conditional probability, which we denote by $q_{j|i}$. We model the similarity of  velocity embedding 
$w_i$ of $v_i$  with the direction from map point $y_i$ to map point $y_j$ by 
\begin{equation}
\label{eq:conditional-probability-q}
q_{j|i} = \frac{1} {Z_{y, i} } \exp( - \beta_{y, i} || \frac {y_j - y_i} {||y_i - y_i||} - \frac{w_i} {||w_i||}||^2 ) 
\end{equation}
where $\beta_{y, i} :=  \frac 1 {2 \sigma_{y ,i}^2}$ is the inverse of the Gaussion variance  and $Z_{y, i}= 1+   \sum_{j \in neighbors \; of \;  i} \exp( - \beta_{y, i} || \frac {y_j - y_i} {||y_j - y_i||} - \frac{w_i} {||w_i||}||^2 ) $ is the normalization factor. 
The $1$ in $Z_{y,i}$ accounts for the pseudo-point $y_i + tw_i$ which is generated by moving point $y_i$ along the velocity direction $w_i$ with time $t$. We include  the $q_{i|i}$ by set the $q_{i|i} = \frac{1} {Z_{y, i} } \exp( - \beta_{y, i} || \frac {(y_i +tw_i)- y_i} {||(y_i +tw_i) - y_i||} - \frac{w_i} {||w_i||}||^2 ) =  \frac{1} {Z_{y, i} }$. 
Define the cosine distance $\tilde \cos_{y, ij} := \langle \frac {y_j - y_i} {||y_j - y_i||},  \frac{w_i} {||w_i||} \rangle$, we can simplify the conditional probability into 
\begin{equation}
\label{eq:conditonal-probability-q-cos}
q_{j|i} = \frac{1} {Z_{y, i}} \exp( - 2 \beta_{y,  i} ( 1- \tilde  \cos_{y, ij})  ) 
\end{equation}
where $Z_{y, i} =  1 + \sum_{j \;  \in\;  neighbors \; of \;  i}   \exp( - 2 \beta_{y, i} ( 1- \tilde \cos_{y, ij})  ) $. 

For notation simplicity, In the flowing description, we denote $ \sum_{j \in neighbors \; of \;  i}$  as $\sum_{ j\ne i}$ and denote 
$ \sum_{j \in \{ \text{neighbors \; of \;  i} \} \cup \{i\} }$  as $\sum_{ j}$;  $\hat x_{i,j}:=  \frac {x_j - x_i} {||x_j - x_i||}$, $\hat y_{ij} := \frac {y_j - y_i} {||y_j - y_i||}$, $\hat v_i =  \frac{v_i} {||v_i||}$ and $\hat w_i =  \frac{w_i} {||w_i||}$. 

If the velocity map points $w_i$ correctly model the direction of the high-dimensional velocity $v_i$ in a local space, then the conditional probability $p_{j|i}$ and $q_{j|i}$ will be equal. Motivated by this observation, we aims to find a low-dimensional velocity representation that minimizes the mismatch between $p_{j|i}$ and $q_{j|i}$. A natural measure of the faithfulness with which $q_{j|i}$ model $p_{j|i}$ is the Kullback-Leibler divergence ( which is in this case equal to the cross-entropy up to an additive constant). We minimizes the sum of Kullback-Leibler divergences and the cost function $C$ is given by 
\begin{equation}
\label{eq:loss}
C = \sum_i KL(P_i || Q_i ) = \sum_i \sum_j p_{j|i} \log \frac { p_{j|i} } {q_{j|i}}
\end{equation}
in which $P_i$ represent the conditional probability distribution over the  directions  from the data point $x_i$ to its neighbor points  and  the pseudo-point $x_i + tv_i$  given the velocity $v_i$ of data point $x_i$, and $Q_i$ represent the conditional probability distribution over the directions from the map point $y_i$ to its   neighbor map points and the pseudo-point $y_i + tw_i$ given the map velocity $w_i$ of point $y_i$,  where $y_i$  using the same  neighbors as in $P_i$. 
%

It seems very reasonable of the above formulation, but when we do experiment on the simulation data, it does the wrong work on the simulation data with exact data points and velocities ( see Section~\ref{sec:approximate-simulation}) but work perfectly on the simulation data with exact map points and velocity embeddings ( see Section~\ref{sec:exact-simulation})  )  where the data points and its velocities  coming  from  the  linear projecting of  the exact map points and its velocity embeddings. So why? We now give a simple analysis of the above formulation. Since the loss function is the KL divergence, in the ideal case, we will get that $p_{j|i} = q_{j|i}$,  in which case the cost $C=0$. Comparing the $p_{j|i}$  with $ q_{j|i}$, we will get the following relations,
\begin{equation}
\label{eq:ideal-case-p-q}
\begin{array}{lcl}
p_{i|i} = q_{i,i}  &\Longrightarrow & Z_{x,i} = Z_{y,i} \\ 
p_{j|i} = q_{j|i}, \; j \ne i   & \Longrightarrow&     \beta_{x,i}(1 - \tilde \cos_{x,ij}) = \beta_{y,i}(1 - \tilde \cos_{y,ij} ) \; \text{for  }  j \ne i  \\
					&\Longleftrightarrow& \beta_{x,i} || \hat v_i - \hat x_{ij}||^2 =  \beta_{y,i} || \hat w_i - \hat y_{ij}||^2 \; \text{for  }  j \ne i  
\end{array}
\end{equation}
The above relations imply that when we minimize the KL divergence, we will find the final solution $\hat w_i$  that satisfy the above linear relations between sphere distances of high dimension space and the low dimension space.
Note that in the high dimension space, the $\{ \hat x_{ij}, \; j \in \text{ i's near neighbors }\}$ usually close to each other. This will cause the problem, since  $\{ || \hat v_i - \hat x_{ij}||^2 , \; j \in \text{ i's near neighbors }\}$ will also close to each other, so we can not faithfully determine $\hat w_i$ from these minor differences $\{ || \hat v_i - \hat x_{ij}||^2 , \; j \in \text{ i's near neighbors } \}$. To make the the  directions to the near neighbors more uniformly distributed on sphere, we now use the view from the end point  of the  mean direction,  which is defined by $\bar x_i := \frac 1 {\text {the number of i's neighbors} } \sum_{j \in \text{i's neighbors}} \hat x_{ij}$, which we will get the following directions,
\begin{equation}
\label{eq:direction-view-from-mean}
\begin{array}{l}
\Delta \hat x_{ij} =\frac{ \hat x_{ij} - \bar x_i } {|| \hat x_{ij} - \bar x_i ||}
\end{array}
\end{equation}.
To get the intuition, let we think a simple example. Suppose that $x_i := (0,0)$ and its $3$ near neighbors are $x_1=(-1,1)$, $x_2 = (0,1)$, $x_3=(1,1)$, then we have the three  directions from point $x_i$ to its there near neighbors,  $\hat x_{i,1} = \frac 1 {\sqrt 2} ( -1,1) \approx (-0.707, 0.707)$, $\hat x_{i,1} = ( 0,1)$,   $\hat x_{i,2} = \frac 1 {\sqrt 2} ( 1,1) \approx (0.707, 0.707)$. Then mean direction will be $\bar x_{i} = \frac 1 3 (\hat x_{i,1} + \hat x_{i,2} + \hat x_{i,3} ) = (0, \frac { 1 + \sqrt 2} 3) \approx (0,0.805) $. Form the view on end point of mean direction, we will have the directions, $\Delta  \hat x_{i,1} = \frac{ \hat x_{i,1}  - \bar x_i } { ||\hat x_{i,1}  - \bar x_i||} \approx (-0.991, 0.137) $,  $\Delta  \hat x_{i,2}= (0, 1) $, $\Delta  \hat x_{i,3} \approx (0.991, -0.137) $. The directions corrected by the mean direction   are more uniformly distributed on the sphere than the original directions to its neighbors. These well-separated directions on the sphere will  help to locate any velocity direction on the sphere more easily.

Now we get the following representation of the current problem. 
\begin{equation}
\label{eq:formulation}
\begin{array}{l}
p_{j|i} = \frac 1 {Z_{x,i} } \exp( - 2 \beta_{x,i}( 1 - \cos_{x,ij})), \; j \in \text{  i's neighbors } \\
p_{i|i}  =  \frac 1 {Z_{x,i} } \\ 
Z_{x,i}  = 1 +  \sum_{j \in \text{  i's neighbors } } \exp( - 2 \beta_{x,i}( 1 - \cos_{x,ij})) \\
\cos_{x,ij} = \langle \hat v_i, \Delta \hat x_{ij} \rangle\\
\Delta \hat x_{ij} =\frac{ \hat x_{ij} - \bar x_i } {|| \hat x_{ij} - \bar x_i ||} \\
 \bar x_i =  \frac 1 {\text {the number of i's neighbors} } \sum_{j \in \text{i's neighbors}} \hat x_{ij} \\
q_{j|i} = \frac 1 {Z_{y, i} } \exp( - 2 \beta_{y,i}( 1 - \cos_{y,ij})), \; j \in \text{  i's neighbors } \\
q_{i|i}  =  \frac 1 {Z_{y,i} }  \\
Z_{y,i}  = 1 +  \sum_{j \in \text{  i's neighbors } } \exp( - 2 \beta_{y,i}( 1 - \cos_{y,ij})) \\
\cos_{y,ij} = \langle \hat w_i, \Delta \hat y_{ij} \rangle\\ 
\Delta \hat y_{ij} =\frac{ \hat y_{ij} - \bar y_i } {|| \hat y_{ij} - \bar y_i ||} \\
 \bar y_i =  \frac 1 {\text {the number of i's neighbors} } \sum_{j \in \text{i's neighbors}} \hat y_{ij} \\

C =  \sum_i \sum_j p_{j|i} \log \frac { p_{j|i} } {q_{j|i}}\\
\end{array}
\end{equation}

There is one problem in the above formulation, note that in the ideal case, we will have that $p_{i|i} = q_{i|i}$, so that one part of loss $p_{i|i} \log \frac {p_{i|i} } { q_{i|i} } =0$ do not contribute to the loss. While the $p_{i|i}$ will take a large part of probability mass ( $ p_{i|i}= \frac 1 {Z_{x,i}} \ge p_{j|i} = \frac{ \exp( - 2 \beta_{x,i}( 1 - \cos_{x,ij})) } {Z_{x,i}}  , \; j \ne i $), which will hinder  the optimization of the loss function. To alleviate  this problem, we use the following probability distribution  without considering the pseudo-point $x_i + tv_i$, 
\begin{equation}
\label{eq:parital-p}
\begin{array}{l}
\tilde p_{j|i} = \frac 1 {\tilde Z_{x,i}  } \exp( - 2 \beta_{x,i}( 1 - \cos_{x,ij})),  \; j \in \text{  i's neighbors } \\
\tilde Z_{x,i}  =  \sum_{j \in \text{  i's neighbors } } \exp( - 2 \beta_{x,i}( 1 - \cos_{x,ij})) \\
\end{array}
\end{equation}
to weight the error term $\log  \frac { p_{j|i} } {q_{j|i}}$. We modify the loss function to the following as the loss of DSNE. 
\begin{equation}
\label{eq:DSNE-loss}
\begin{array}{l}
C = \sum_i \sum_{j \in \text{i's neighbors}} \tilde p_{j|i}  \log \frac { p_{j|i} } {q_{j|i}}
\end{array}
\end{equation}

Finally, we get the the optimization problem of DSNE as follows,
\begin{equation}
\label{eq:formulation-final}
\begin{array}{l}
p_{j|i} = \frac 1 {Z_{x,i} } \exp( - 2 \beta_{x,i}( 1 - \cos_{x,ij})), \; j \in \text{  i's neighbors } \\
p_{i|i}  =  \frac 1 {Z_{x,i} } \\ 
Z_{x,i}  = 1 +  \sum_{j \in \text{  i's neighbors } } \exp( - 2 \beta_{x,i}( 1 - \cos_{x,ij})) \\
\cos_{x,ij} = \langle \hat v_i, \Delta \hat x_{ij} \rangle\\
\tilde p_{j|i} = \frac 1 {\tilde Z_{x,i}  } \exp( - 2 \beta_{x,i}( 1 - \cos_{x,ij})),  \; j \in \text{  i's neighbors } \\
\tilde Z_{x,i}  =  \sum_{j \in \text{  i's neighbors } } \exp( - 2 \beta_{x,i}( 1 - \cos_{x,ij})) \\
q_{j|i} = \frac 1 {Z_{y, i} } \exp( - 2 \beta_{y,i}( 1 - \cos_{y,ij})) \; j \in \text{  i's neighbors } \\
q_{i|i}  =  \frac 1 {Z_{y,i} }  \\
Z_{y,i}  = 1 +  \sum_{j \in \text{  i's neighbors } } \exp( - 2 \beta_{y,i}( 1 - \cos_{y,ij})) \\
\cos_{y,ij} = \langle \hat w_i, \Delta \hat y_{ij} \rangle\\ 
C =  \sum_i \sum_{j \ne i}  \tilde p_{j|i} \log \frac { p_{j|i} } {q_{j|i}}\\
\end{array}
\end{equation}

The remaining parameter to be selected is the inverse of  variance $\beta_{x,i}:=\frac 1 { 2 \sigma^2_{x,i}  }$ of the Gaussian.
It is not likely that there is a single value of $\beta_{x,i}$ that is optimal for all velocities in the dataset because the density of the data is likely to vary. In dense regions, a large value of $\beta_{x,i}$ ( a smaller value of $\sigma_{x,i}$)  is usually more appropriate than in sparser regions, since it will  scale the distance  $\beta_{x, i}2 ( 1- \cos_{x, ij})$ separate each other well which aids to optimization. Any particular value of $\beta_{x, i}$ includes a probability distribution, $P_i$, over the directions from point $x_i$ to its neighbor points  and the pseudo-point  $x_i + tv_i$.  This distribution has an entropy which increase as $\beta_{x,i}$ decreases ( $\sigma_{x,i}$ increases), DSNE performs a binary search for the value of $\beta_{x,i}$ that produces a $P_i$ with a fixed perplexity that is specified by the user. The perplexity is defined as 
\begin{equation}
\label{eq:perplexity}
Perp(P_i) = 2^{H(P_i)}
\end{equation}
where $H(P_i)$ is the Shannon entropy of $P_i$ measured in bits
\begin{equation}
\label{eq:entropy}
H(P_i) = - \sum_j p_{j|i} \log_2 p_{j|i}
\end{equation}
The perplexity can be interpreted as a smooth measure of the effective number of neighbors. The performance of DSNE is  relatively robust to changes in the perplexity and it prefers  the lower value  of perplexity, typical values are between $1$ to $6$  and the corresponding $K$ are between $6$ to $16$ which are based on the experiences on the simulation data.

The minimization of the cost function in Equation \ref{eq:loss} is performed using a gradient descent method  for $w_i$ and binary search for $\beta_{y,i}$. The gradient with respect to $w_i$  has a surprisingly simple form 
\begin{equation}
\label{eq:gradient-w}
\frac{\partial C} {\partial w_i} = \sum_{ j \ne i}  (\tilde p_{j|i} - q_{j|i}) \frac {2 \beta_{y,i}}   { ||w_i||}  ( - \Delta \hat y_{ij} + cos_{y,ij} \hat w_i )
\end{equation}
And  the second order partial derivatives with respect to $w_i$   is given by
\begin{equation}
\label{eq:hessian-w}
\begin{array}{ll}
G_{w_i} &= \frac { 2 \beta_{y,i} } { ||w_i||^2 } \sum_{ j \ne i}  (\tilde p_{j|i} - q_{j|i})  [ \Delta \hat y_{ij}  \hat w_i^T + \hat w_i \Delta  \hat y_{ij}^T + \cos_{y,ij} I - 3 \cos_{y,ij} \hat w_i \hat w_i^T ] \\
	& \quad - \frac{ 4\beta_{y,i}^2} {||w||^2} \sum_{ j \ne i}  q_{j|i}  [ -\Delta  \hat y_{ij} + cos_{y, ij} \hat w_i] [ ( \Delta  \hat y_{ij} - \mathbb E \Delta  \hat y_{i}) - \hat w_i ( \cos_{y, ij} - \mathbb E \cos_{y, i} ) ]^T
\end{array}
\end{equation}
where $G_{w_i}(k,l) = \frac{\partial C} {\partial w_i(k) \partial w_i(l) }$,  $\cos_{y, ij} := \langle \hat w_i, \Delta \hat y_{ij}  \rangle$,  $\Delta \hat y_{ij} := \frac{  \hat y_{ij} - \bar y_{i} } { ||  \hat y_{ij} - \bar y_{i} ||} $,  $ \hat y_{ij} :=  \frac { y_j - y_i} { ||y_j - y_i||}$, $\bar y_i = \frac 1 {\text {the number of i's neighbors} } \sum_{j \in \text{i's neighbors}} \hat y_{ij}$, 
$\hat w_i := \frac{w_i} {||w_i|| } $, $ \mathbb E \Delta \hat y_i := \sum_{k \ne i} q_{k|i}  \Delta  \hat y_{ik}$,  $ \mathbb E \cos_{y,i} := \sum_{k \ne i} q_{k|i}  \cos_{y,ik}$. 
Note that the Hessian matrix  $G_{w_i}$  has a scalar $\beta_{y,i}$ which is common with the gradient $\frac{\partial C} {\partial w_i}$, by mimicking the Newton's method,  we can use the scaled gradient
\begin{equation}
\label{eq:gradient-w}
g_{w_i} = \sum_{ j  \ne i} (\tilde p_{j|i} - q_{j|i})   ( - \Delta \hat y_{ij}+ \cos_{y,ij} \hat w_i)
\end{equation}
to update $w_i$. Also note that  the loss is independent of the norm of $w_i$, we can restrict the $w_i$ on the sphere with  $||w_i||=1$, which can be finished  by  scaling $w_i$  with  $ w_i = \frac{w_i} {||w_i|| }$ after each updating of $w_i$. 

The gradient of the loss w.r.t  $\beta_{y_i}$ is given by 
\begin{equation}
\label{eq:gradient-beta}
\frac{\partial C} {\partial \beta_{y, i} } = \sum_{ j \ne i}  (\tilde p_{j|i} - q_{j|i}) 2 ( 1 -  cos_{y,ij} )
\end{equation}
and the second order derivatives is given by
\begin{equation}
\label{eq:hessian-beta}
\frac{\partial^2 C} {\partial^2 \beta_{y, i} } = 4[\sum_{ j \ne i }  q_{j|i} ( 1 -cos_{y,ij})^2 - (\sum_{ j \ne i }   q_{j|i} ( 1 -cos_{y,ij}))^2 ]
\end{equation}
Note that $\frac{\partial^2 C} {\partial^2 \beta_{y, i} } \ge 0$ by the Cauchy inequality, so the cost is a convex function about $\beta_{y,i}$, which is easy to optimize. 
Note that we should update $\beta_{y,i}$ toward the direction such that $p_{j|i} = q_{j|i}$ also $Z_{x,i} = Z_{y,i}$. 
We can use the binary search to  adjust $\beta_{y,i}$ to make that the conditional distribution $Q_i$ has a fixed perplexity the same as $P_i$, which will be satisfied when $p_{j|i} = q_{j|i}$.   To make a dedicate control the update of $\beta_{y,i}$, we only use the binary split rule  to update $\beta_{y,i}$ when the gradient of $\beta_{y,i}$ and the $dH = H(Q_i) - \log (Perp)$ with the different signs. The reason behind this is that when  the $dH>0$,   the entropy of $Q_i$ is too large, we should reduce the entropy hence increase the value of $\beta_{y,i}$ ( reduce the variance $\sigma_{y,i}^2$). This is only reasonable when we have a negative gradient, in which increasing  the value of $\beta_{i}$ will reduce the current cost.  The opposite site has a similar reason.

To accelerate the convergence speed, we use an adaptive momentum gradient update scheme (\cite{Jacobs1988}) for $w_i$ and update the value of $\beta_{y,i}$ using  conditioned binary search method described above.

Note that this algorithm will produce the direction of the low-dimensional velocity, but ideally we want to get the velocity  embedding with the norm on the low dimension space. Note that there are   approximately  relation $||x_i|| / ||v_i|| \approx ||y_i||/||w_i||$  which tell us $||w_i|| \approx ||y_i||/ ||x_i|| ||v_i|| $. We use the following approximation to get the norm of $||w_i||$.


\begin{equation}
\label{eq:norm-w}
||w_i|| = [\frac 1 N  \sum_{j =1}^N \frac{ ||y_{j}|| + d  } {  ||x_{j}|| +D }] ||v_i||
\end{equation}
where we add $d , D$ to $||y_{j}||, \; ||x_{j}||$ respectvely for numerical stability. 

And the final velocity embedding  is given by 


\begin{equation}
\label{eq:w-with-norm}
w_i= [\frac 1 N  \sum_{j =1}^N \frac{ ||y_{j}||+d } {  ||x_{j}|| +D}] ||v_i|| \hat w_i
\end{equation}

Now, we give the DSNE algorithm~\ref{alg:dsne} to guide the details of imagination.

\begin{algorithm}
\caption{DSNE: Direction Stochastic Neighbor Embedding}
\label{alg:dsne}
\begin{algorithmic}[1]

\Function{DSNE}{$X$, $V$, $Y$, perlexity, $N$,  $K$, $D$, $d$} 
   \State Data format: data points matrix $X \in \mathbb R^{N \times D}$, velocities matrix $V \in \mathbb R^{N \times D}$, 
   	     low-dimensional map points matrix $Y  \in \mathbb R^{N \times d}$. 
    \State Initializing velocity embedding matrix $W  \in \mathbb R^{N \times d}$ with random uniform variable and normalized $W$ by row to the surface of standard ball, i.e 
    $w_i := \frac { w_i } { ||w_i|| }, \; i= 1, \ldots N$. 
    \State Initializing $\text{gains}_W \in \mathbb R^{N \times d}$ with  values $1$;
    \State Initializing the moment accumulate gradient $u_W \in \mathbb R^{N \times d}$ with values $0$. 
    \State Initializing the $\beta_{y,i} = 1, \; i = 1,\ldots, N$.  
    \State Search the K nearest neighbors for each $x_i$ with Euclidean distance $d_{ij} := || x_i - x_j||^2$ which finished by the vantage point tree algorithm(\cite{vptree}). And store the $K$ nearest neighbor index of each data point $i$ into matrix $B \in \mathbb R^{N \times K}$ where $B(i,k)$ is the index of the k-th nearest neighbor of $i$. 
    \State Using the nearest neighbor index $B$ to compute the $P_{j|i},  i=1, \ldots, N, j \in \{ i \} \cup \{ B(i,k), k=1, \ldots, K\}  $ where using the binary search method to compute the inverse of the variance $\beta_{x,i}$ such that  the entropy $H(P_i)$ of $P_i$ equals the $\log(Perplexity)$. Get the value $\tilde p_{j|i} = \frac{ p_{j|i} } { \sum_{j   \in \{ B(i,k), k=1, \ldots, K\} }  p_{j|i} }$.   Storing the conditional probability into the matrix $\tilde P \in \mathbb R^{N \times K}$ where $\tilde P[i,k] = \tilde p_{B(i,k)|i},  \; i=1, \ldots, N, \; k= 1,\ldots, K$.
    \State Compute the unit-length neighbor direction $\hat y_{ij} := \frac { y_j - y_i} { ||y_j - y_i||}, \; i=1, \ldots, N, j=B(i,k), k= 1,\ldots, K$; and then compute the mean directions $\bar y_{i} = \frac 1 K \sum_{j \in  \{ B(i,k), k=1, \ldots, K\} } \hat y_{ij}$; compute the mean direction corrected direction $\Delta \hat y_{ij}= \frac{  \hat y_{ij} - \bar y_i } {|| \hat y_{ij} - \bar y_i|| },   \; i=1, \ldots, N, j=B(i,k), k= 1,\ldots, K$, and then  store them into the array $\Delta \hat Y \in \mathbb R^{N \times K \times d}$, where $\Delta \hat Y[i,k] =\Delta \hat y_{i,B(i,k)} \; i=1, \ldots, N, \; k= 1,\ldots, K$.  
        \Repeat
            	\State $W \leftarrow $ {UpateVelocityEmbedding}( $\tilde P$,  $B$, $\Delta \hat Y$,  $\beta_y$, $W$, $\text{gains}_W$, $u_W$, $N$,  $K$, $d$ )
        		\State $ \beta_y \leftarrow $ {UpateBetaQ} ( $\Delta \hat Y$, $B$, $\beta_y$, $W$, $perplexity$, $N$, $K$, $d$)	
         \Until{convergence}
         \State Compute the the $W$ with the norm $w_i =  [\frac 1 N  \sum_{j =1}^N \frac{ ||y_{j}||+d } {  ||x_{j}|| +D}] ||v_i||  w_i, \; i = 1, \ldots, N$
   
   \State \Return{$W$}
\EndFunction

\end{algorithmic}
\end{algorithm}

We update the velocity embeddings $W$ by the gradient descent  method with momentum, which is given in the following algorithm~\ref{alg:update-W}. Note that we use the adaptive learning rate scheme described by Jacobs \cite{Jacobs1988}, which gradually increases the learning rate in the direction in which the gradient is stable.

\begin{algorithm}
\caption{Updating the Velocity Embedding}
\label{alg:update-W}
\begin{algorithmic}[1]

\Function{UpateVelocityEmbedding}{ $\tilde P$, $B$, $\Delta \hat Y$,  $\beta_y$, $W$, $\text{gains}_W$, $u_W$, $N$,  $K$, $d$ }
    \State Initializing the learning rate $\eta$.
    \State Initializing the momentum scalar $\gamma$.
    \Repeat
            \State Compute the scaled gradient $g_W \in \mathbb R^{N \times d}$ of $W$ with $g_{w_i }= \sum_{j \in \{ B[i,k], k=1, \ldots, K\} }  (\tilde p_{j|i} - q_{j|i})   ( - \Delta \hat y_{ij}+ cos_{y,ij} \frac{w_i} {||w_i|| } )$.
            \State Update the gains of gradient with $\text{gains}_{w_i}  = (sign(g_{w_i}) != sign(u_{w_i}) ? (\text{gains}_{w_i} + 0.2) : (\text{gains}_{w_i} * 0.8), \; i=1, \ldots, N$.
            \State Update the momentum accumulated gradient $u_{w_i} =  \gamma * u_{w_i} - \eta * \text{gains}_{w_i} *  g_{w_i}, \; i= 1, \ldots, N$. 
            \State Update $W$ with $w_i = w_i + u_{w_i}, \; i=1, \ldots, N$.
             \State Normalize $W$ with unit length, i.e. $w_i = \frac{w_i} {||w_i||}, \; i = 1, \ldots, N$. 
   \Until{convergence}
   \State \Return{$W$}
\EndFunction

\end{algorithmic}
\end{algorithm}

We update the inverse of Variance $\beta_y$ with the conditional binary search with the Algorithm~\ref{alg:update-beta}
\begin{algorithm}
\caption{Updating the Inverse of Variance}
\label{alg:update-beta}
\begin{algorithmic}[1]

\Function{UpateBetaQ}{$\Delta \hat Y$,  $B$,  $\beta_y$, $W$, $perplexity$, $N$, $K$, $d$}
    \State Initializing the threshold $tol = 1e-5$.  
    	\For {$i \leftarrow 1\ldots N$}		
		\State Initialize $\beta = \beta_{y,i}$.
		\State Initialize $\beta_{max} = DBLMAX$, i.e. the maximum of the double type. 
		\State Initialize $\beta_{min} = -DBLMAX$,  i.e. the minimum of the double type. 
		\Repeat
			\State Compute $cos_{y,ij}$ and  $q_{j|i}$ with $\beta$.
			\State Compute the scaled gradient $g_{\beta}$ of $\beta$ with $g_{\beta}= \sum_{j \in \{ B[i,k], k=1, \ldots, K\} }   (\tilde p_{j|i} - q_{j|i})   2( 1-  \cos_{y,ij})$. 
			\State Compute the entropy $H = -\sum_{ j \in \{ B[i,k], k=1, \ldots, K\} \cup \{ i\} }  q_{j|i} \log q_{j|i}$.
			\State Compute the entropy difference $dH = H - \log(perplexity)$.
			\If{($|g_{\beta_i}| < tol$) || ($|dH| < tol$) || ($dH*g_{\beta_i} \ge 0$)}
				\State $\beta_{y,i} = \beta$. 
				\State  Break the Repeat loop
			\Else
				\If{ $dH>0$}
					\State $\beta_{min} = \beta$. 
					\If{$\beta_{max} = DBLMAX$ || $\beta_{max} = -DBLMAX$}
						\State $\beta = 2\beta$.
					\Else
						\State $\beta = \frac{ \beta + \beta_{max}} 2$.
					\EndIf 
				\Else
					\State $\beta_{max} = \beta$. 
					\If{($\beta_{min} = -DBLMAX$) || ($\beta_{min} = DBLMAX$)}
						\State $\beta = \beta/2$.
					\Else
						\State $\beta = \frac{ \beta + \beta_{min}} 2$.
					\EndIf 
					
				\EndIf
			\EndIf
			
		\Until{convergence}
		\State $\beta_{y,i} = \beta$.
	\EndFor	
   \State \Return{$\beta_{y}$}
\EndFunction

\end{algorithmic}
\end{algorithm}

 \textit{ Implementation details.} 
 We only find the velocity embedding for $v_i$ with $||v_i|| >0$. We use the vantage point tree C code implemented in BH-SNE (\cite{BH-SNE})  package ( \url{https://github.com/danielfrg/tsne} ). 
%

\section{Comparement with scVelo Velocity Embedding}
\cite{dynamical-RNA-velocity}  proposed the following velocity embedding. 

\begin{equation}
\label{eq:scvelo-velocity-embedding}
\begin{array}{ll}
w_i = \sum_{ j \in \text{i's near neighbors}}  \tilde p_{j|i} \hat y_{ij} -  \bar  y_{full, i } 
\end{array}
\end{equation}
where $\hat y_{ij} =  \frac { y_j - y_i} {|| y_j - y_i|| }$, $\bar y_{full, i }= \frac 1 N \sum_{j=1}^N  \hat y_{ij}$, 
$\tilde p_{j|i} =  \frac{1} {\tilde Z_i} \exp( - 2 \beta_{x, i} ( 1- \tilde \cos_{x, ij})  ) $,  $\tilde Z_i = \sum_{ j \in \text{i's near neighbors}}  \exp( - 2 \beta_{x, i} ( 1- \tilde \cos_{x, ij})  )$ and $\tilde \cos_{x, ij} = \langle v_i, \hat x_{ij} \rangle$ , $\hat x_{ij} :=  \frac { x_j - x_i} {|| x_j - x_i|| }$, where $i$'s near neighbors were chosen from the  K nearest neighbors of $x_i$ under the Euclidean distance essentially, excluding the point $i$ itself, i.e. $\tilde P_{i|i} =0$. We termed this algorithm by the name scVeloEmbedding. 

It works relative well in the experiments, although not as good as DSNE. We first make a connection between the two kinds of algorithms, And then we give some explanations why the scVeloEmbedding works well and why DSNE is a more accurate method than scVeloEmbedding. 

Note that $w_i$  in by the equation (\ref{eq:scvelo-velocity-embedding})  can be viewed as 
\begin{equation}
\label{eq:scvelo-velocity-embedding-gradient-view}
\begin{array}{ll}
 \sum_{ j \in \text{i's near neighbors}}  \tilde p_{j|i}( w_i -    (   \hat y_{ij}  -  \bar  y_{full, i } ) )  =0
\end{array}
\end{equation}
, which is the gradient of the following loss function
\begin{equation}
\label{eq:scvelo-velocity-embedding-loss}
\begin{array}{ll}
 C_{scVelo,i } := \sum_{ j \in \text{i's near neighbors}}  \tilde p_{j|i}||  w_i -    (   \hat y_{ij}  -  \bar y_{full, i }) ||^2
\end{array}
\end{equation}
This loss function is closely related to the DSNE loss function (\ref{eq:DSNE-loss}). To see this, we decompose the DSNE loss function as follows, 
\begin{equation}
\label{eq:DSNE-loss-decomposition}
\begin{array}{ll}
 C_{DSNE,i } &:= \sum_{ j \in \text{i's near neighbors}}  \tilde p_{j|i} \log \frac{  p_{j|i}  } {q_{j|i}} \\
 	&=   - \sum_{ j \in \text{i's near neighbors}}  \tilde p_{j|i} \log q_{j|i} - \tilde H      \\
	&=  - \sum_{ j \in \text{i's near neighbors}}  \tilde p_{j|i}  \log \frac{ \exp( -\beta_{y,i} || \hat w_i - \Delta \hat y_{ij}||^2 ) } { Z_{y,i} }  - \tilde H \\
	&=  \beta_{y,i}  \sum_{ j \in \text{i's near neighbors}}      \tilde p_{j|i} || \hat w_i - \Delta \hat y_{ij}||^2   + \log  Z_{y,i} - H
\end{array}
\end{equation}
where $\tilde H = - \sum_{ j \in \text{i's near neighbors}}  \tilde p_{j|i}  \log  p_{j|i}$ is a scaled entropy of $\tilde P_i$ do not involve with $w_i$, which can be viewed as a constant. If we drop out the normalization term  $\log  Z_{y,i}$, we will get  almost the same loss function as scVeloEmbedding's, $ \sum_{ j \in \text{i's near neighbors}}     \tilde p_{j|i}  || \hat w_i - \Delta \hat y_{ij}||^2 $. This may be the reason why scVeloEmbedding work relatively well in practice. We note that there are several differences between DSNE and scVeloEmbedding. First, DSNE use the local average $\bar y_i$ rather than the global average direction $\bar y_{full,i}$, we choose the local average direction is because that the usually used dimension reduction algorithm, e.g. t-SNE, UMAP, preserve local structure better than the global structure. So the local average seems more reasonable than the global average. Second, DSNE use the unit direction $\hat w_i, \; \Delta \hat y_{ij}$, while the scVeloEmbedding use the un-normalized direction $w_i$ and $  \hat y_{ij}  -  \bar y_{full, i }$. Using the unit direction is more reasonable since we can not tell which is better $  \hat y_{ij}  -  \bar y_{full, i }$ for different $j$ ( If $\hat y_{ij}$  close to the mean direction $\bar y_{full, i }$, it will have little norm if $ \bar y_{full, i }$ was not near zero,  so $ \hat y_{ij}  -  \bar y_{full, i }$ will contribute little to $w_i$),  also in the DSNE loss, the unit direction $ \Delta \hat y_{ij}$  is comparable with the unit direction $\hat w_i$. Although these minor differences may contribute to work better, the essential difference between DSNE and scVeloEmbedding is that DSNE seek to find a linear relation between the sphere distance of velocity and the directions to near neighbors , i.e.  $\beta_{x,i} || \hat v_i - \Delta \hat x_{ij}||^2 =  \beta_{y,i} || \hat w_i -  \Delta \hat y_{ij}||^2 $. If the dimension reduction algorithm will preserve the sphere distance  up to a scalar in the local structure, i.e,  $ || \bar x_i - \Delta \hat x_{ij}||^2 =  \alpha || \hat y_i -  \Delta \hat y_{ij}||^2, \; j \in \text{i's neighbors}$, then we  can figure out the direction $\hat w_i$  of velocity $v_i$ in the low-dimension space with the DSNE algorithm. The scVeloEmbedding  relying on the  probability weighting of the directions $  \hat y_{ij}  -  \bar y_{full, i }$ is a suboptimal choice.

%
%
\section{Approximate DSNE} 
Based the above discussion, we can use the following formula to compute the velocity embedding $w_i$ approximately. 
\begin{equation}
\label{eq:DSNE-approximate}
\begin{array}{ll}
 \tilde w_i = \sum_{ j \in \text{i's near neighbors}}  \tilde p_{j|i} \Delta \hat y_{ij}\\
 \hat w_i = \frac {\tilde w_i} {|| \tilde w_i ||} \\
 w_i = [\frac 1 N  \sum_{j =1}^N \frac{ ||y_{j}||+d } {  ||x_{j}|| +D}] ||v_i||   \hat w_i 
\end{array}
\end{equation}
We term this method by the name DSNE\_approximate, which is implemented in dsne package. In the numerical experiment, its performance is a little  better than scVeloEmbedding, while less performed  as well as DSNE. For clarity, in the following experiments, we omits its numerical outputs.

\section{Experiments}
To evaluate the performance of  DSNE, we performed experiments in on the simulated data and the Pancreas scRNA-seq data. Note that there seems no velocity embedding algorithm to be compared with (I do not do a full survey), we only compare with the simple intuitive algorithm  scVeloEmbedding presented in scVelo (\cite{dynamical-RNA-velocity}). 

\subsection{Simulated Data}

\subsubsection{Simulated Data with Exact Velocity and Approximate Velocity Embedding. }
\label{sec:approximate-simulation}
To test the performance of DSNE and compare it with the scVeloEmbedding, we generated the simulated data with the exact velocities and then go along with the velocity with one time step one-by-one  from three  start  points to get the data points.  
\begin{enumerate}
\item  Generate the  velocity $V \in \mathbb R^{N \times D}$ by random sampling from the normal distribution, i,e, $ V_{il} \sim \mathcal N(0,36),  \; i=1 \ldots, N; \; l = 1, \ldots, D$. where we take $N = 3N_s$;  
\item  Choose three start points of data points $x_{start,1} = \mathbf 0$, $x_{start,2}  =  50* \mathbf 1$.
    $x_{start,3}  =160 * \mathbf 1$.  where $ \mathbf 0$ is the zeros vector with length $D$  and $\mathbf 1$ is the ones vector with length $D$. 
\item  Generate the data points $X \in \mathbb R^{N \times D}$  from the three starting points and moving along the velocity $v_i$ one by one. i.e. 
\begin{equation*}
\begin{array}{l}
    	x_1 = x_{start, 1}\\
	x_{N_s+1} = x_{start, 2}\\
	x_{2N_s+1} = x_{start, 3}\\
	x_{i+1} = x_i + v_i, \; i = 1, \ldots, N_s-1 \\
	x_{N_s + i+1} = x_{N_s + i}  + v_{N_s + i}, \; i = 1, \ldots, N_s-1 \\
	x_{2N_s + i+1} = x_{2N_s + i}  + v_{2N_s + i}, \; i = 1, \ldots, N_s-1   	
\end{array}
\end{equation*}
\end{enumerate}
By changing the number  of point  $N=N_s*3$ and the dimension of $D$, we can get different sizes of data. 

Note that since we have $x_{i+1} = x_i +v_i, \; i=1, \ldots, N_s-1, N_s +1 , \ldots, 2*N_s-1,   2*N_s+1, \ldots, 3*N_s-1$, we have the reasonable guess that $y_{i+1} = y_i + w_i \; i=1, \ldots, N_s-1, N_s +1 , \ldots, 2*N_s-1,   2*N_s+1, \ldots, 3*N_s-1$, 
so we use the  $w_{true, i} := \frac { y_{i+1} - y_i   } {|| y_{i+1} - y_i|| }$ as the true direction of  velocity embeddings. With   $w_{true, i} \; i=1, \ldots, N_s-1, N_s +1 , \ldots, 2*N_s-1,   2*N_s+1, \ldots, 3*N_s-1$, we define the following accuracy of velocity embeddings $W$,
\begin{equation}
\label{eq:accuracy-approximate}
accu :=  \frac 1 {N-3} \sum_{i=1}^{N_s-1} ( \langle \frac { w_i } { ||w_i||} ,   \frac { w_{true, i} } { ||w_{true, i}||} \rangle
	+ \langle \frac { w_{i+N_s} } { ||w_{i+N_s}||} ,   \frac { w_{true,i+N_s} } { ||w_{true, i+N_s}||} \rangle   
	+ \langle \frac { w_{i+2N_s} } { ||w_{i+2N_s}||} ,   \frac { w_{true,i+2N_s} } { ||w_{true, i+2N_s}||} \rangle    ) 
\end{equation}

We first give a small simulated data with $N=150$ and $d=30$ to check that DSNE can do the correct work and compare it with the result of scVeloEmbedding.  

\begin{figure}
  \centering
\includegraphics[width=150mm]{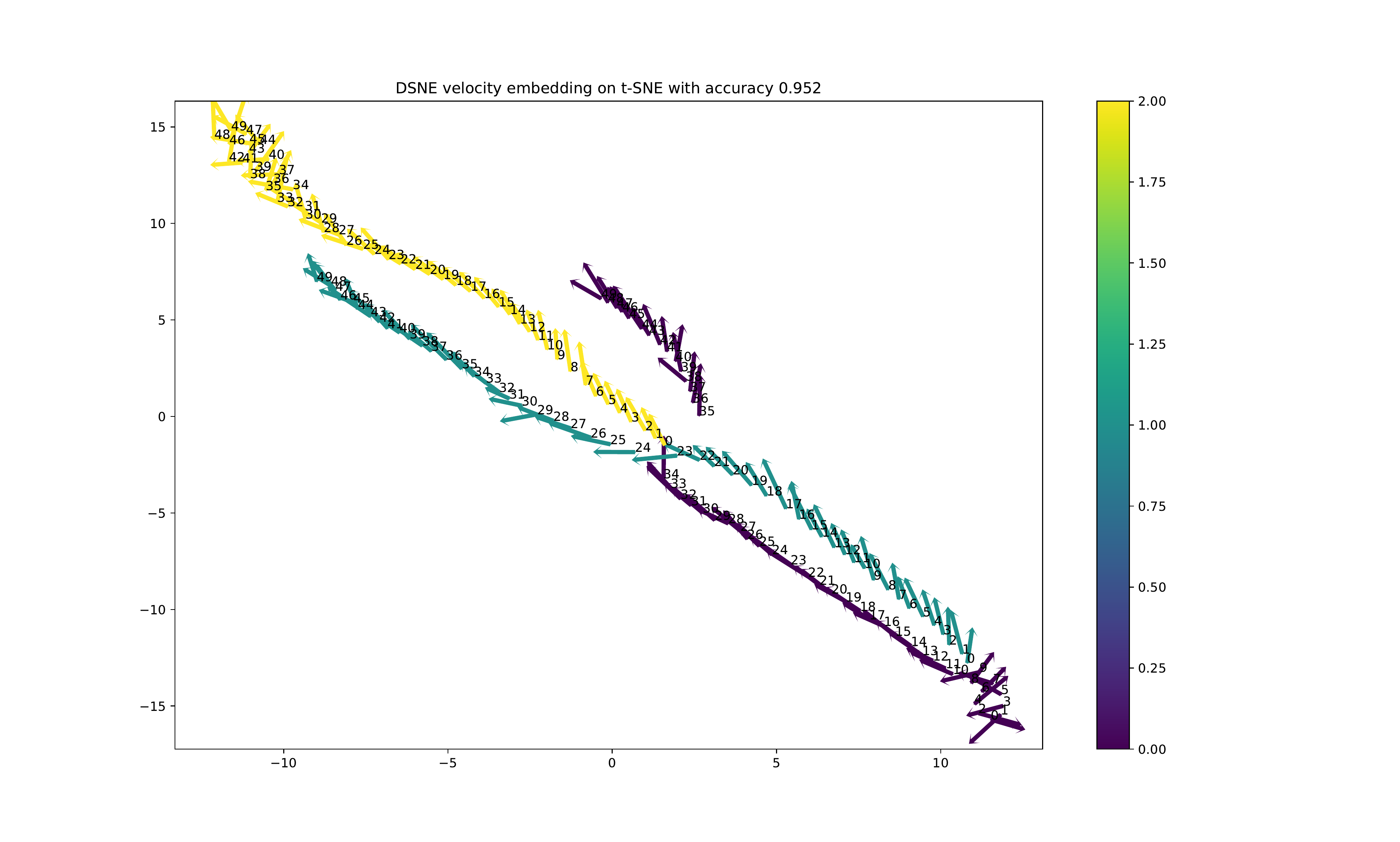}
\includegraphics[width=150mm]{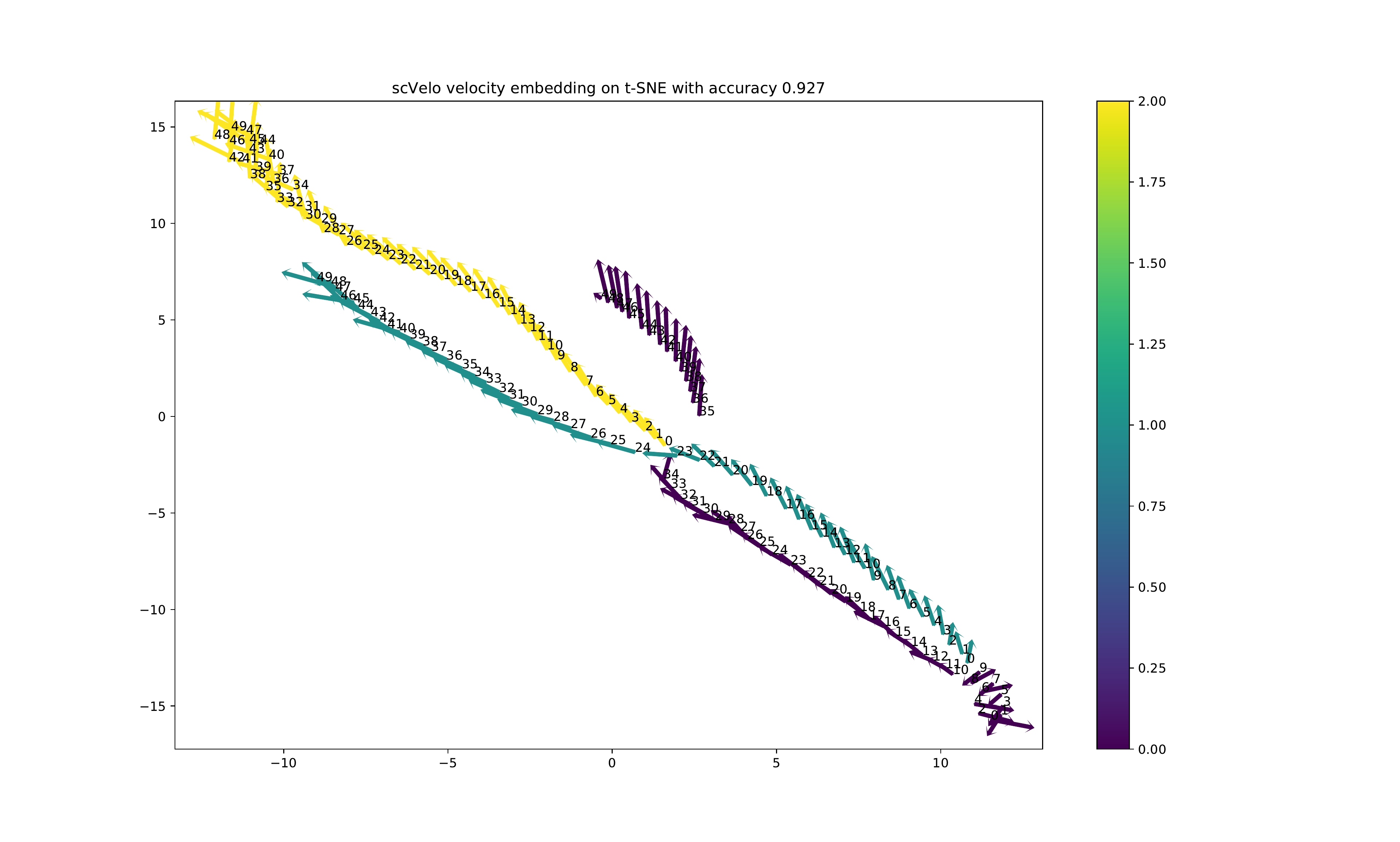}
  \caption{The toy example on  the  simulated data based on exact data points and velocities with $N=150$, $D=30$. The top figure shows the results of DSNE on the t-SNE map points, which has the accuracy  $0.952$ of velocity embeddings compared with the approximate true direction on the t-SNE map points. The bottom figure shows the results of scVeloEmbedding on the t-SNE map points, which has the accuracy  $0.927$ of velocity embeddings compared with the approximate true direction on the t-SNE map points. }
  \label{fig:toy-approximate-tsne-arrow}
  \end{figure}

\begin{figure}
  \centering
\includegraphics[width=150mm]{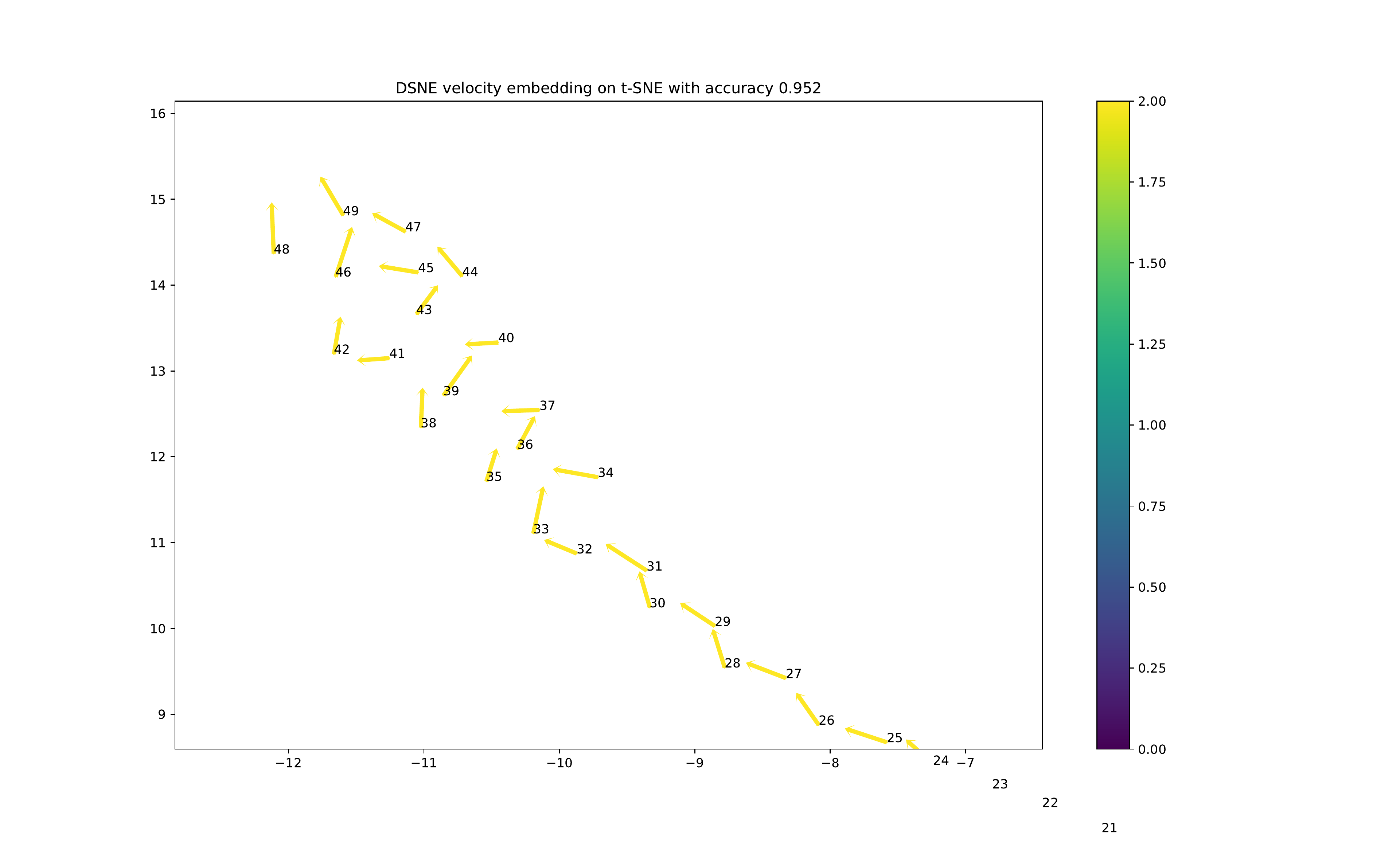}
\includegraphics[width=150mm]{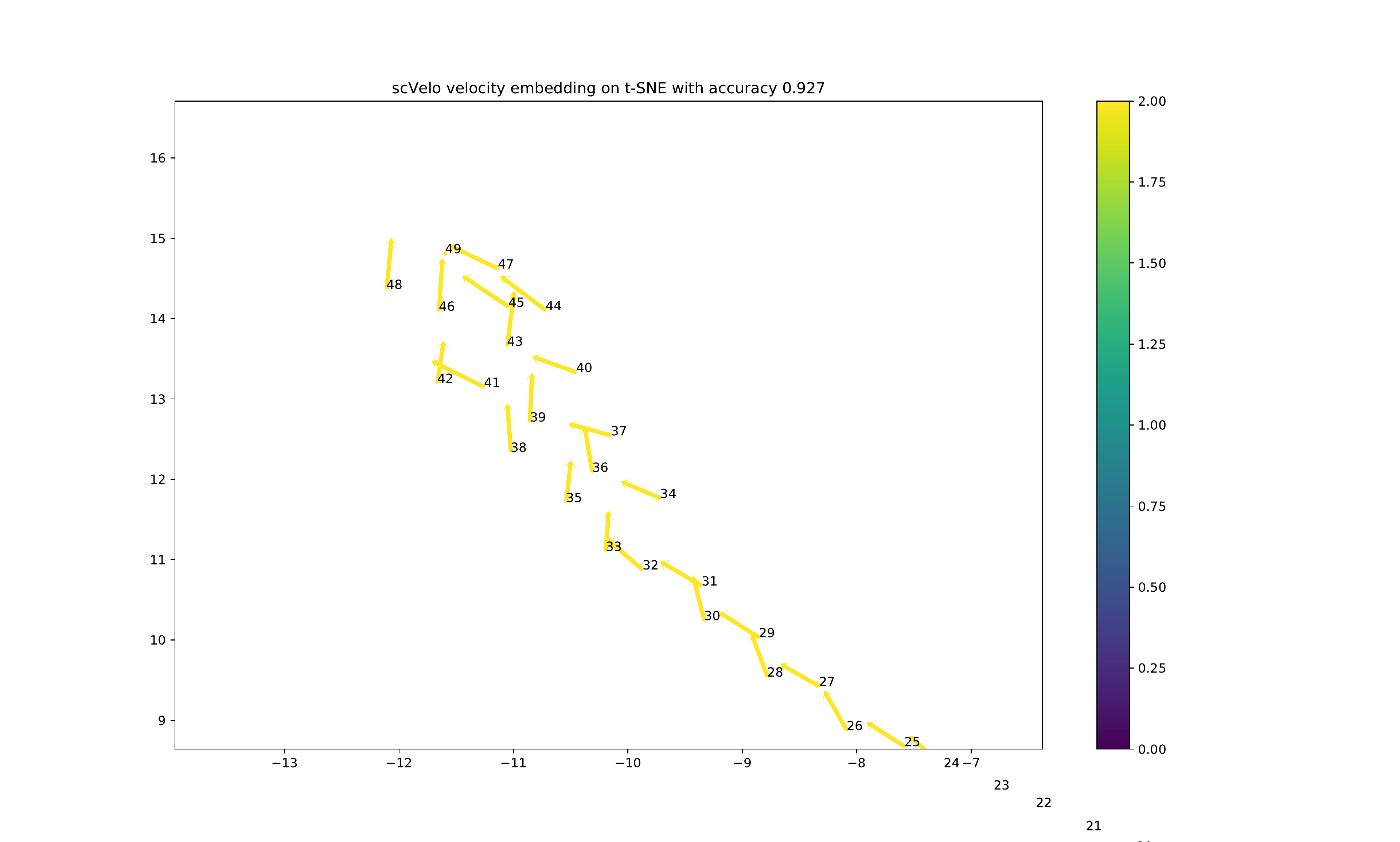}
  \caption{The enlarged plot of  top left part  of Figure~\ref{fig:toy-approximate-tsne-arrow}}
  \label{fig:toy-approximate-tsne-arrow-top-left}
  \end{figure}

\begin{figure}
  \centering
\includegraphics[width=150mm]{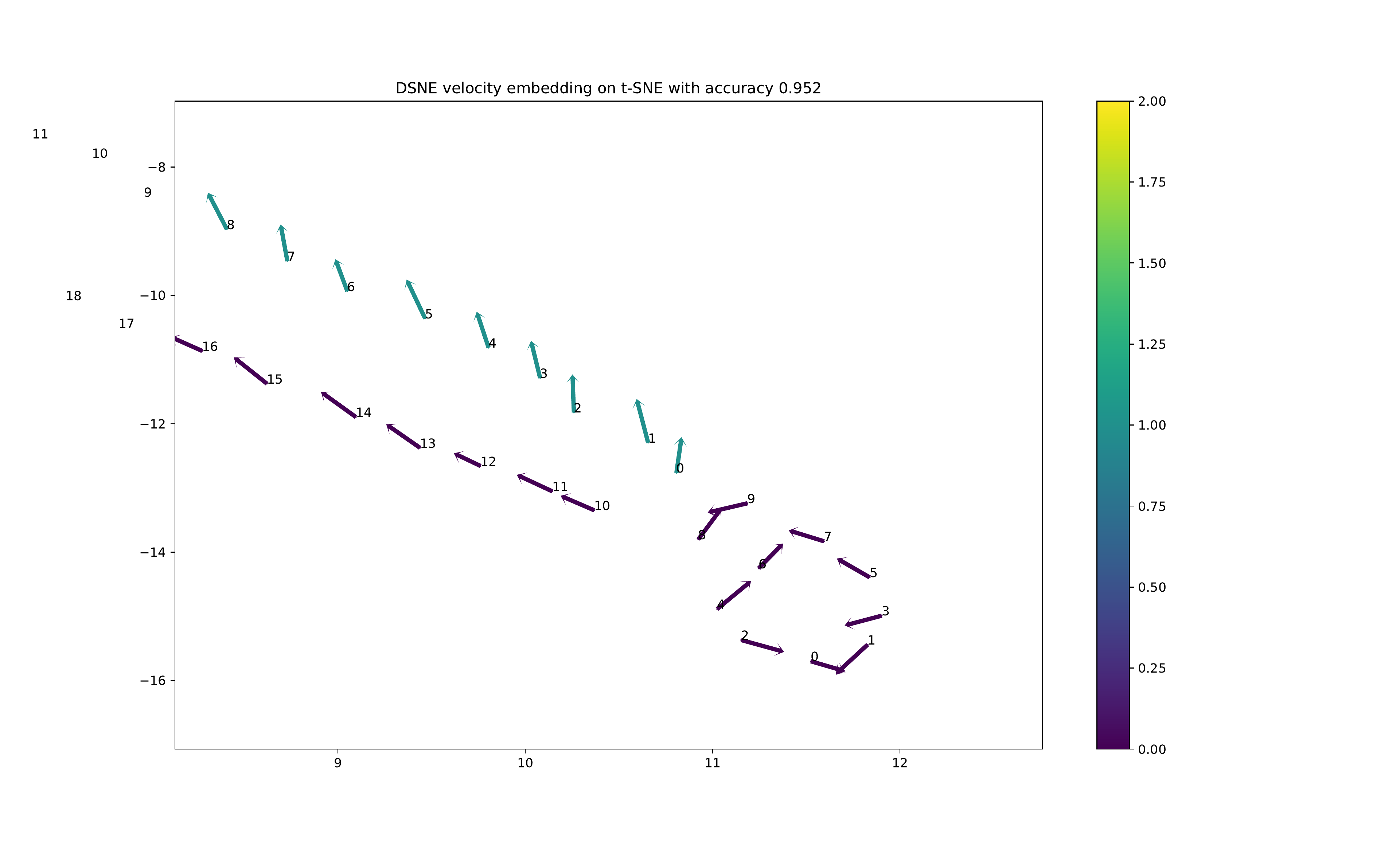}
\includegraphics[width=150mm]{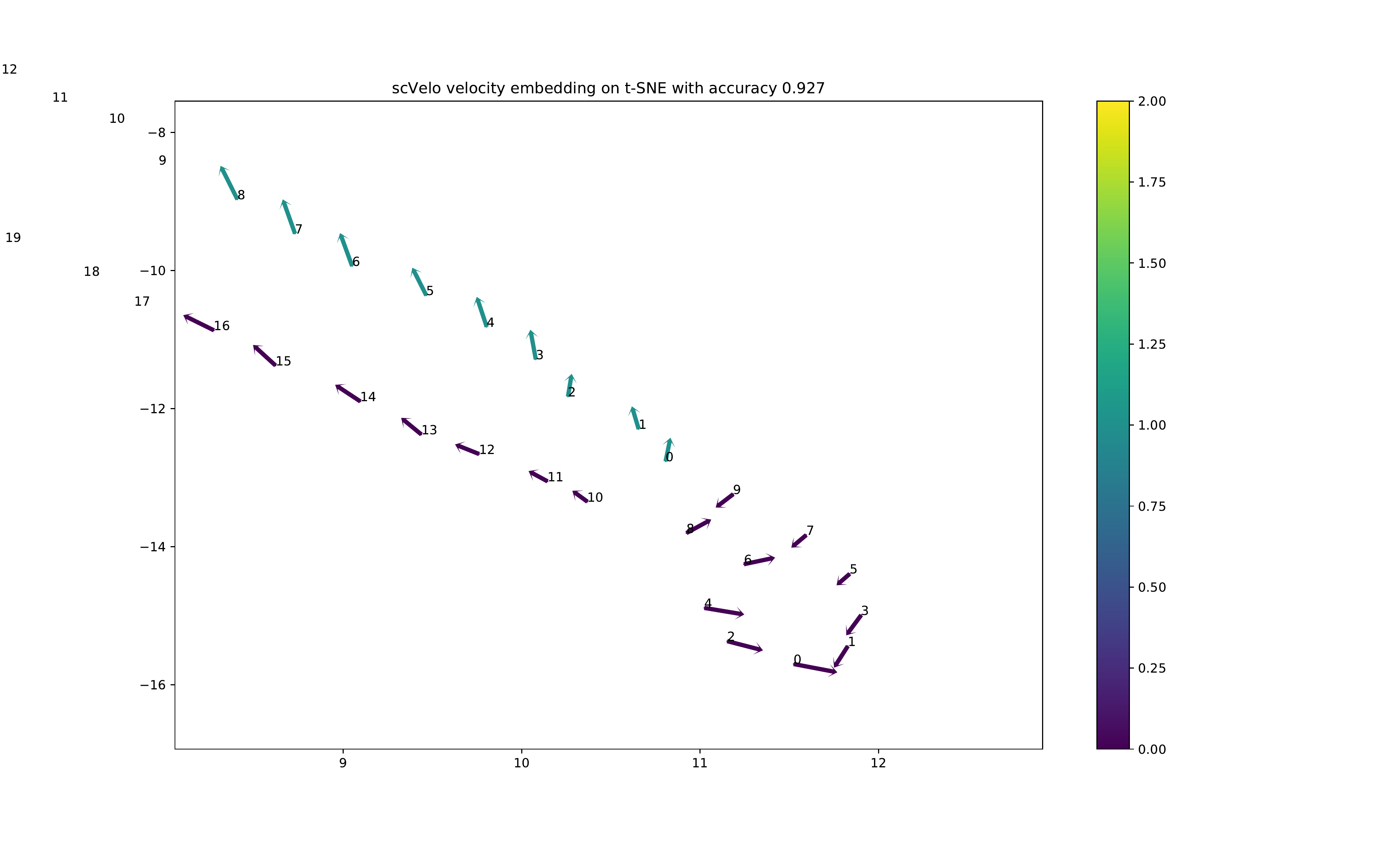}
  \caption{The enlarged plot of  bottom right part  of Figure~\ref{fig:toy-approximate-tsne-arrow}}
  \label{fig:toy-approximate-tsne-arrow-bottom-right}
  \end{figure}

\begin{figure}
  \centering
\includegraphics[width=150mm]{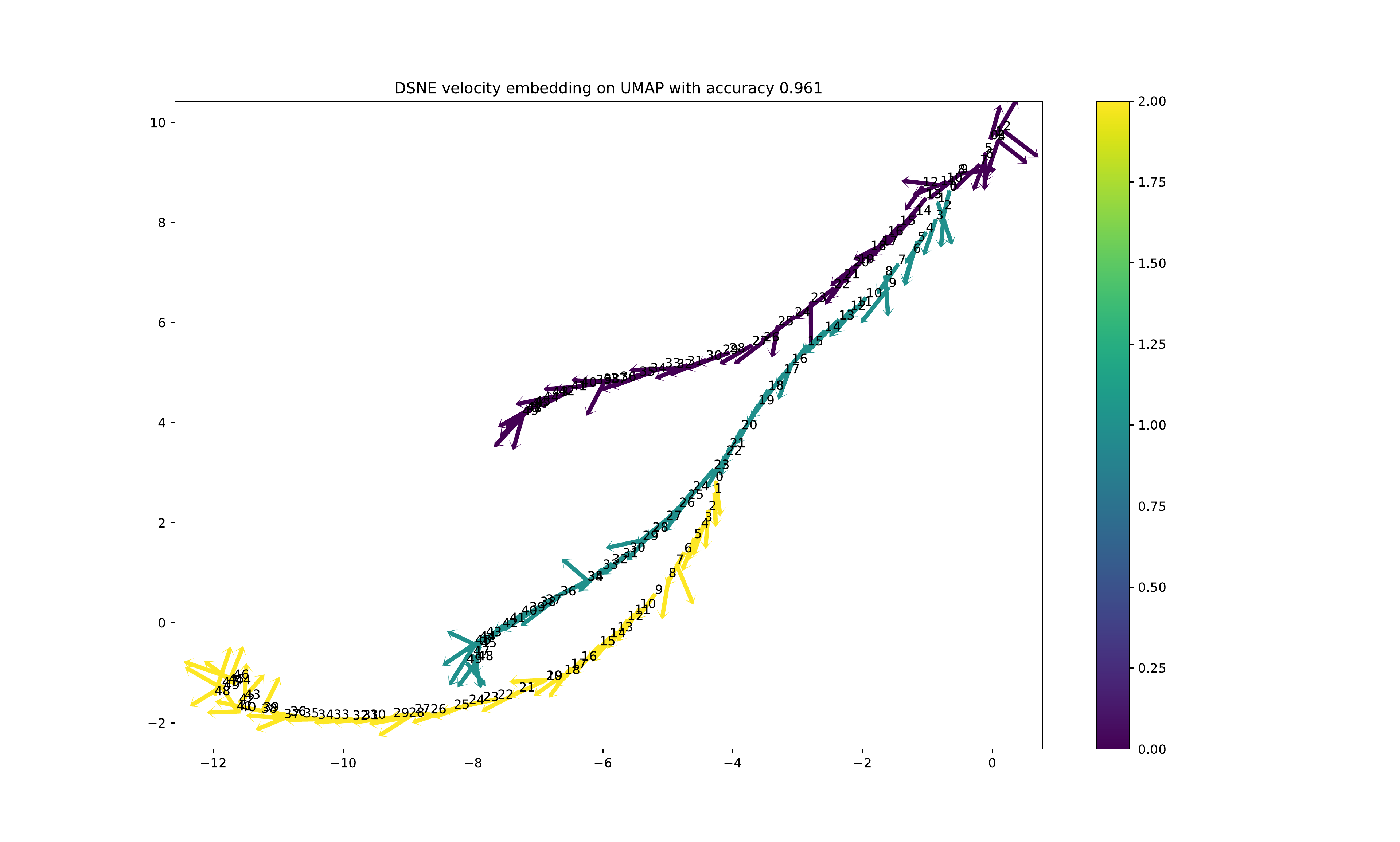}
\includegraphics[width=150mm]{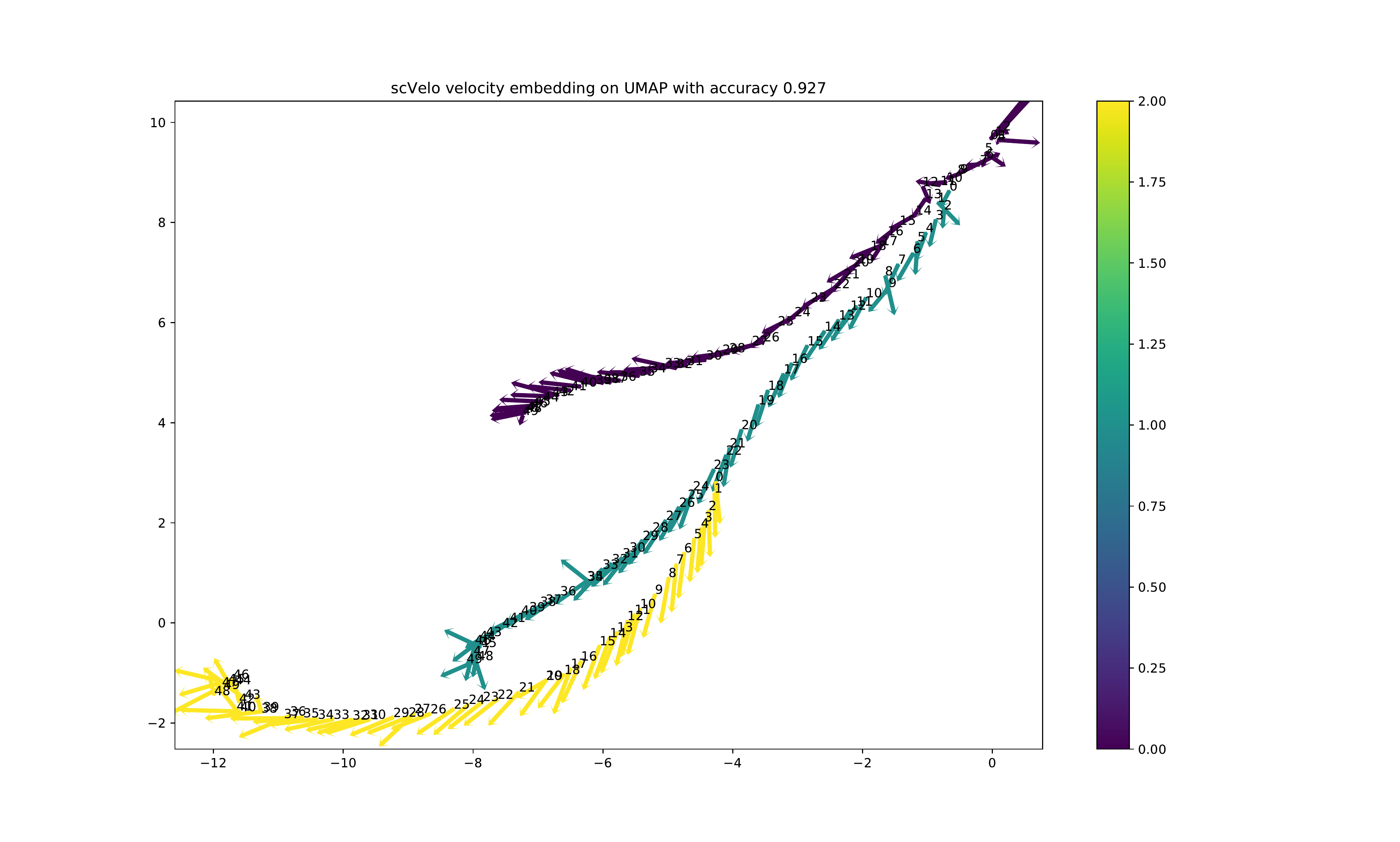}
  \caption{The toy example on  the  simulated data based on exact data points and velocities with $N=150$, $D=30$. The top figure shows the results of DSNE on the UMAP map points, which has the accuracy  $0.961$ of velocity embeddings compared with the approximate true direction on the UMAP map points. The bottom figure shows the results of scVeloEmbedding on the UMAP map points, which has the accuracy  $0.927$ of velocity embeddings direction compared with the approximate true direction on the UMAP map points.  Zoom in for details.  }
  \label{fig:toy-approximate-umap-arrow}
  \label{fig:LISI-cell-line}
  \end{figure}

\begin{figure}
  \centering
\includegraphics[width=150mm]{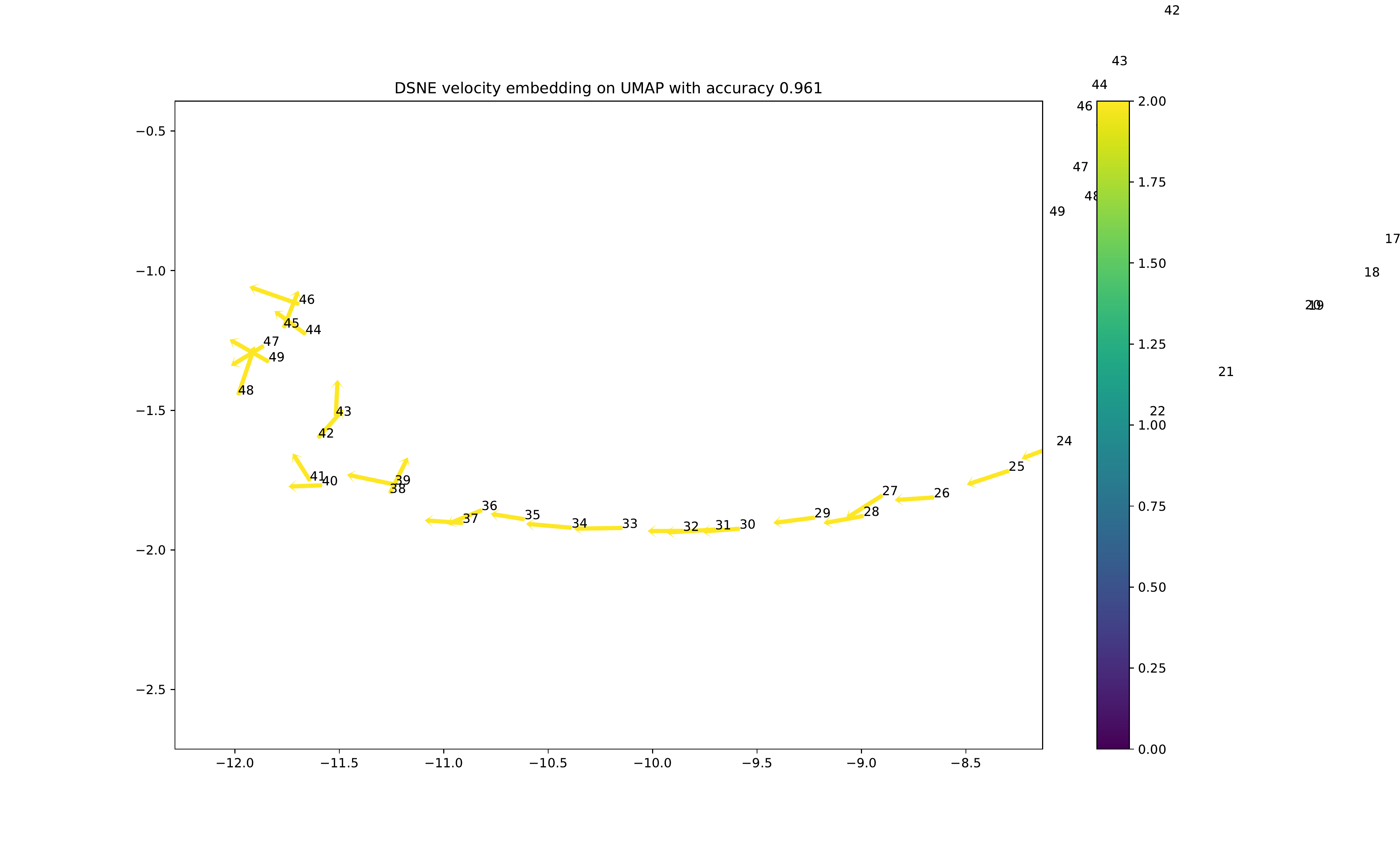}
\includegraphics[width=150mm]{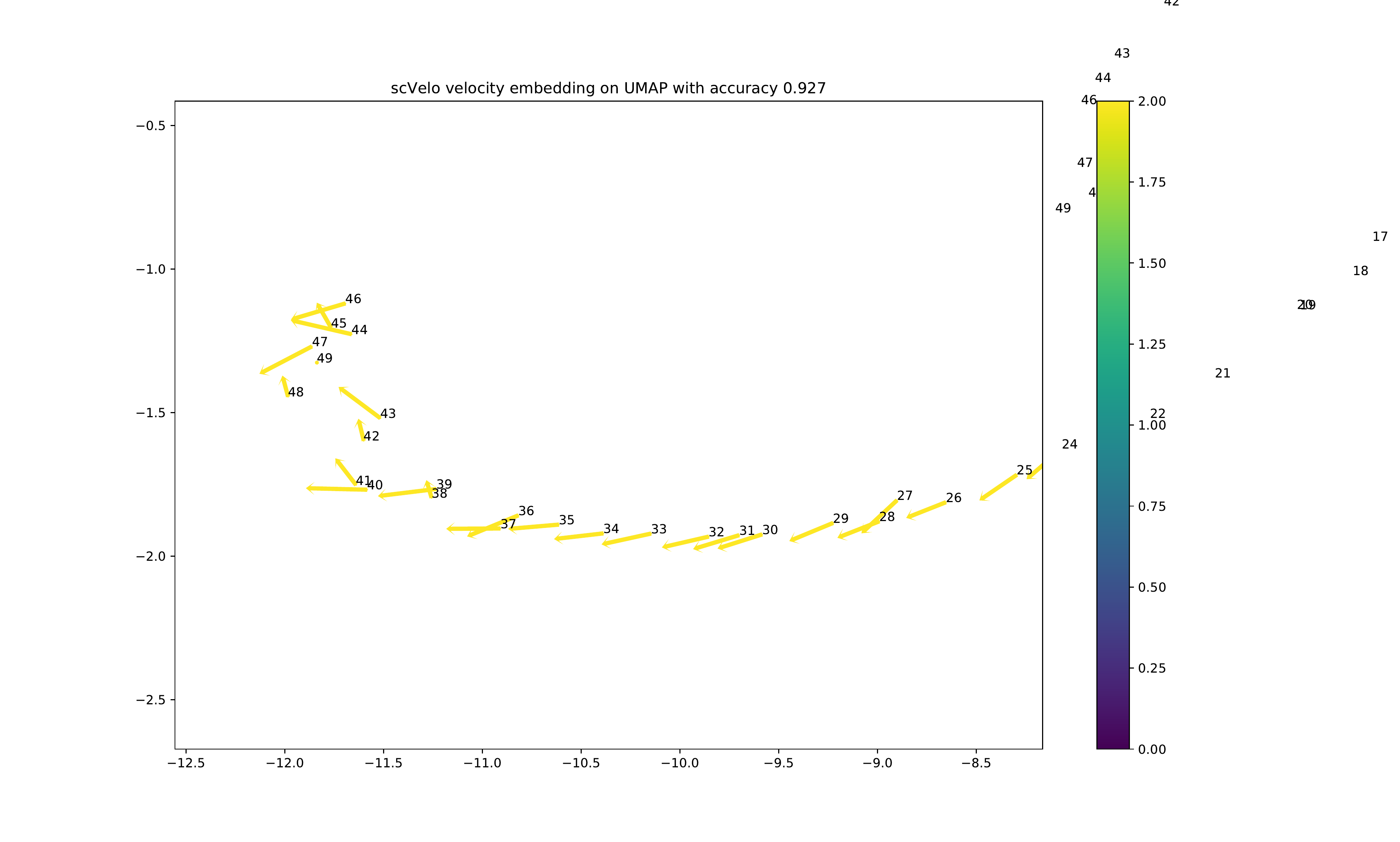}
  \caption{The enlarged plot of  top left part  of Figure~\ref{fig:toy-approximate-umap-arrow}}
  \label{fig:toy-approximate-umap-arrow-top-left}
  \end{figure}

\begin{figure}
  \centering
\includegraphics[width=150mm]{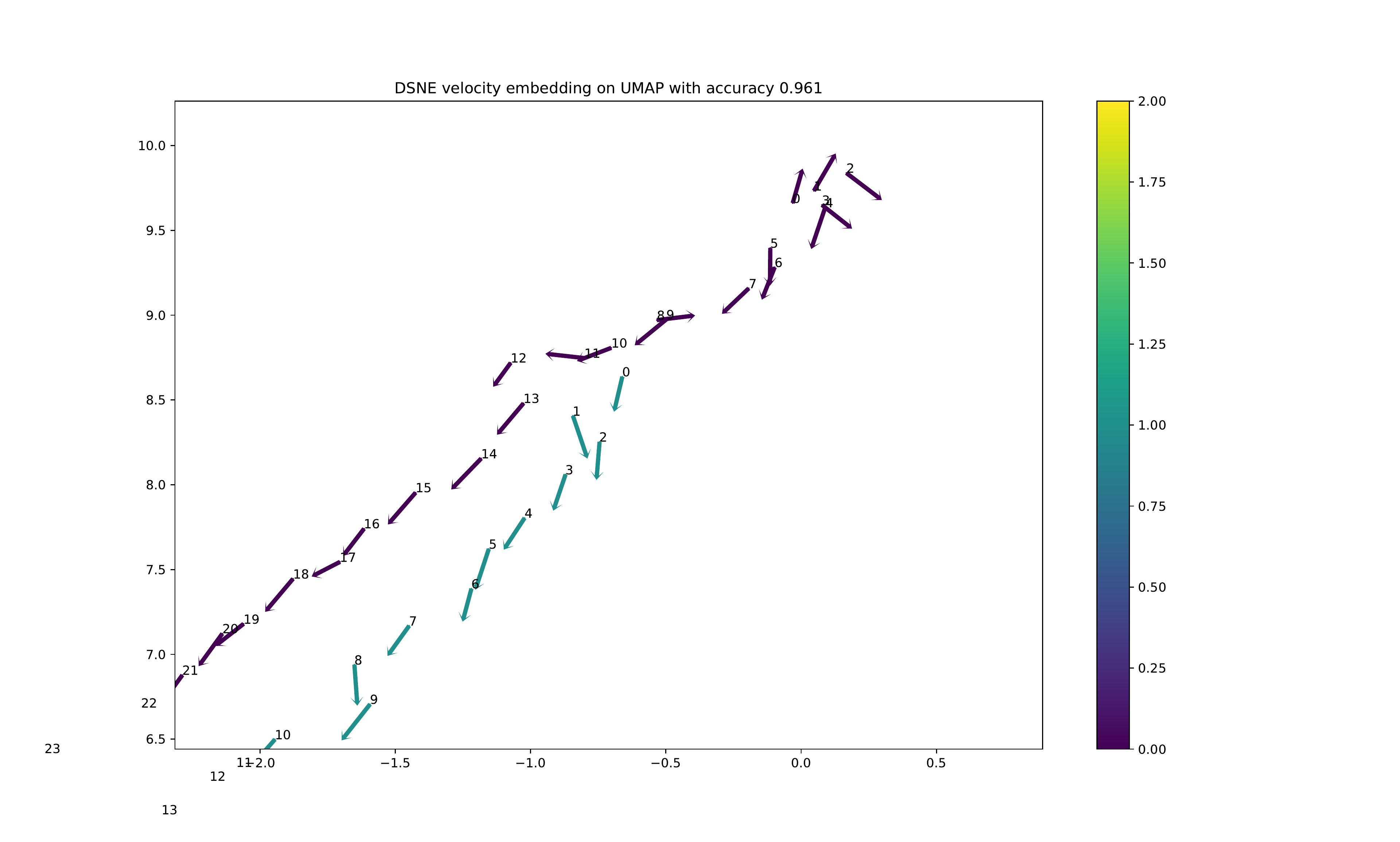}
\includegraphics[width=150mm]{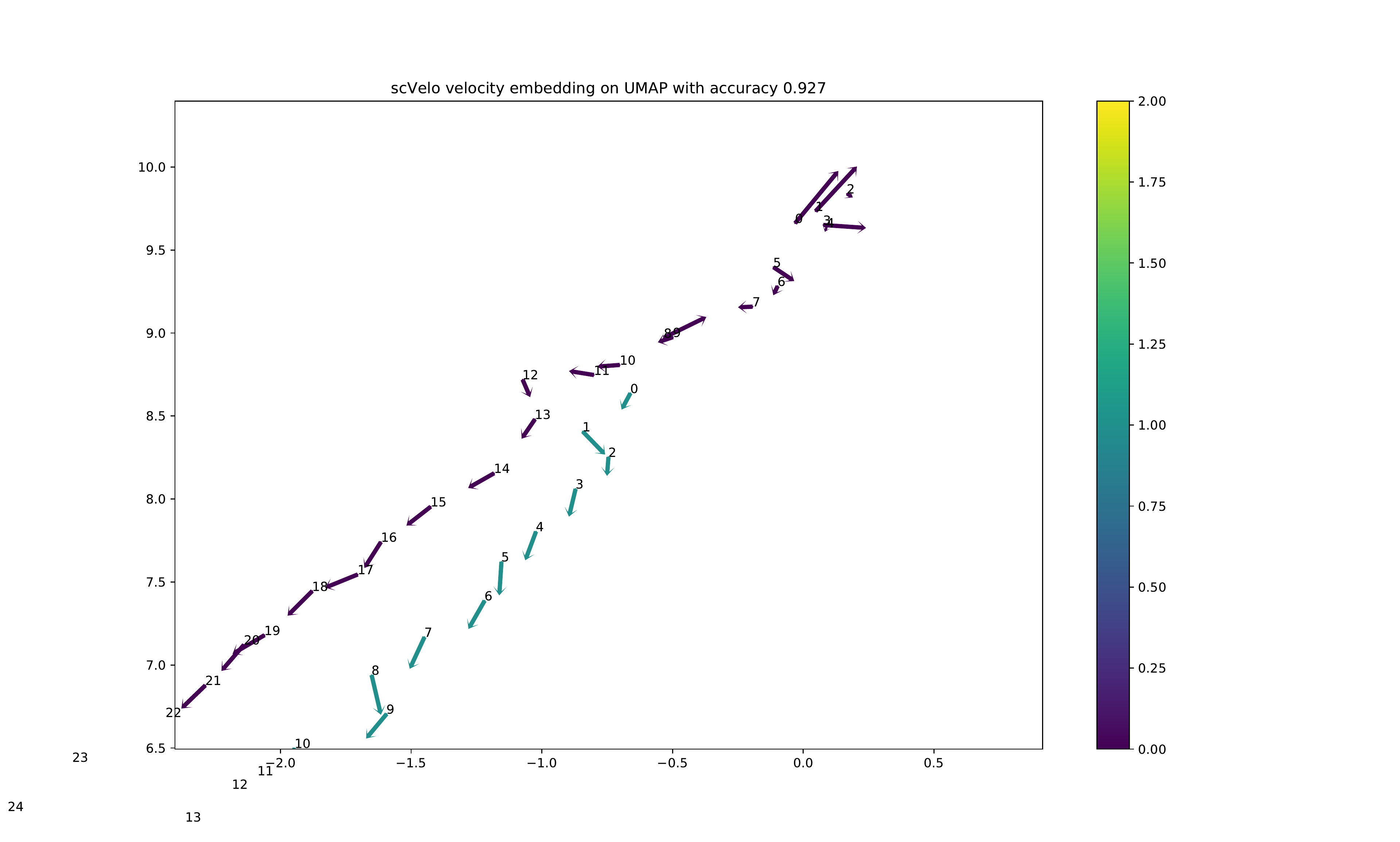}
  \caption{The enlarged plot of  bottom right part  of Figure~\ref{fig:toy-approximate-umap-arrow}}
  \label{fig:toy-approximate-umap-arrow-bottom-right}
  \end{figure}

We set the the parameters of  DSNE with learning rate $\eta=0.1$, in the first $250$ steps momentum $\gamma = 0.5$ and the later steps $\gamma = 0.8$, the perplexity $Perplexity = 1$, $K=6$.
 We run BH-SNE (\cite{BH-SNE}) (\url{https://github.com/danielfrg/tsne}) with parameter $\theta=0.5, \; perplexity=20$, UMAP (\cite{UMAP}), to get the low-dimensional embedding $Y$, respectively. 
 And then on these embeddings to learn the velocity embedding on the low-dimensional space with DSNE and scVeloEmbedding. The results are presented in Figure~\ref{fig:toy-approximate-tsne-arrow} ( with enlarged  local parts  Figure~\ref{fig:toy-approximate-tsne-arrow-top-left} and Figure~\ref{fig:toy-approximate-tsne-arrow-bottom-right}) on the t-SNE map points; in Figure~\ref{fig:toy-approximate-umap-arrow} ( with enlarged  local parts  Figure~\ref{fig:toy-approximate-umap-arrow-top-left} and Figure~\ref{fig:toy-approximate-umap-arrow-bottom-right}) on the UMAP map points. 
 On both t-SNE and UMAP map points, DSNE get a more accurate velocity embeddings than scVeloEmbeddin's. This can be verified  with the accuracy and  the velocity arrows on Figure~\ref{fig:toy-approximate-tsne-arrow-top-left}, e.g, on the point $36$, DSNE will put the velocity to point $37$, while the velocity embedding of  scVeloEmbedding  on point $36$  was  point to  point $40$, which is not correct. The similar phenomena were happened  on some other points. 

%
%
%
%
To more throughly test the performance of DSNE and  compare with the scVeloEmbedding, we simulate the data $10$ times with different dimensions of $N,D$, and compute the low-dimension map points with BH-SNE (\cite{BH-SNE}) (\url{https://github.com/danielfrg/tsne}) with parameter $\theta=0.5, \; Perplexity=20$, UMAP (\cite{UMAP}), run DSNE and scVeloEmbedding on the same simulated data with same map points each time. For DSNE, we use $K=6, \; perplexity =1$ on the setting $N=150, \; D=30$ and $K=16, \; perplexity =3$ for all other settings. For scVeloEmbedding, we run it with the default parameter in the scVelo (\url{https://github.com/theislab/scvelo}) package. Finally, we output the mean and standard deviation of the accuracy in Table~\ref{tab:accuracy-approximate}. From the table, we see that DSNE do a better work than scVeloEmbedding for all the test settings.

\begin{table}
 \caption{Accuracy of Direction of Velocity Embeddins on the Simulation Data with Approximate Map Points and Velocity Embeddings }
  \centering
   {
  \begin{tabular}{lcc}
    \toprule
    Dimension (Reduction Method)     & Accuracy mean (std) of DSNE &  Accuracy mean (std) of scVeloEmbedding  \\
    \midrule
     $N=150, \; D=30$ \;( UMAP) & $0.965~(	0.007)	$ & $0.936~(	0.011)$	\\
     $N=150, \; D=30$ \; ( t-SNE ) & $0.959~(	0.006)	$ & $0.925~(	0.004)$	\\
      $N=1500, \; D=10$ \;( UMAP) & $0.985~(	0.002)	$ & $0.944~(	0.009)$	\\
     $N=1500, \; D=10$ \; ( t-SNE ) & $0.982~(	0.003)	$ & $0.936~(	0.001)$	\\
     $N=1500, \; D=300$\;( UMAP) & $0.988~(	0.002)	$ & $0.969~(	0.003)$	\\
     $N=1500, \; D=300$ \; ( t-SNE ) & $0.985~(	0.002)	$ & $0.961~(	0.002)$	\\
      $N=15000, \; D=300$  \;( UMAP ) & $0.988~(	0.001)	$ & $0.969~(	0.001)$	\\
     $N=15000, \; D=300$  \; ( t-SNE ) & $0.993~(	0.001)	$ & $0.981~(	0.001)$	\\
    \bottomrule
  \end{tabular}
  }
  \label{tab:accuracy-approximate}
\end{table}

\begin{figure}
  \centering
\includegraphics[width=150mm]{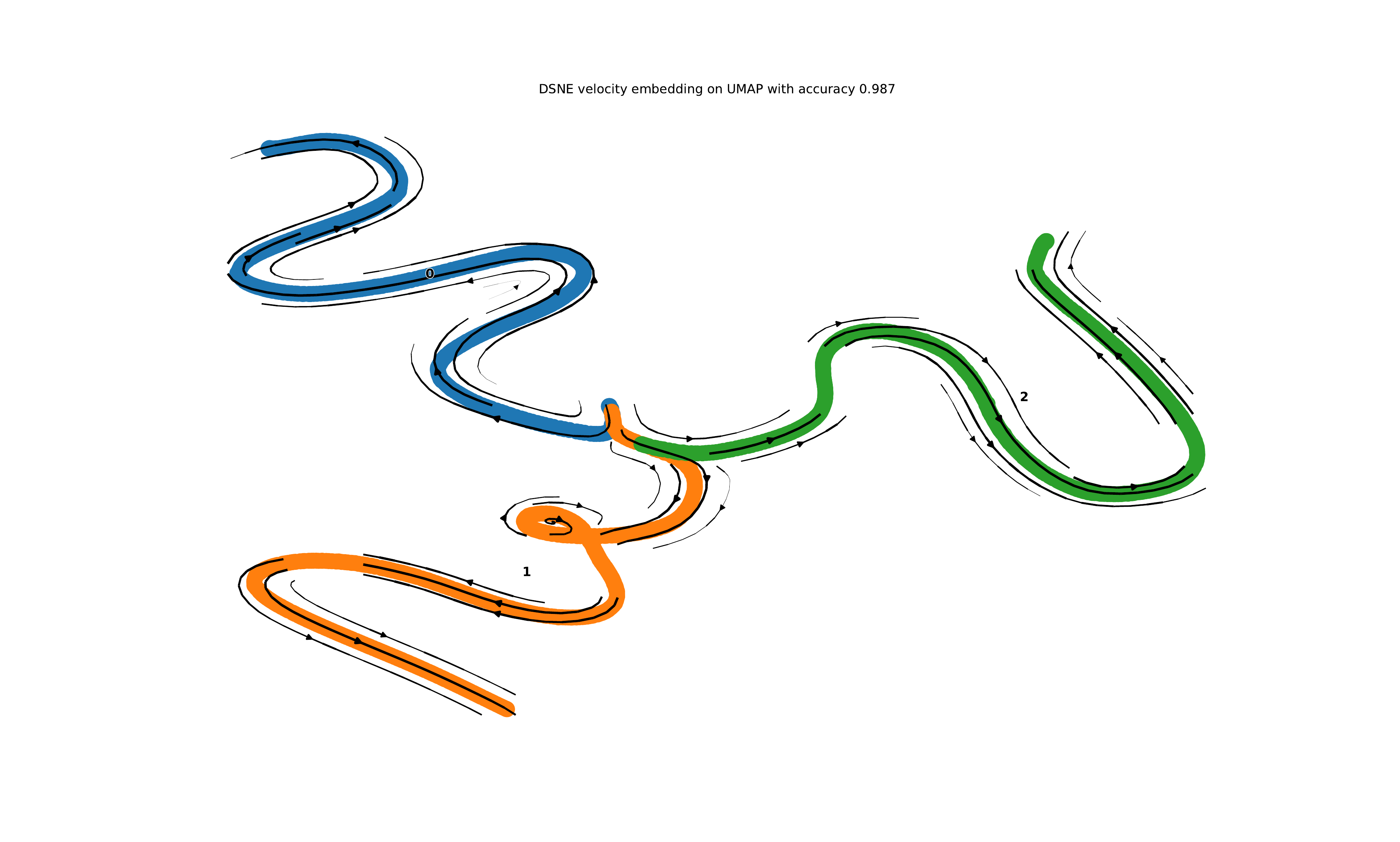}
\includegraphics[width=150mm]{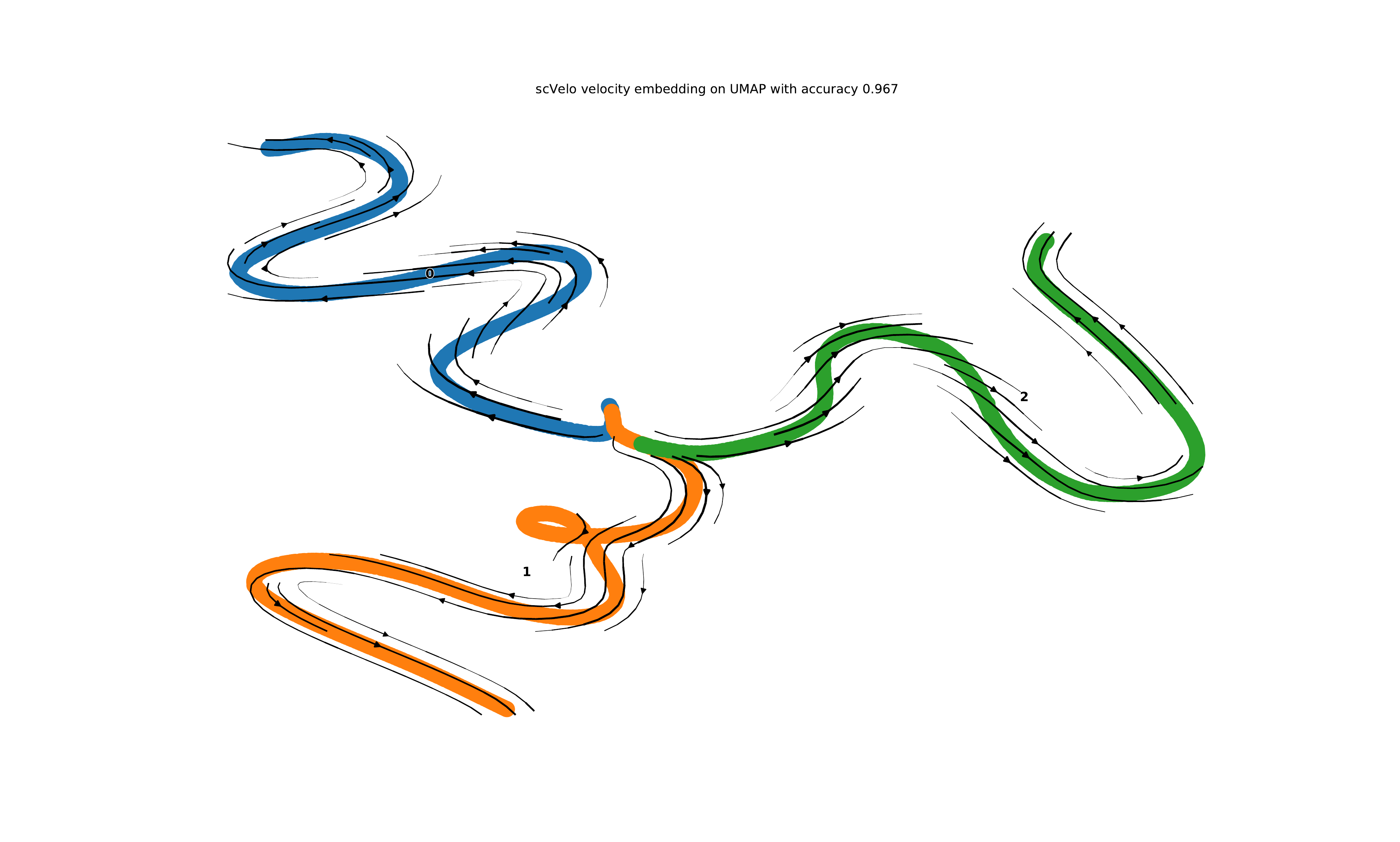}
  \caption{The stream plot of  the simulated data based on exact data points and velocities  with $N=1500$, $D=300$  and $d=2$. The top figure shows the stream plot of  the velocity embeddings output by  DSNE on the UMAP map points of the data points, which has the  direction accuracy $0.987$ compared with the approximate true velocity embeddings;  the bottom figure shows the stream plot of the velocity embeddings output by scVeloEmbeddings on the UMAP map points of the data points, which has the direction accuracy $0.967$.  Zoom in for details.  }
  \label{fig:approximate-embeddings-umap-stream}
  \end{figure}

\begin{figure}
  \centering
\includegraphics[width=150mm]{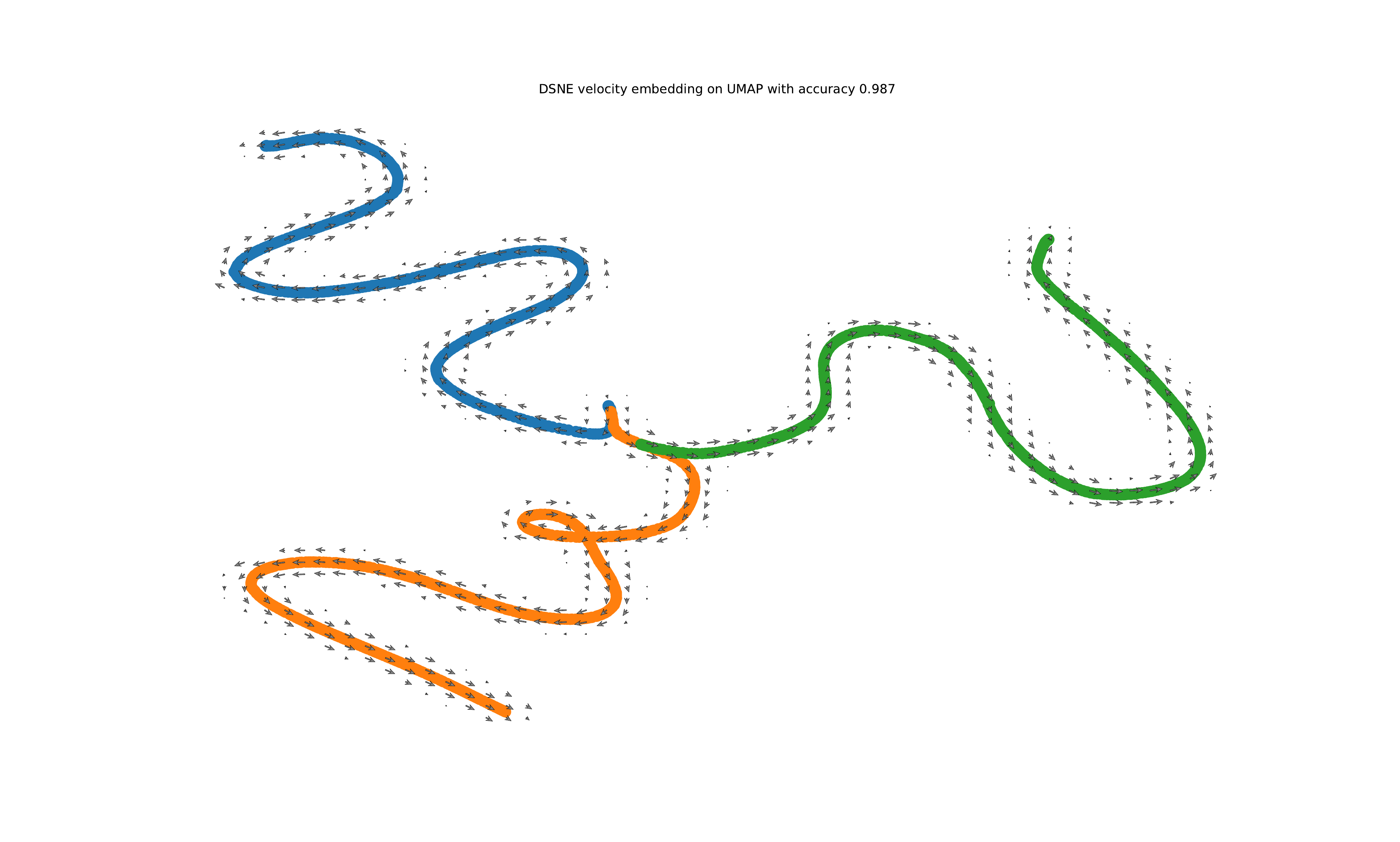}
\includegraphics[width=150mm]{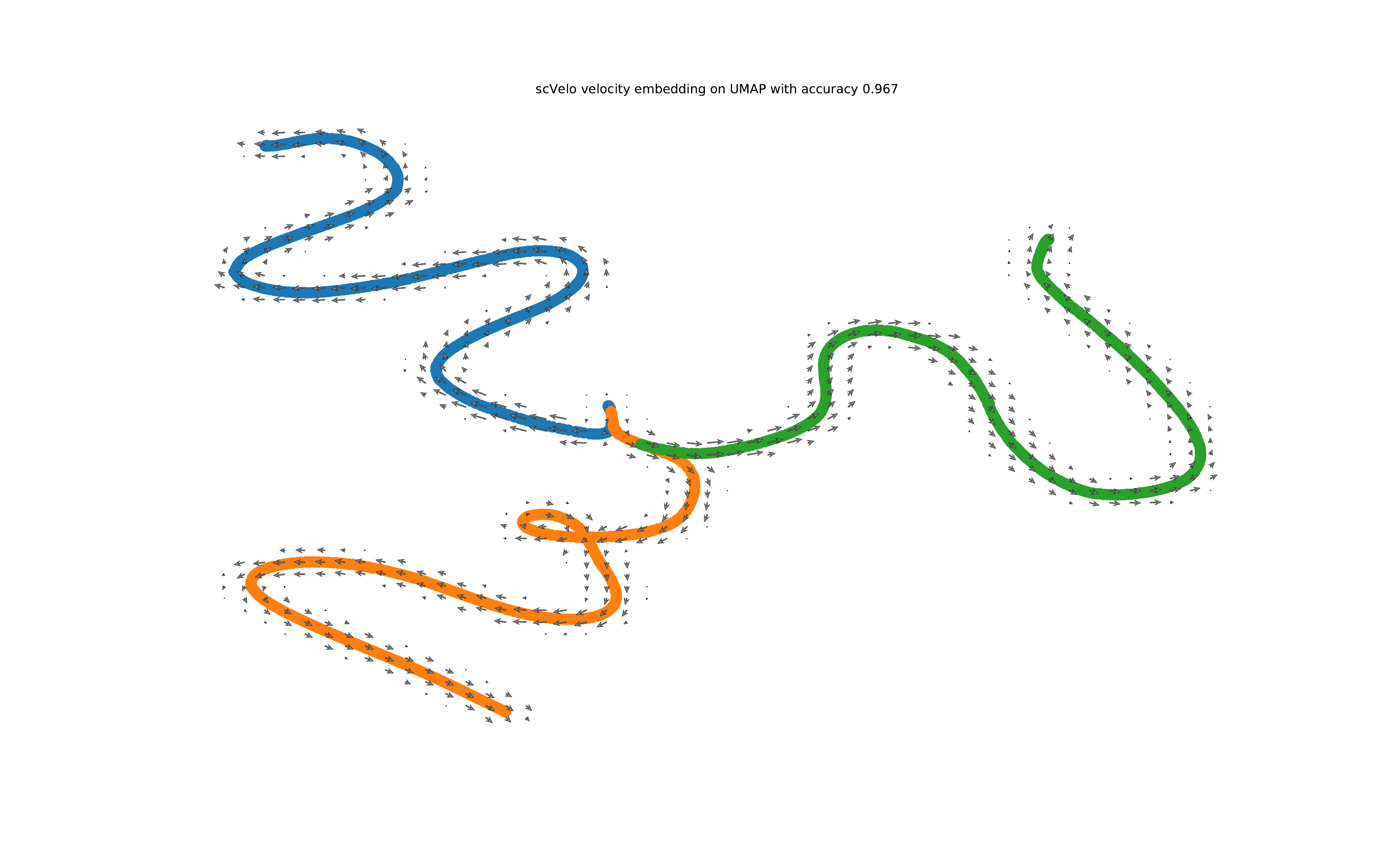}
  \caption{The grid plot of  the simulated data based on exact data points and velocities  with $N=1500$, $D=300$  and $d=2$. The top figure shows the grid plot of  the velocity embeddings output by  DSNE on the UMAP map points of the data points, which has the  direction accuracy $0.987$ compared with the approximate true velocity embeddings;  the bottom figure shows the grid plot of the velocity embeddings output by scVeloEmbeddings on the UMAP map points of the data points, which has the direction accuracy $0.967$.  Zoom in for details.  }
  \label{fig:approximate-embeddings-umap-grid}
  \end{figure}

\begin{figure}
  \centering
\includegraphics[width=150mm]{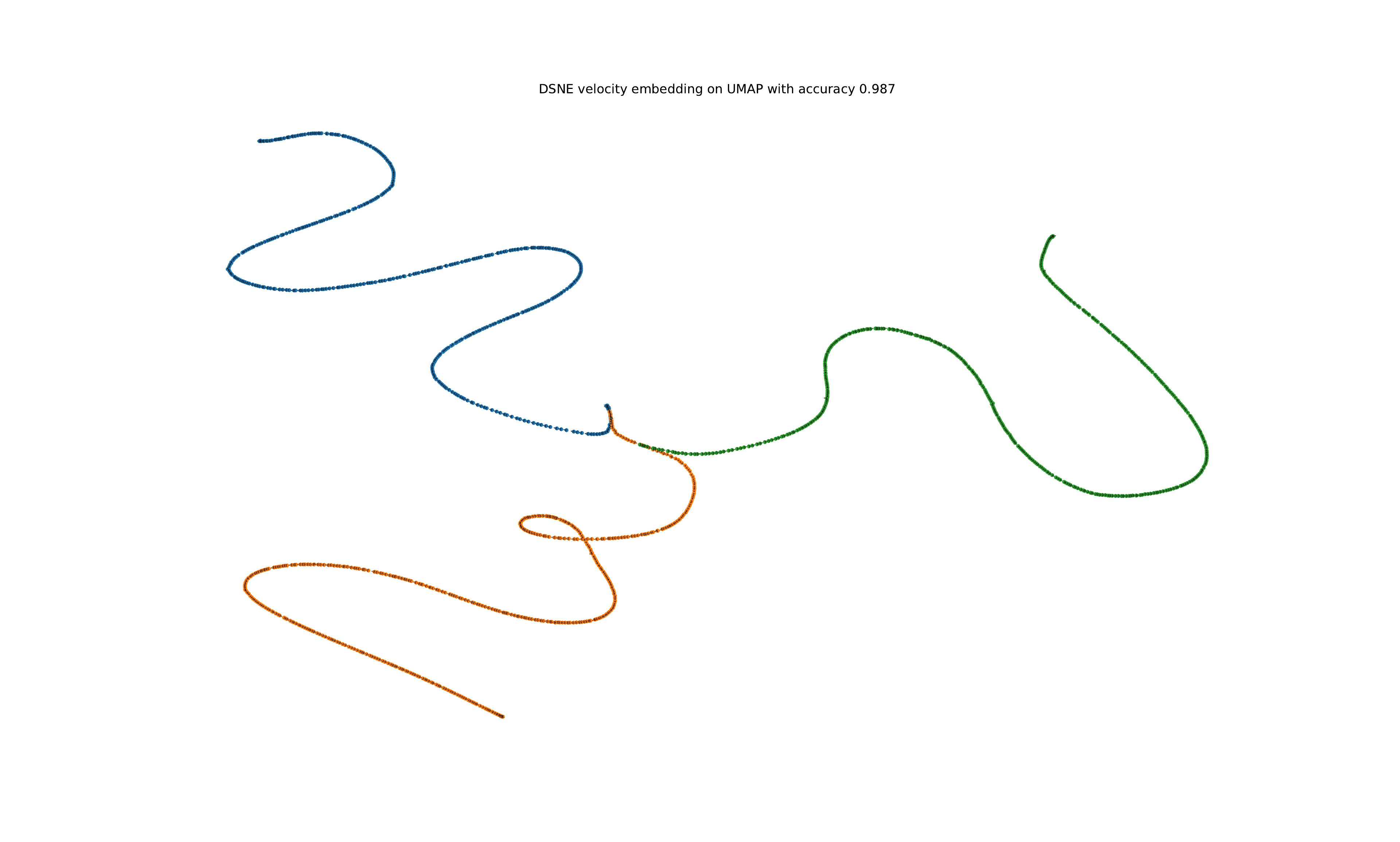}
\includegraphics[width=150mm]{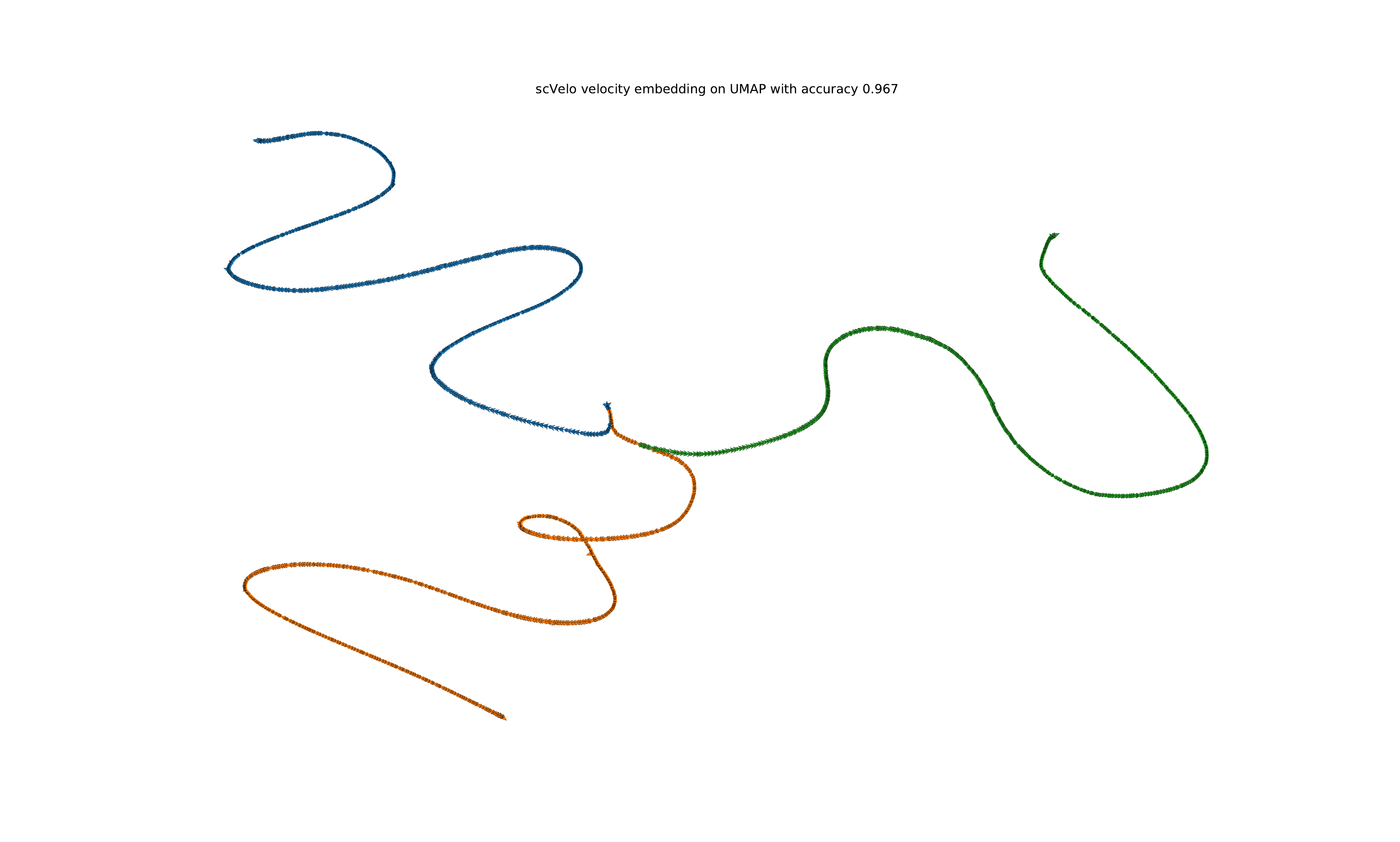}
  \caption{The arrow plot of  the simulated data based on exact data points and velocities  with $N=1500$, $D=300$  and $d=2$. The top figure shows the arrow plot of  the velocity embeddings output by  DSNE on the UMAP map points of the data points, which has the  direction accuracy $0.987$ compared with the approximate true velocity embeddings;  the bottom figure shows the arrow plot of the velocity embeddings output by scVeloEmbeddings on the UMAP map points of the data points, which has the direction accuracy $0.967$.   Zoom in for details. }
  \label{fig:approximate-embeddings-umap-arrow}
  \end{figure}

\begin{figure}
  \centering
\includegraphics[width=150mm]{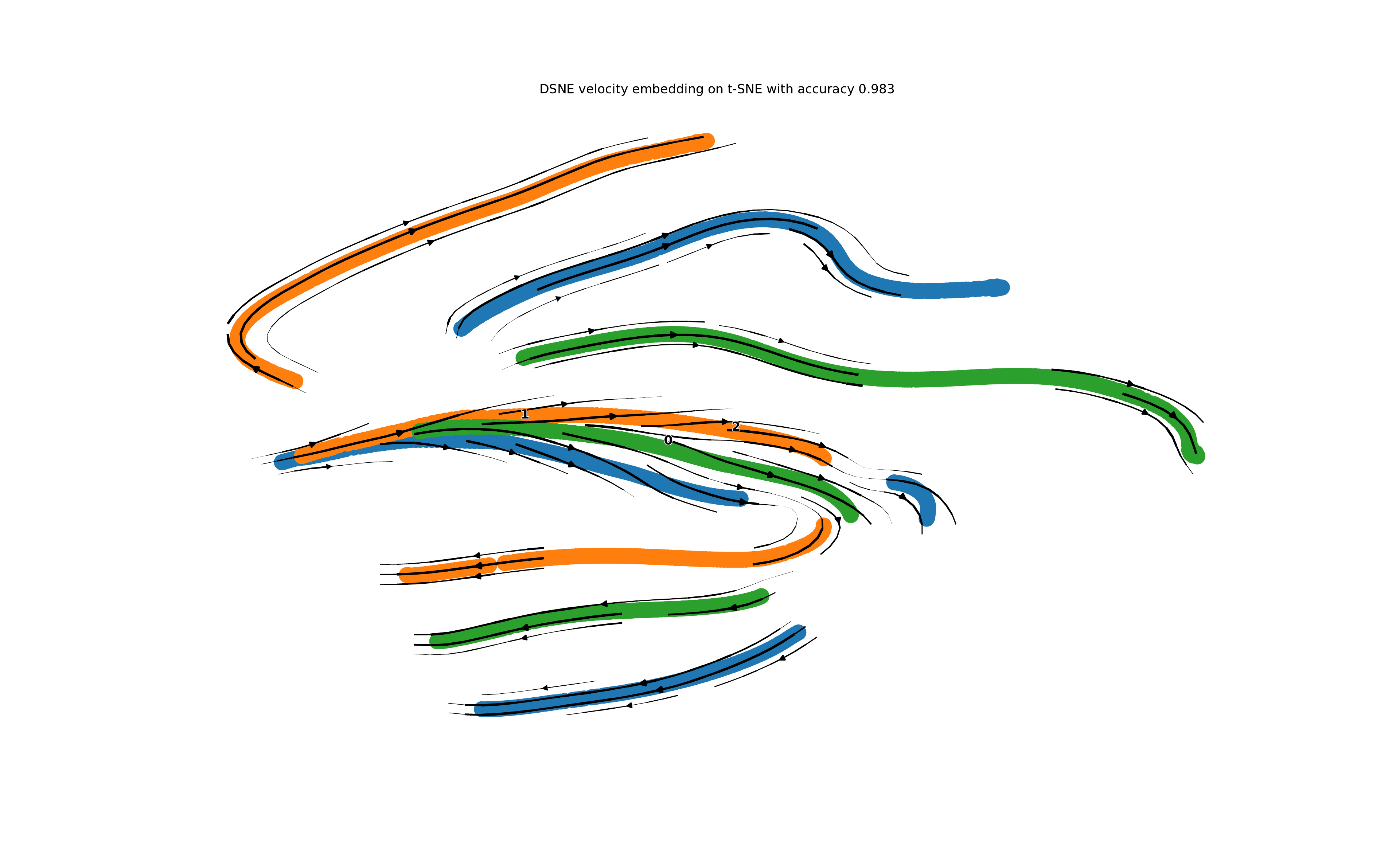}
\includegraphics[width=150mm]{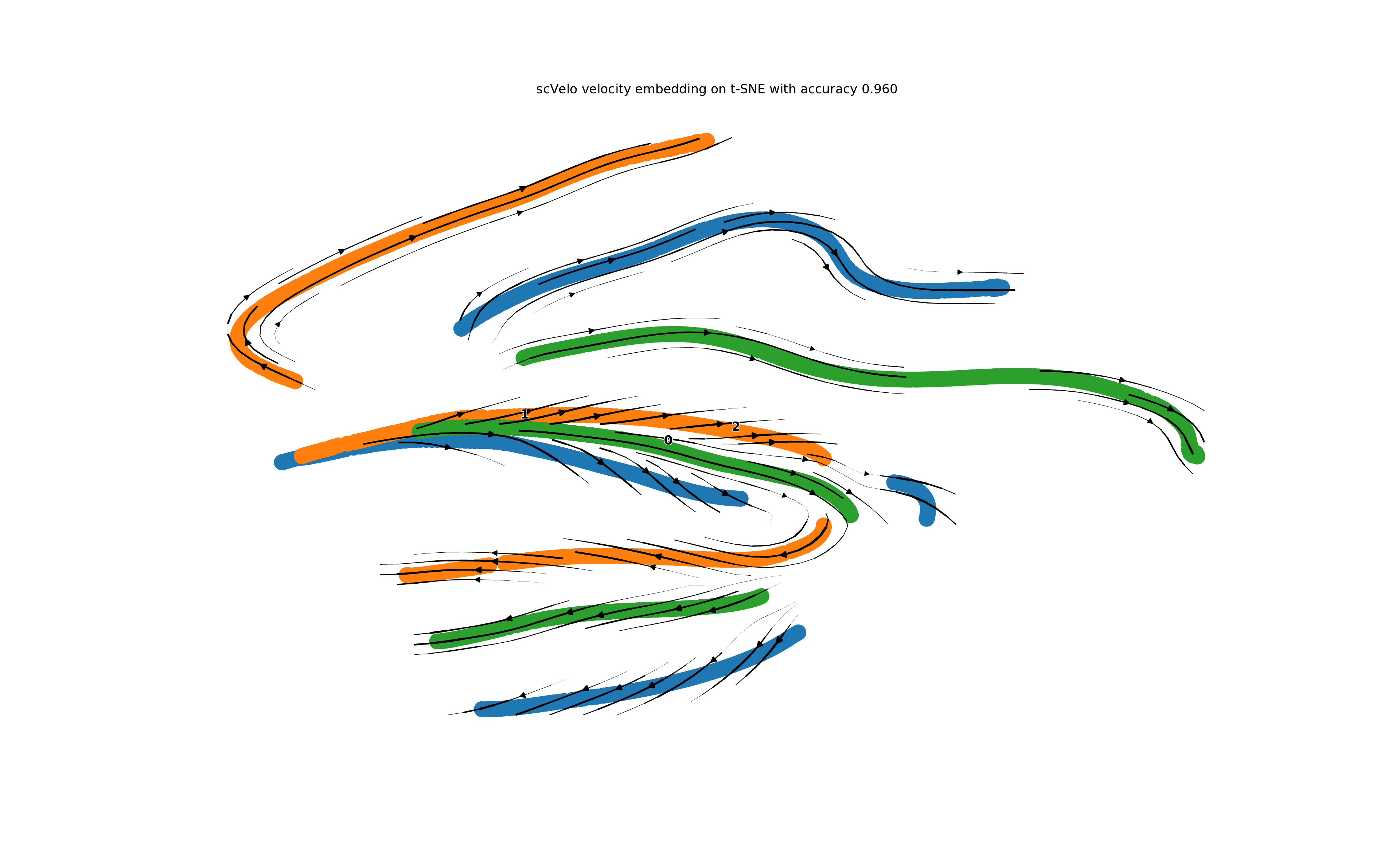}
  \caption{The stream plot of  the simulated data based on exact data points and velocities  with $N=1500$, $D=300$  and $d=2$. The top figure shows the stream plot of  the velocity embeddings output by  DSNE on the t-SNE map points of the data points, which has the  direction accuracy $0.983$ compared with the approximate  true velocity embeddings;  the bottom figure shows the stream plot of the velocity embeddings output by scVeloEmbeddings on the  t-SNE map points of the data points, which has the direction accuracy $0.960$.  Zoom in for details.  }
  \label{fig:approximate-embeddings-tsne-stream}
  \end{figure}

\begin{figure}
  \centering
\includegraphics[width=150mm]{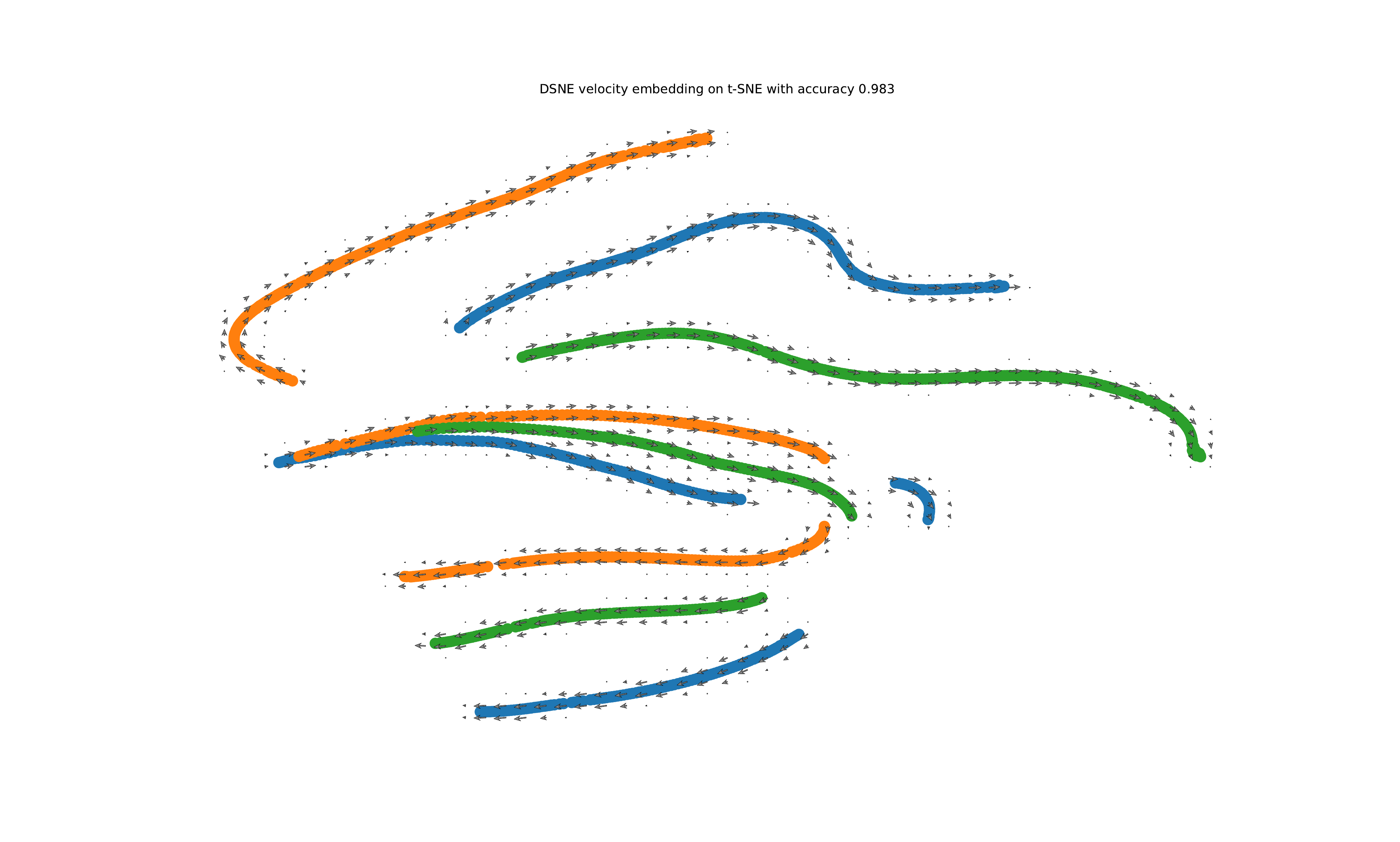}
\includegraphics[width=150mm]{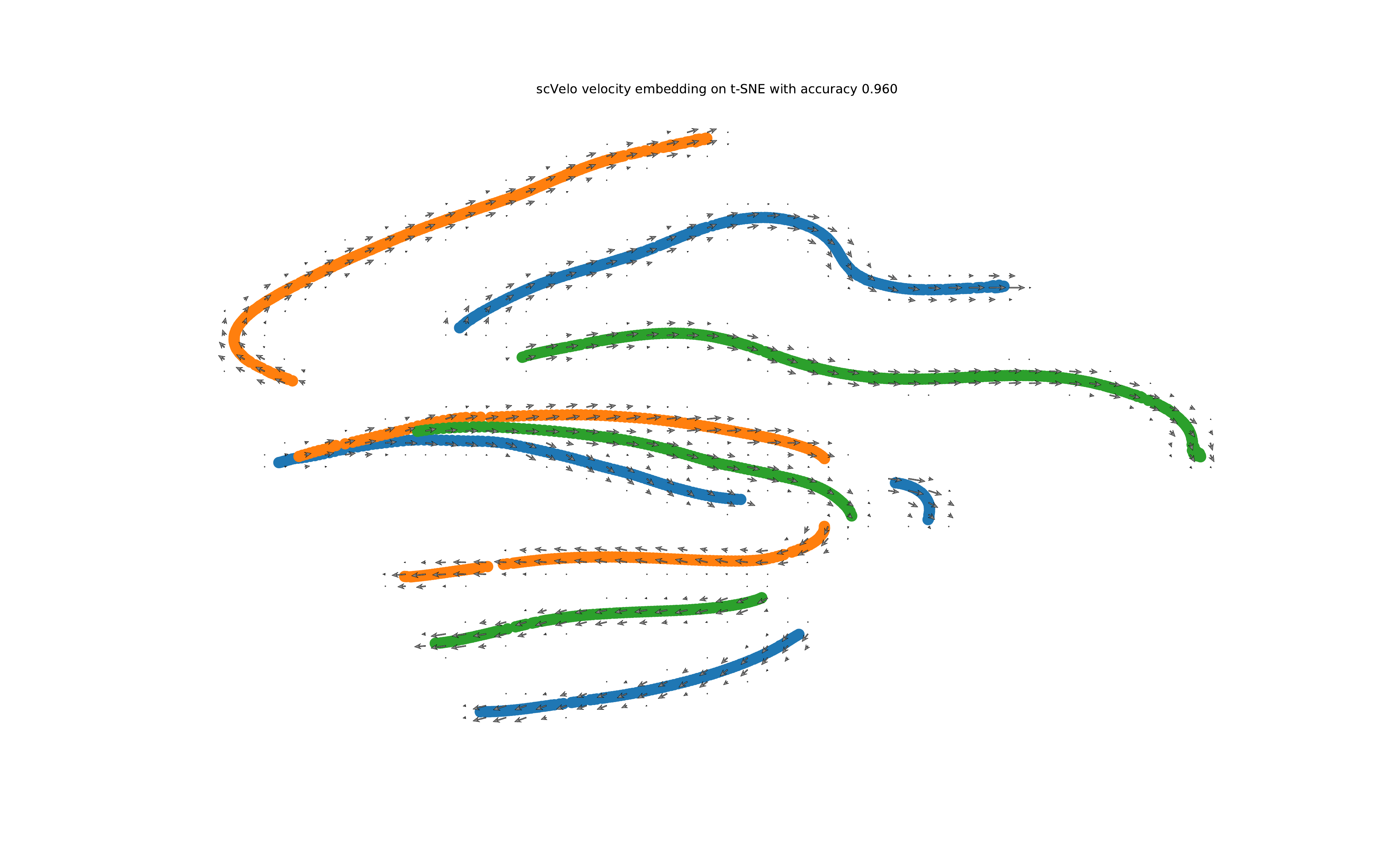}
  \caption{The grid plot of  the simulated data based on exact data points and velocities  with $N=1500$, $D=300$  and $d=2$. The top figure shows the grid plot of  the velocity embeddings output by  DSNE on the t-SNE map points of the data points, which has the  direction accuracy $0.983$ compared with the approximate true  velocity embeddings;  the bottom figure shows the grid plot of the velocity embeddings output by scVeloEmbeddings on the  t-SNE map points of the data points, which has the direction accuracy $0.960$.  Zoom in for details.  }
  \label{fig:approximate-embeddings-tsne-grid}
  \end{figure}

\begin{figure}
  \centering
\includegraphics[width=150mm]{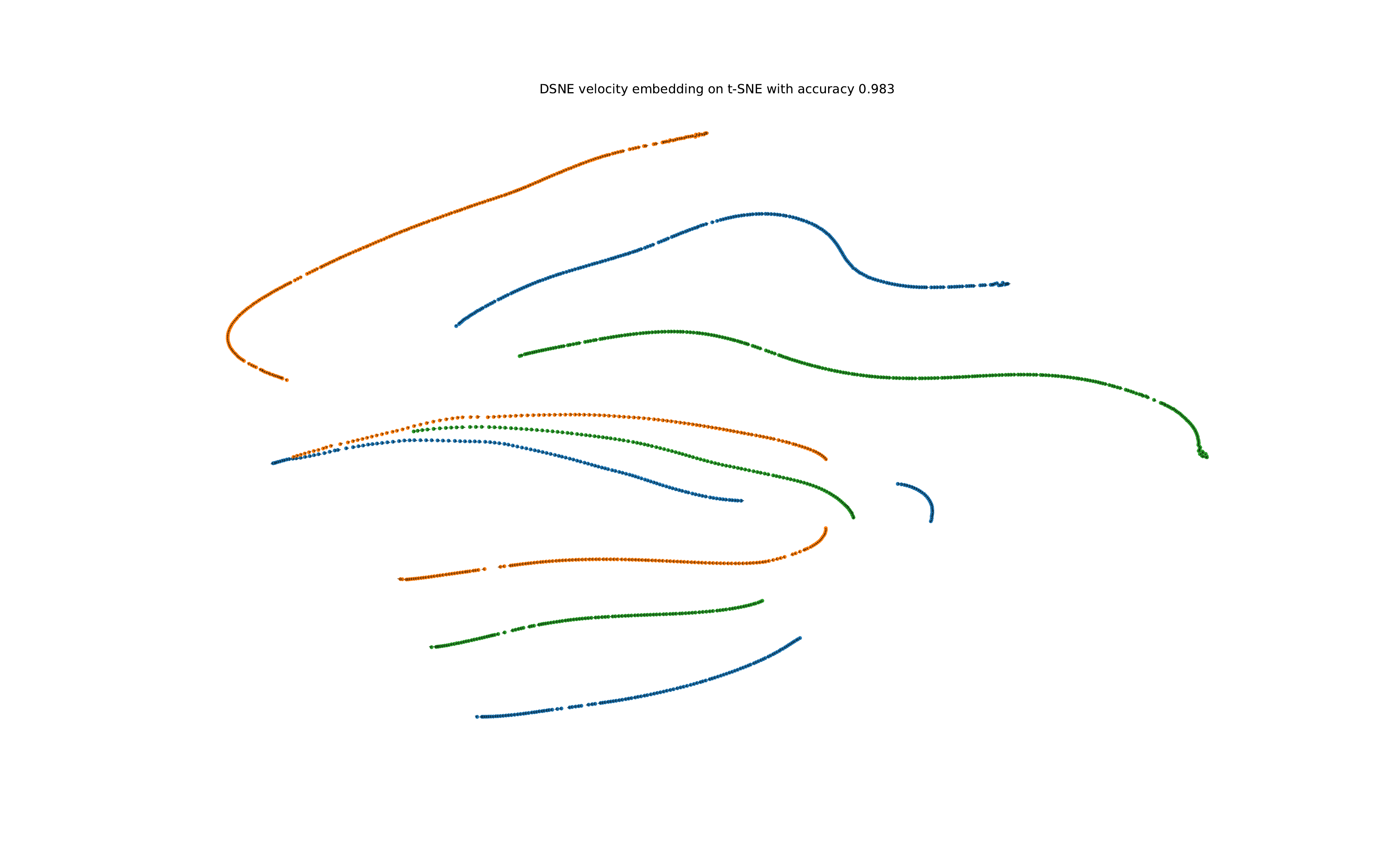}
\includegraphics[width=150mm]{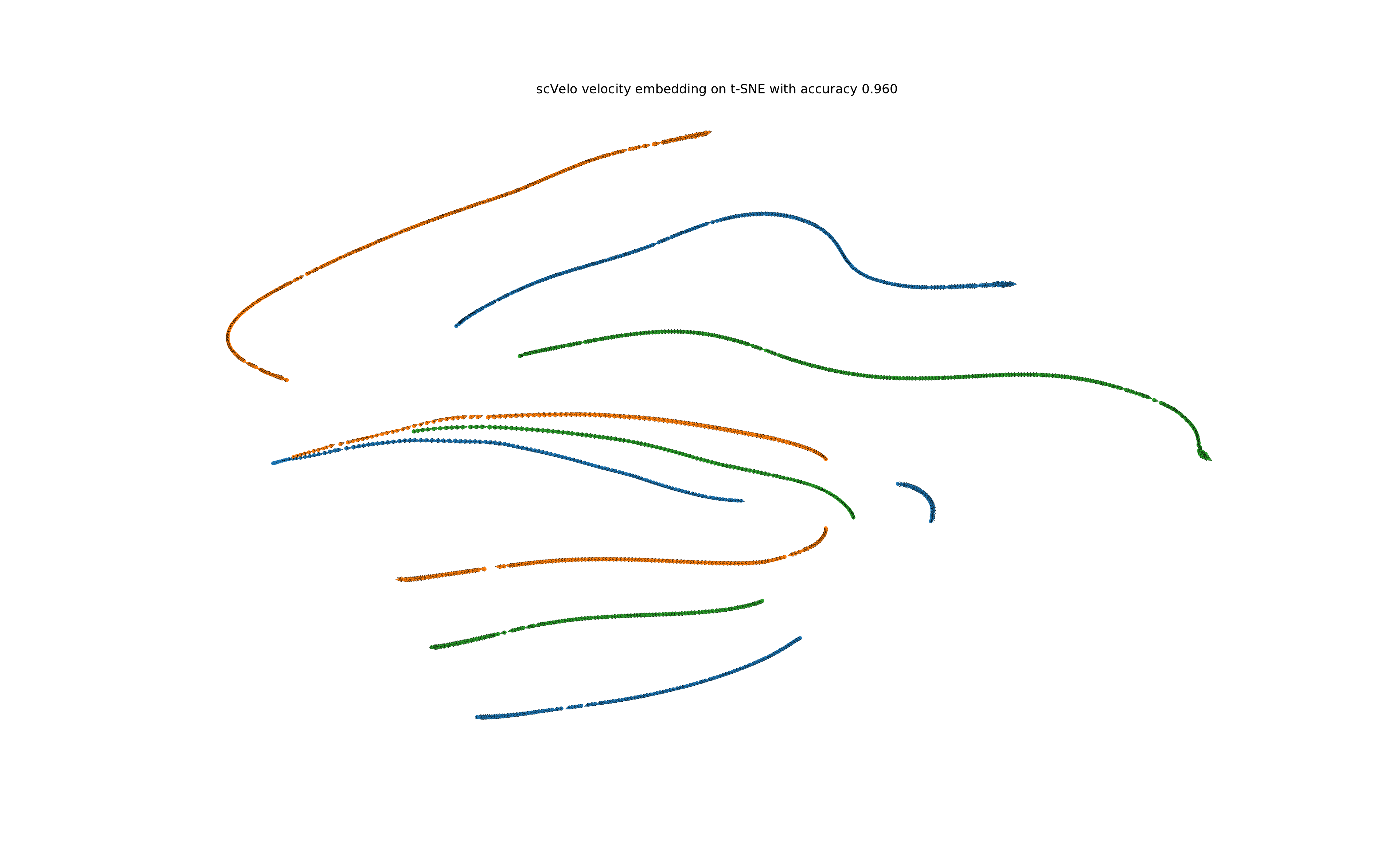}
  \caption{The arrow plot of  the simulated data based on exact data points and velocities  with $N=1500$, $D=300$  and $d=2$. The top figure shows the arrow plot of  the velocity embeddings output by  DSNE on the t-SNE map points of the data points, which has the  direction accuracy $0.983$ compared with the approximate true  velocity embeddings;  the bottom figure shows the arrow plot of the velocity embeddings output by scVeloEmbeddings on the  t-SNE map points of the data points, which has the direction accuracy $0.960$.  Zoom in for details.  }
  \label{fig:approximate-embeddings-tsne-arrow}
  \end{figure}

To get a visual feeling on the velocity embeddings, we plot the stream, grid, arrow plot of the results of DSNE and scVeloEmbedding on the UMAP map points ( see Figure~\ref{fig:approximate-embeddings-umap-stream}, 
 Figure~\ref{fig:approximate-embeddings-umap-grid},  Figure~\ref{fig:approximate-embeddings-umap-arrow})  and on the t-SNE map points ( see Figure~\ref{fig:approximate-embeddings-tsne-stream}, 
 Figure~\ref{fig:approximate-embeddings-tsne-grid},  Figure~\ref{fig:approximate-embeddings-tsne-arrow}). From the UMAP stream plot (Figure~\ref{fig:approximate-embeddings-umap-stream}), we see that DSNE present a well stream line along the map points. while scVeloEmbedding present  over-smoothed stream lines along the map points. Also we note that UMAP is a better representation of the global structure than t-SNE map points, since each color line is a swig line in the high dimension space, t-SNE map points break the line into small pieces in the low-dimensional space, while UMAP keeps the continuous line for each color.

%

\subsubsection{Simulation with known low dimensional velocity} 
\label{sec:exact-simulation}
To get the exact quantitive measure how the DSNE and scVeloEmbedding behave, we generate the simulation data  which begin with   velocity embeddings on the low dimensional space,  and then moving along the velocity embedding with one time step one-by-one from three start points to get the map points. Then we  linear project the map points and their velocities to the high dimensional space. By this way, we have the true velocity embeddings and map points of the corresponding high dimensional data points and velocities.  We  compare the velocity embeddings $W$  with the true velocity embeddings $W_{true}$ by the cosine distances, i.e, we define the accuracy of velocity embeddings $W$  with  the true velocity embeddings $W_{true}$ by 
\begin{equation}
\label{eq:accuracy-exact}
accu :=  \frac 1 N \sum \langle \frac { w_i } { ||w_i||} ,   \frac { w_{true, i} } { ||w_{true, i}||} \rangle 
\end{equation}
where $accu \in [-1, 1]$, the perfect accuracy is $1$ with all the velocity embeddings direction correct,  $\frac{ w_i } { ||w_i||} =   \frac { w_{true, i} } { ||w_{true, i}}, \; i=1, \ldots, N$; the lowest accuracy is $-1$ with  all the velocity embeddings direction are the opposite of the true  velocity embeddings direction, $\frac { w_i } { ||w_i||} =  - \frac { w_{true, i} } { ||w_{true, i}||}, \; i=1, \ldots, N$. 

The simulate data was generated similarly as above. 
\begin{enumerate}
\item  Generate the  low-dimensional velocity $W_{true} \in \mathbb R^{N \times D}$  by random sampling from the Normal distributions, $ W_{true, il}  \sim \mathcal N(0,36),  \; i=1 \ldots, N; \; l = 1, \ldots, d$. where we take $N = 3N_s$;  
\item  Choose there start points of map points $y_{start,1} = \mathbf 0$, $y_{start,2}  =  50* \mathbf 1$.
    $y_{start,3}  =160 * \mathbf 1$.  where $ \mathbf 0$ is the zeros vector with length $d$  and $\mathbf 1$ is the ones vector with length $d$. 
\item  Generate the map points  by moving  from the three starting points along  with the velocity embedding $w_i$ one by one, i.e. 
\begin{equation*}
\begin{array}{l}
	y_1 = y_{start, 1}\\
	y_{N_s+1} = y_{start, 2}\\
	y_{2N_s+1} = y_{start, 3}\\
	y_{i+1} = y_i + w_{true, i}, \; i = 1, \ldots, N_s-1 \\
	y_{N_s + i+1} = y_{N_s + i}  + w_{true, i}, \; i = 1, \ldots, N_s-1 \\
	y_{2N_s + i+1} = y_{2N_s + i}  + w_{true, i}, \; i = 1, \ldots, N_s-1   \\
\end{array}
\end{equation*}

\item Generate the projection matrix $U \in \mathbb R^{d \times D}$ by random sampling from the standard normal distributions, i.e. 
$U_{kl} \sim \mathcal N(0,1), \; k=1, \ldots, d; \; l= 1, \ldots, D$.

\item Projection the map points $Y$ and the  true velocity embeddings $W_{true}$ by the projection matrix $U$ to get the data points $X= YU$ and velocity matrix $V= W_{true} U$. 

\end{enumerate}

We run the DSNE and scVeloEmbedding  to learn the velocity embeddings and finally compare the accuracy defined  in equation (\ref{eq:accuracy-exact}) to see how good the two algorithms behave. To compare the performance, we run  simulation data with same $N,D$ $10$ times,  run the DSNE and scveloEmbedding algorithm on the same simulated data each time. For DSNE, we select the parameter $K=16$, $perplexity = 6$ for all settings. We run scVeloEmbedding with default parameters in scVelo package ($K=100$). 
The mean with the standard deviation of the  accuracies of the $10$ times for different $N$ and $D$ are presented in Table~\ref{tab:accuracy-exact}. It obviously that  DSNE do a better work than scVeloEmbedding on all the test settings.  

\begin{table}
 \caption{Accuracy of Direction of Velocity Embedding on the Simulation Data with Exact Map Points and Velocity Embeddings }
  \centering
  \begin{tabular}{lll}
    \toprule
    Name     & Accuracy mean (std) of DSNE &  Accuracy mean (std) of scVeloEmbedding  \\
    \midrule
     $N=150, \; D=30$ & $0.980~(	0.011)	$ & $0.914~(	0.010)$	\\
     $N=1500, \; D=10$ & $0.985 ~(		0.007)	$		&  $0.965 ~(		0.003)$ \\
   $N=1500, \; D=300$ & $0.994 ~(		0.002)	$		&  $0.966 ~(		0.007)$ \\
   $N=15000, \; D=300$  & $0.995 ~(		0.001)$ 		&  $0.983~(0.003)$   \\
    \bottomrule
  \end{tabular}
  \label{tab:accuracy-exact}
\end{table}

\begin{figure}
  \centering
\includegraphics[width=150mm]{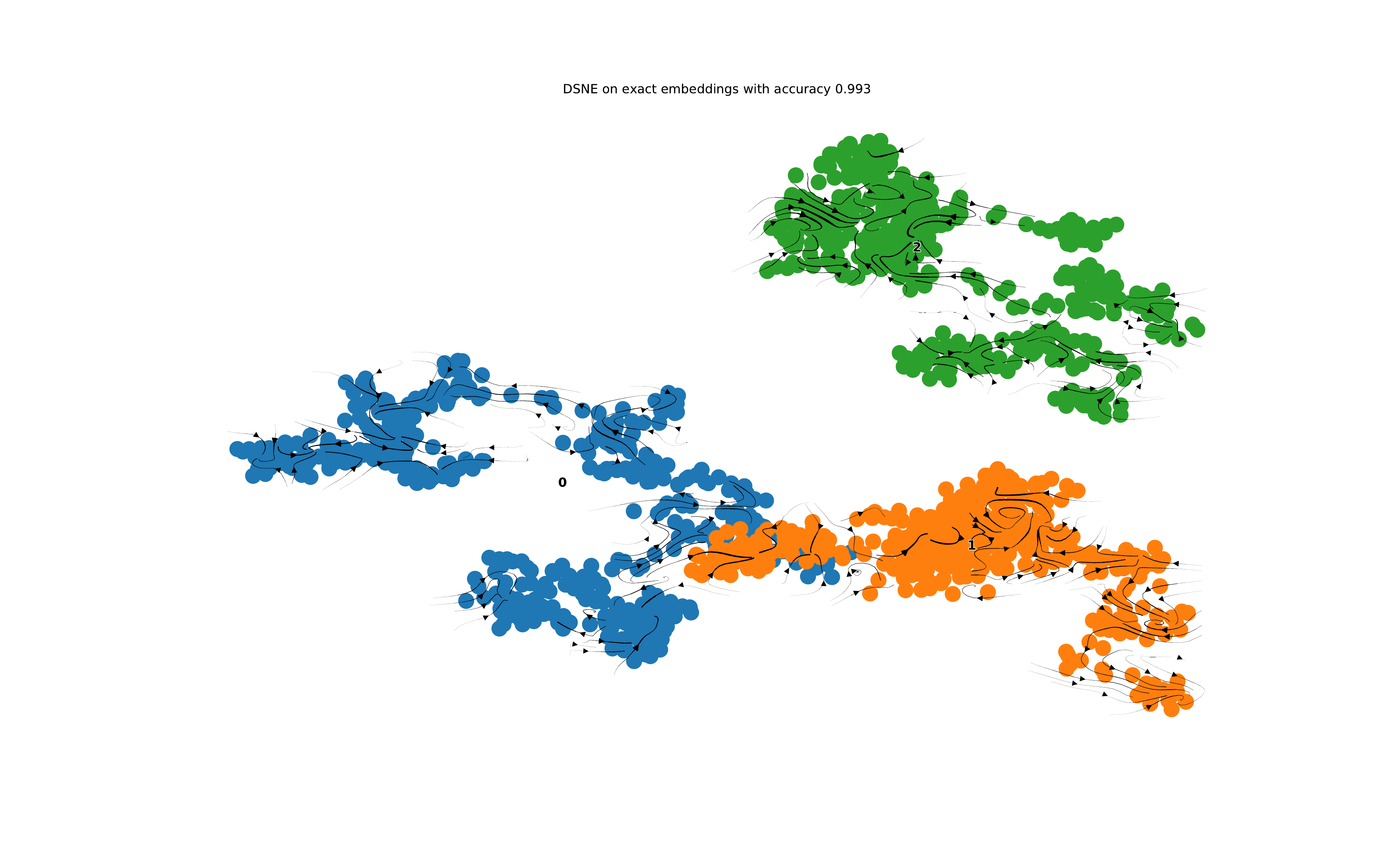}
\includegraphics[width=150mm]{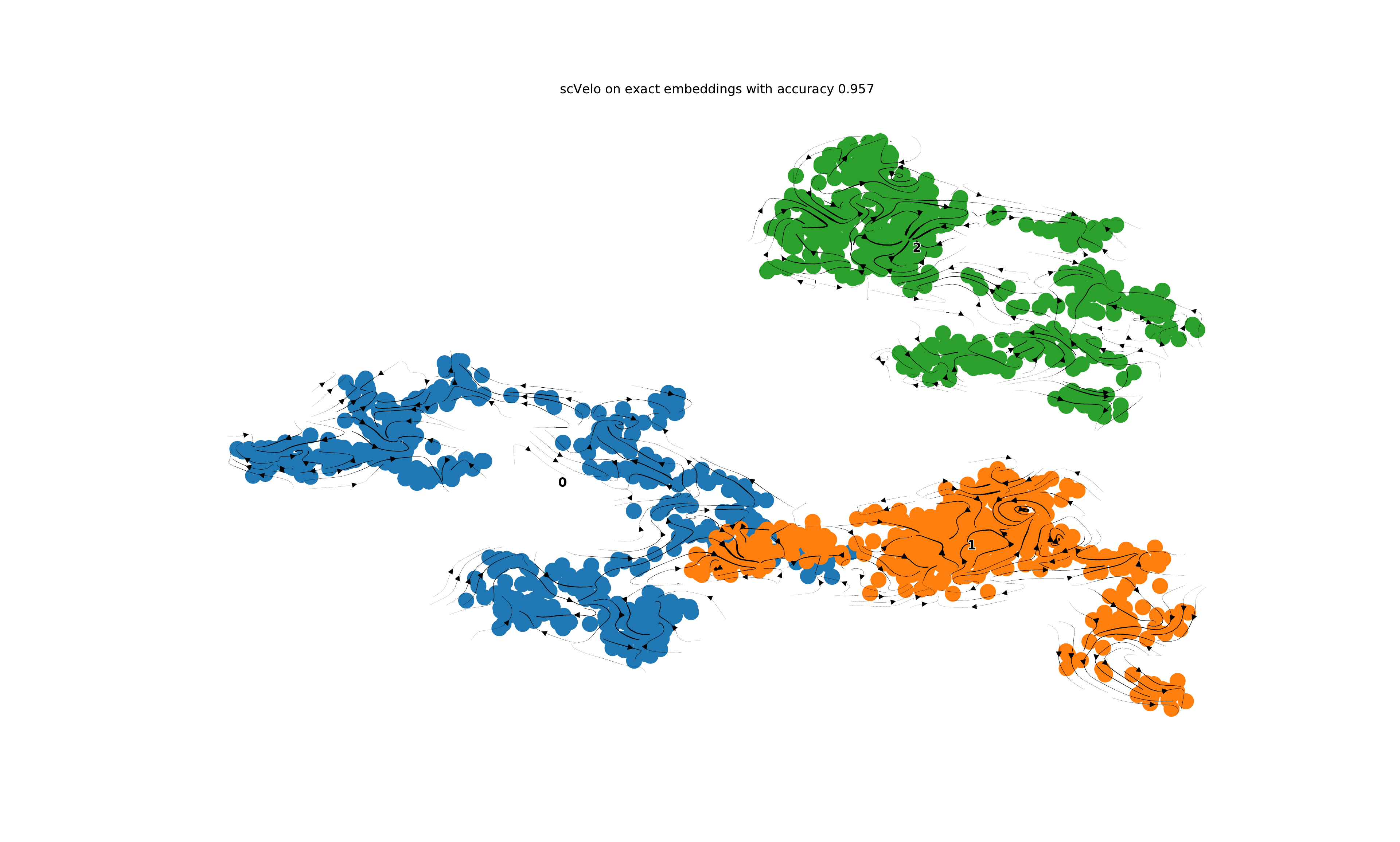}
  \caption{The stream plot of the simulated data based on exact map points and velocity embeddings with $N=1500$, $D=300$  and $d=2$. The top figure shows the stream plot of  the velocity embeddings output by  DSNE, which has the   direction accuracy $0.993$ compared with the true velocity embeddings;  the bottom figure shows the stream plot of the velocity embeddings output by scVeloEmbedding, which has the direction accuracy $0.957$.  Zoom in for details.  }
  \label{fig:exact-embeddings-stream}
  \end{figure}

\begin{figure}
  \centering
\includegraphics[width=150mm]{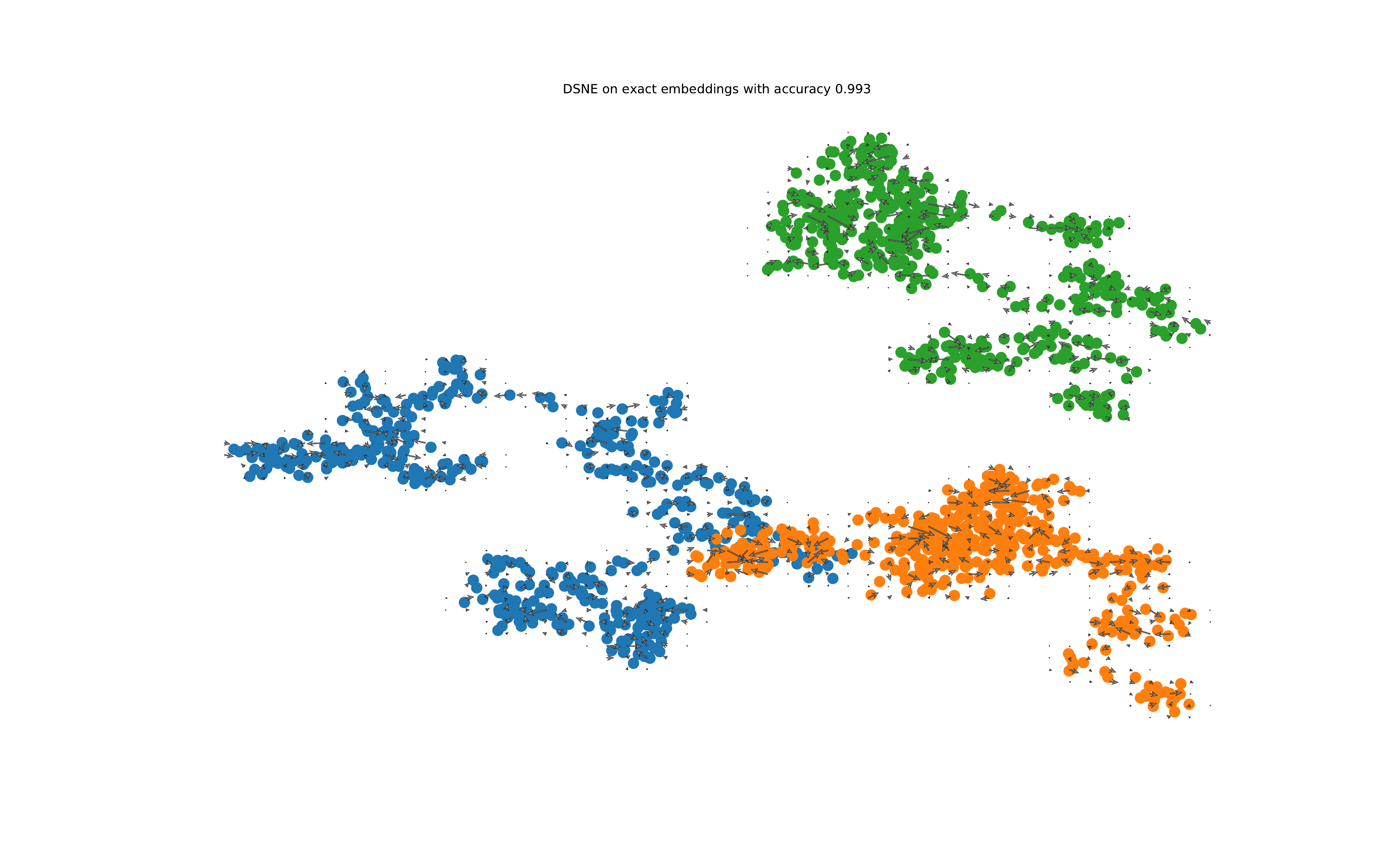}
\includegraphics[width=150mm]{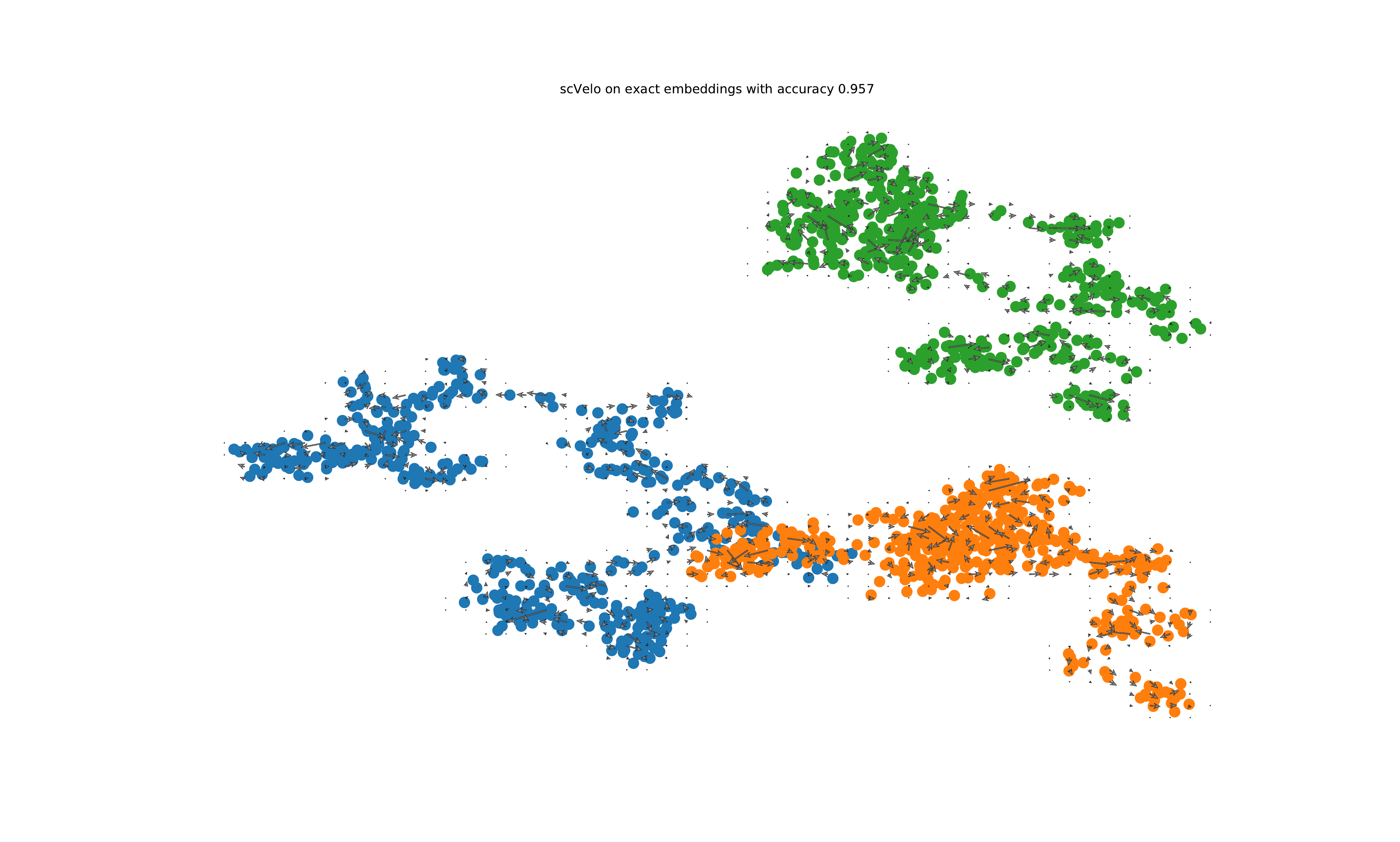}
  \caption{The grid plot  of  the simulated data based on exact map points and velocity embeddings with $N=1500$, $D=300$  and $d=2$. The top figure shows the stream plot of  the velocity embeddings output by  DSNE, which has the   direction accuracy $0.993$ compared with the true velocity embeddings;  the bottom figure shows the stream plot of the velocity embeddings output by scVeloEmbeddings, which has the direction accuracy $0.957$. Zoom in for details. }
  \label{fig:exact-embeddings-grid}
  \end{figure}

\begin{figure}
  \centering
\includegraphics[width=150mm]{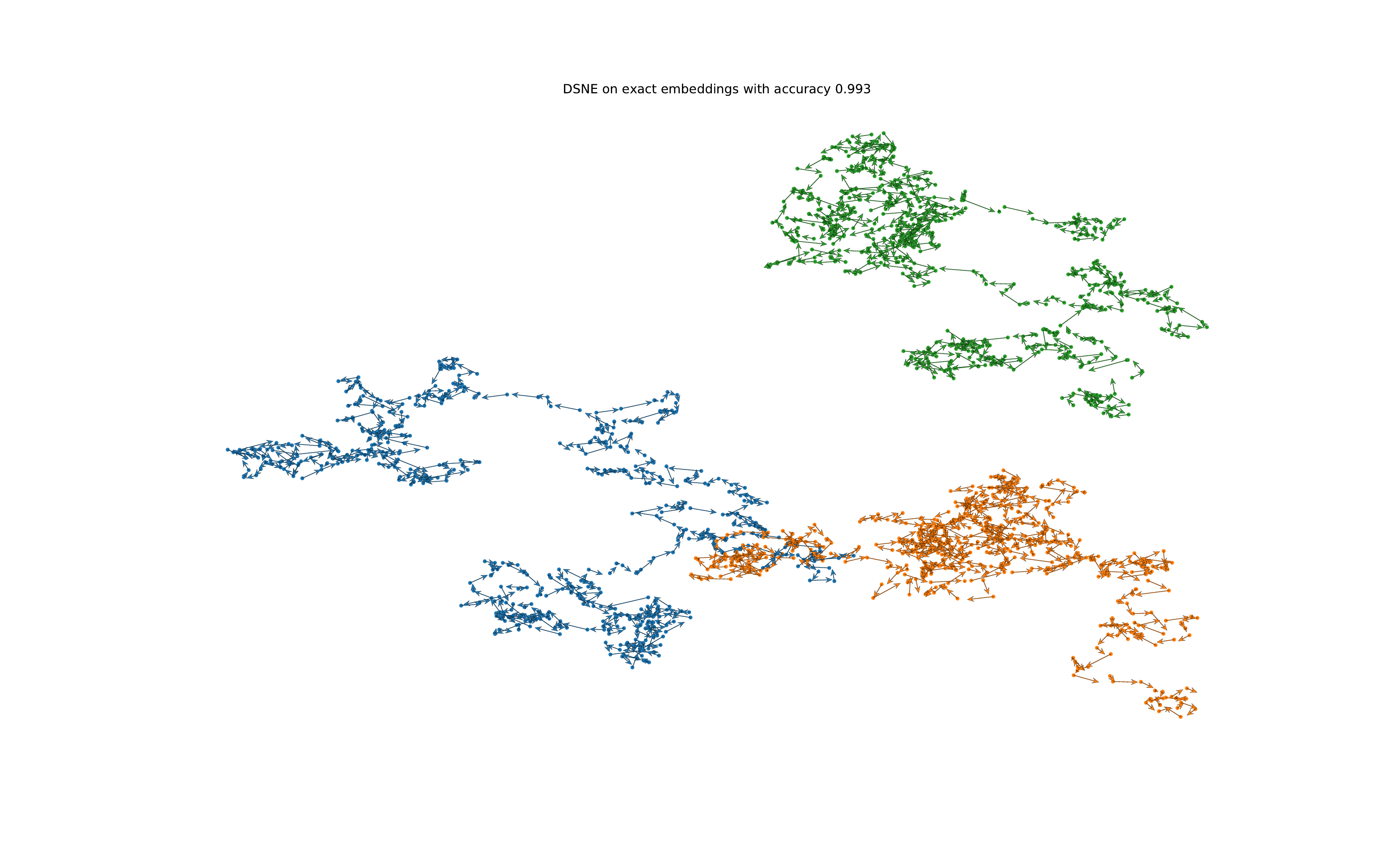}
\includegraphics[width=150mm]{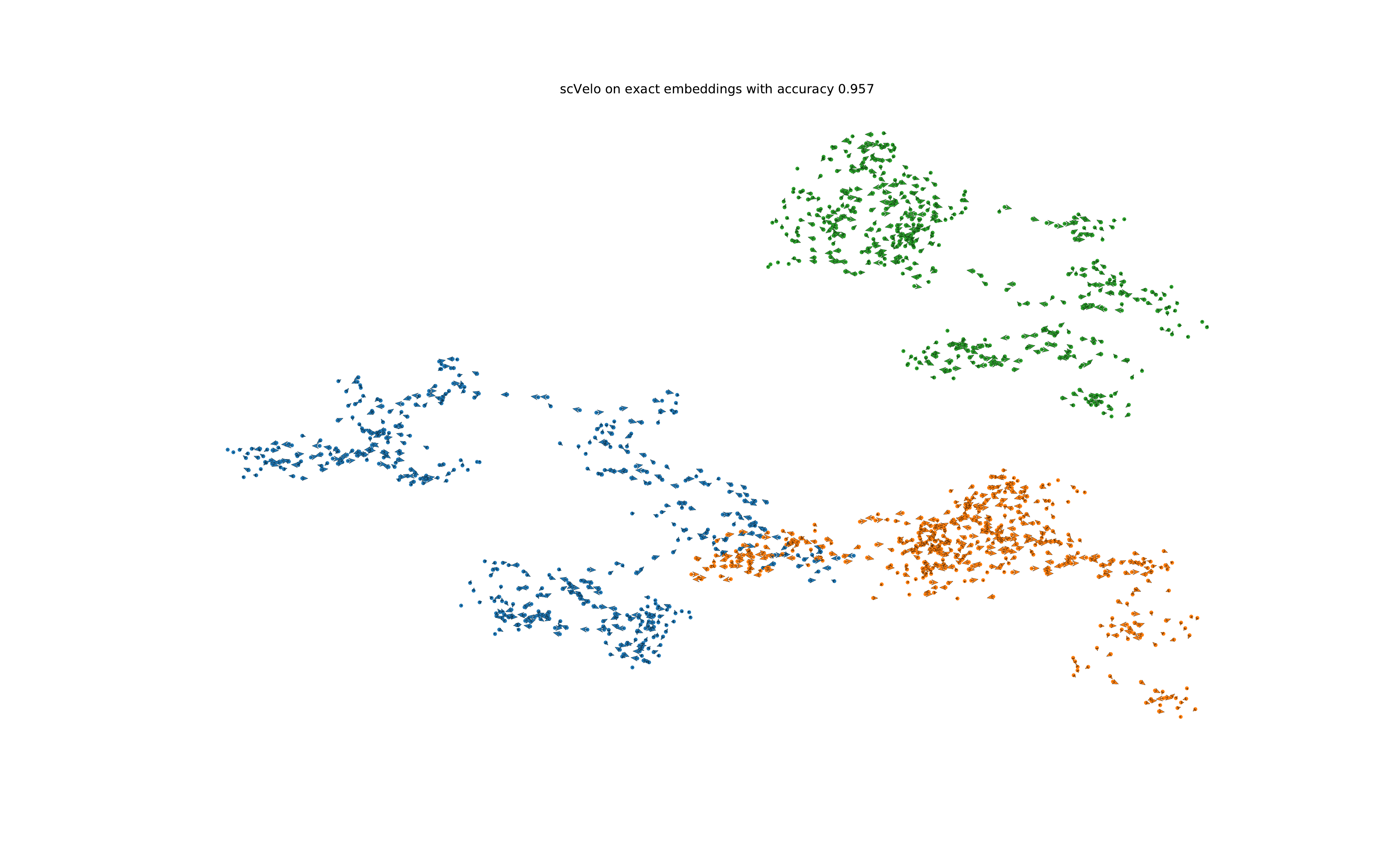}
  \caption{The arrow plot  of  the simulated data based on exact map points and velocity embeddings with $N=1500$, $D=300$  and $d=2$. The top figure shows the stream plot of  the velocity embeddings output by  DSNE, which has the   direction accuracy $0.993$ compared with the true velocity embeddings;  the bottom figure shows the stream plot of the velocity embeddings output by scVeloEmbeddings, which has the direction accuracy $0.957$.  Zoom in for details.  }
  \label{fig:exact-embeddings-arrow}
  \end{figure}

To get a feel about the velocity embedding, we plot the stream, grid, arrow  picture  in Figure~\ref{fig:exact-embeddings-stream}, Figure~\ref{fig:exact-embeddings-grid}, Figure~\ref{fig:exact-embeddings-arrow}, respectively. From the arrow picture ( Figure~\ref{fig:exact-embeddings-arrow} ), we found that arrow length of DSNE  was better presented than scVeloEmbedding's, this verifies the effectiveness of the approximate formula (\ref{eq:w-with-norm}).

\subsection{scRNA-seq data: Endocrine Pancreas }
The cell  differentiation and embryo development  is the fundamental problems in biology. RNA velocity techniques greatly aid to make a visually view how the cell trajectory presented on the low dimensional space.  Here, we use the Pancreas data which was analyzed in \cite{dynamical-RNA-velocity} to compare DSNE with the scVeloEmbedding.

We run DSNE with parameters $K=16$, $perplexity=3$ and run scVeloEmbedding with the default parameters, which is based on the notebooks from \url{https://github.com/theislab/scvelo_notebooks/Pancreas.ipynb}.  For the low dimensional map points, we use the UMAP map points and VeloViz(\cite{VeloViz}) map points which was based the tutorial \url{https://github.com/JEFworks-Lab/veloviz/vignettes/pancreas.Rmd}. 

\begin{figure}
  \centering
  \includegraphics[width=150mm]{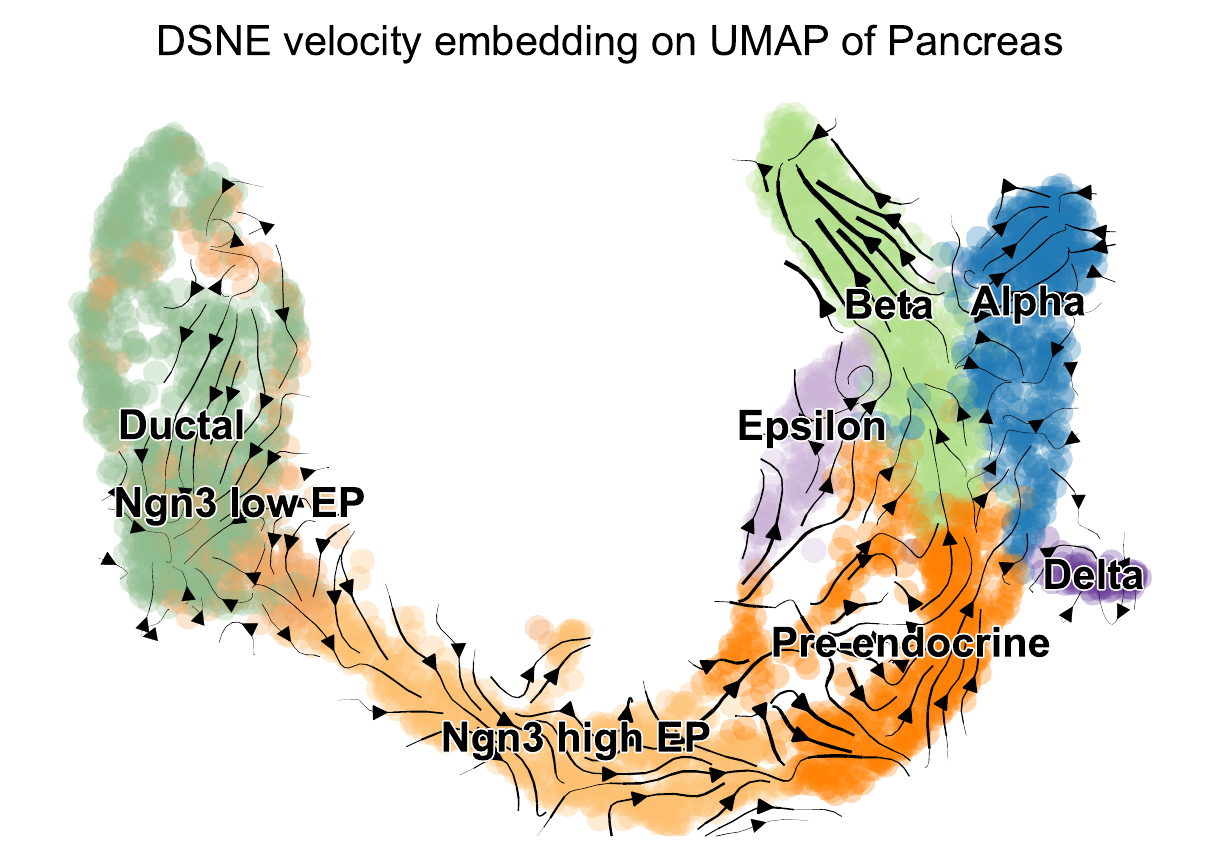}
\includegraphics[width=150mm]{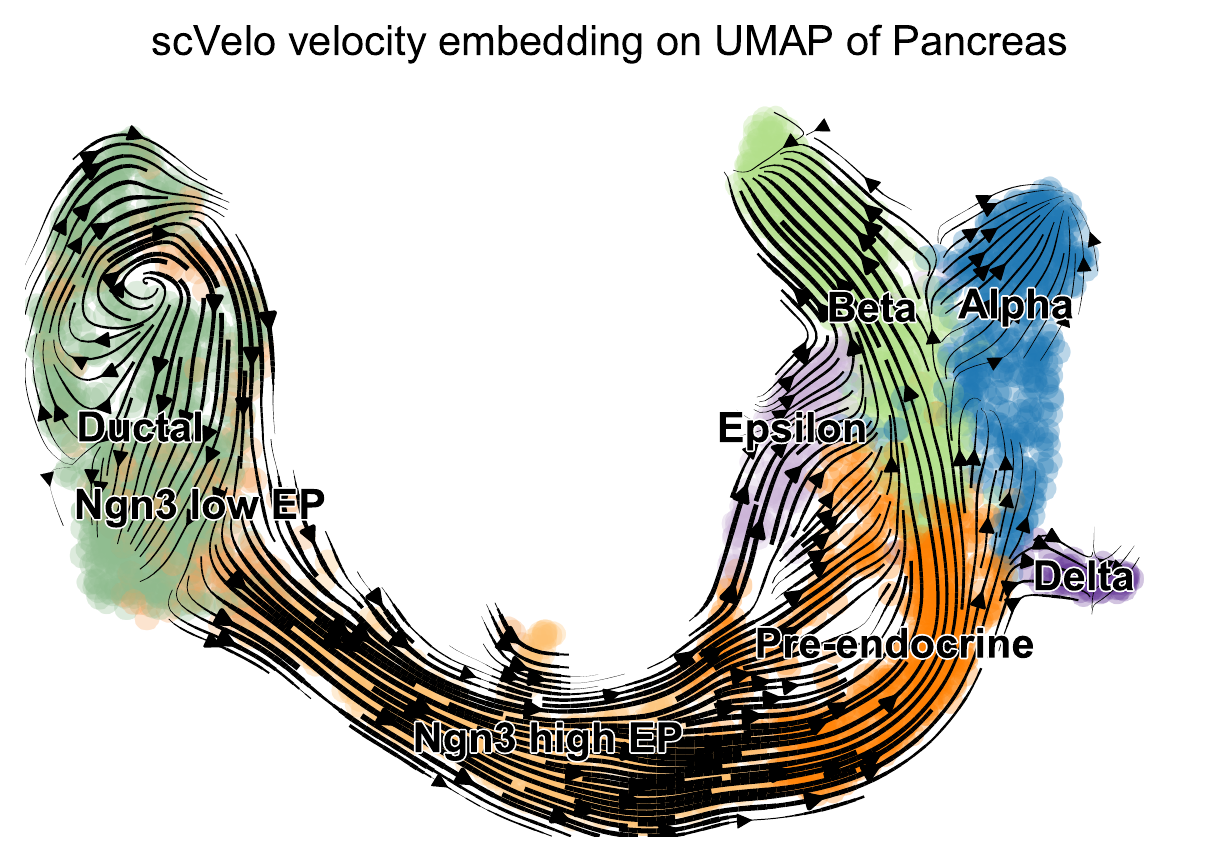}
  \caption{The stream plot of the velocity embeddings on the UMAP of the pancreas data. The top figure shows the results of DSNE   and  the bottom figure shows the results of scVeloEmbedding.  Zoom in for details. }
  \label{fig:pancreas-umap-stream}
  \end{figure}

\begin{figure}
  \centering
  \includegraphics[width=150mm]{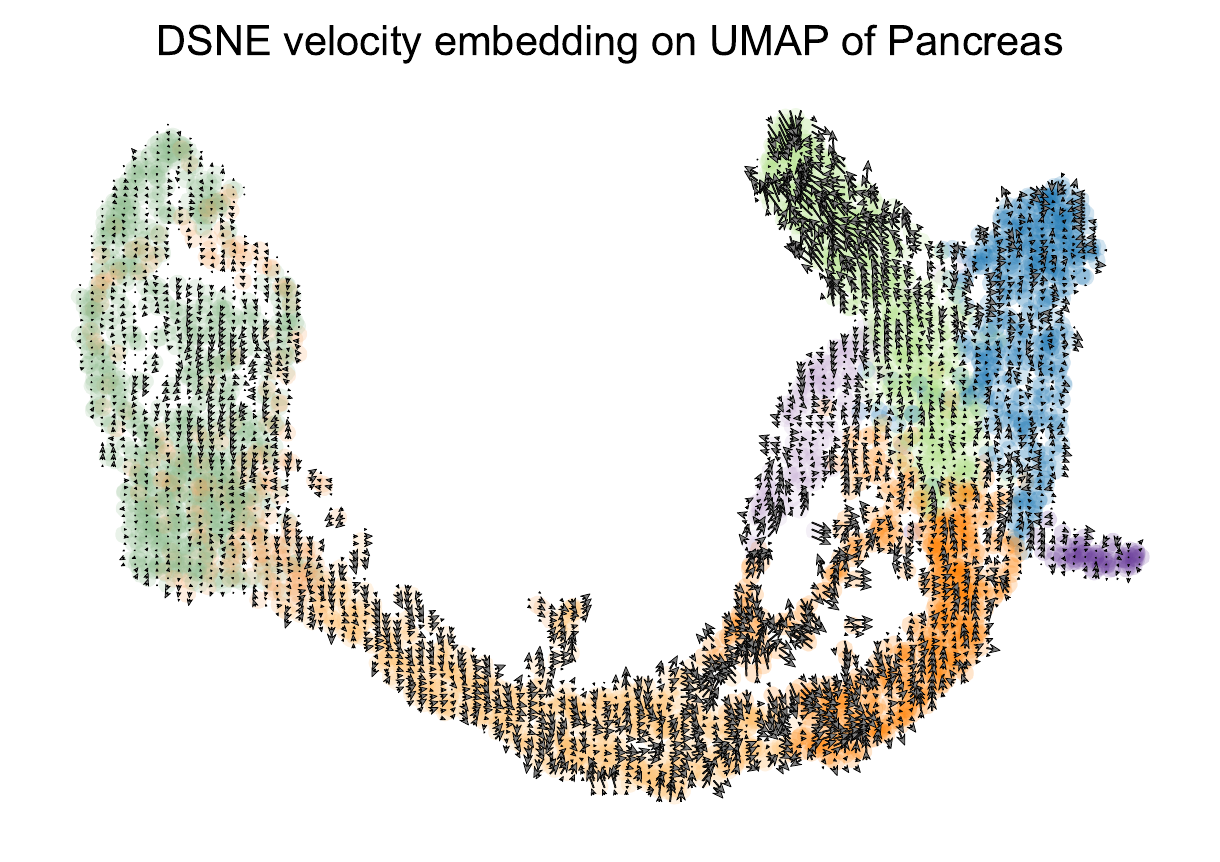}
\includegraphics[width=150mm]{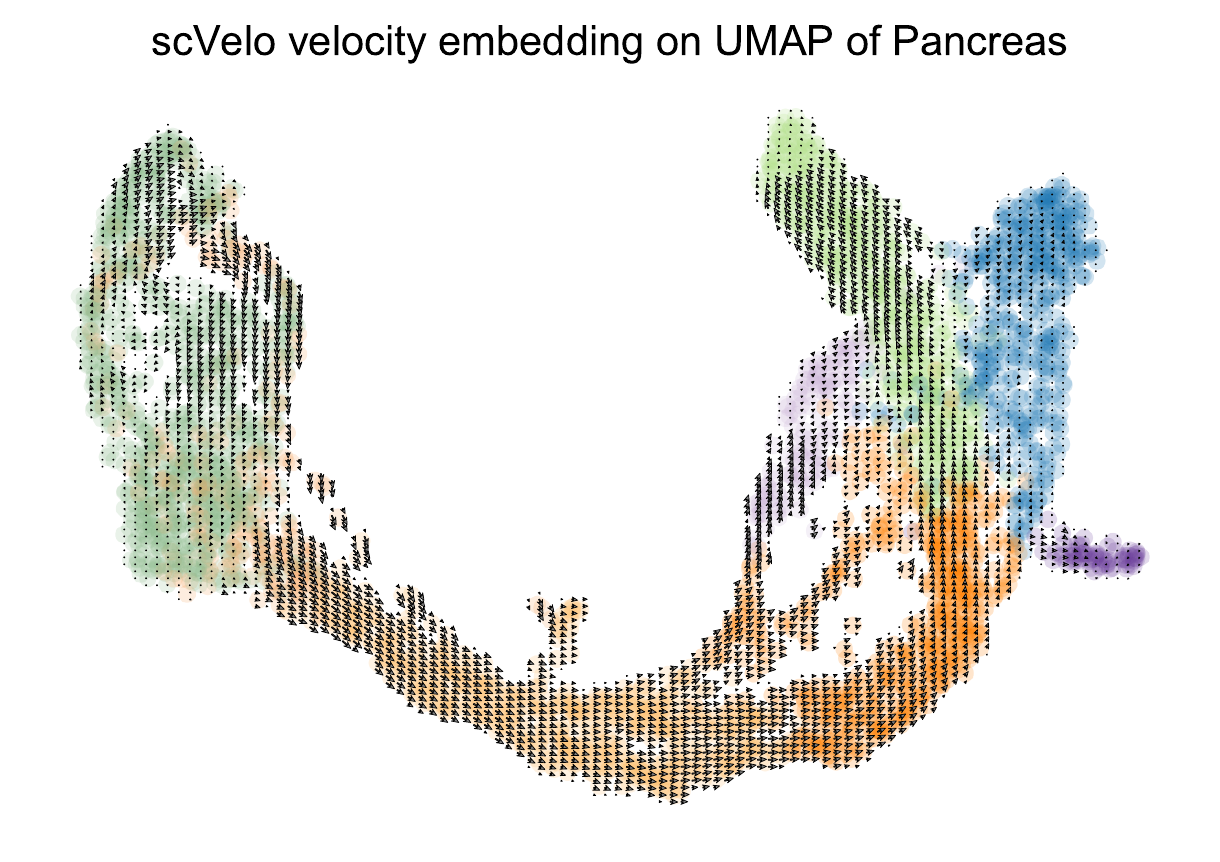}
  \caption{The grid plot of the velocity embeddings on the UMAP of the pancreas data. The top figure shows the results of DSNE and  the bottom figure shows the results of scVeloEmbedding. Zoom in for details. }
  \label{fig:pancreas-umap-grid}
  \end{figure}

\begin{figure}
  \centering
  \includegraphics[width=150mm]{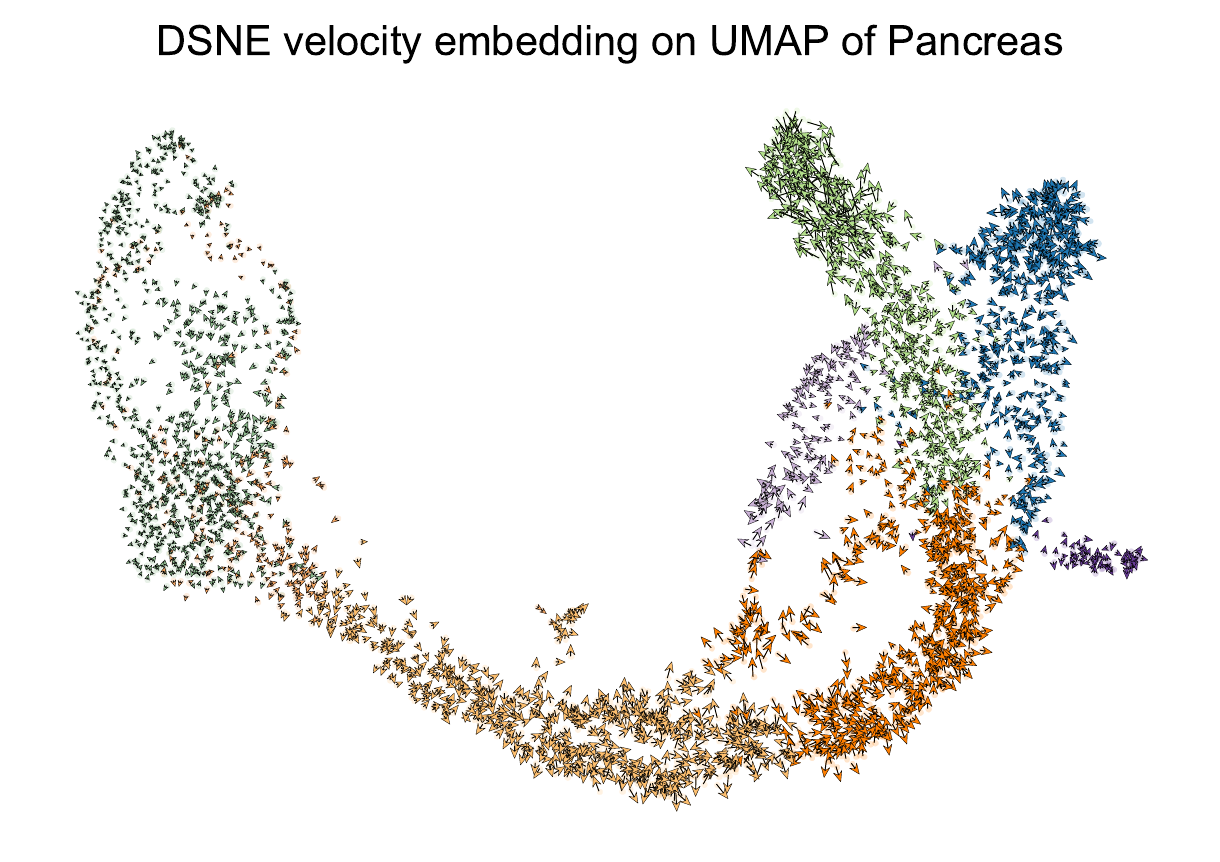}
\includegraphics[width=150mm]{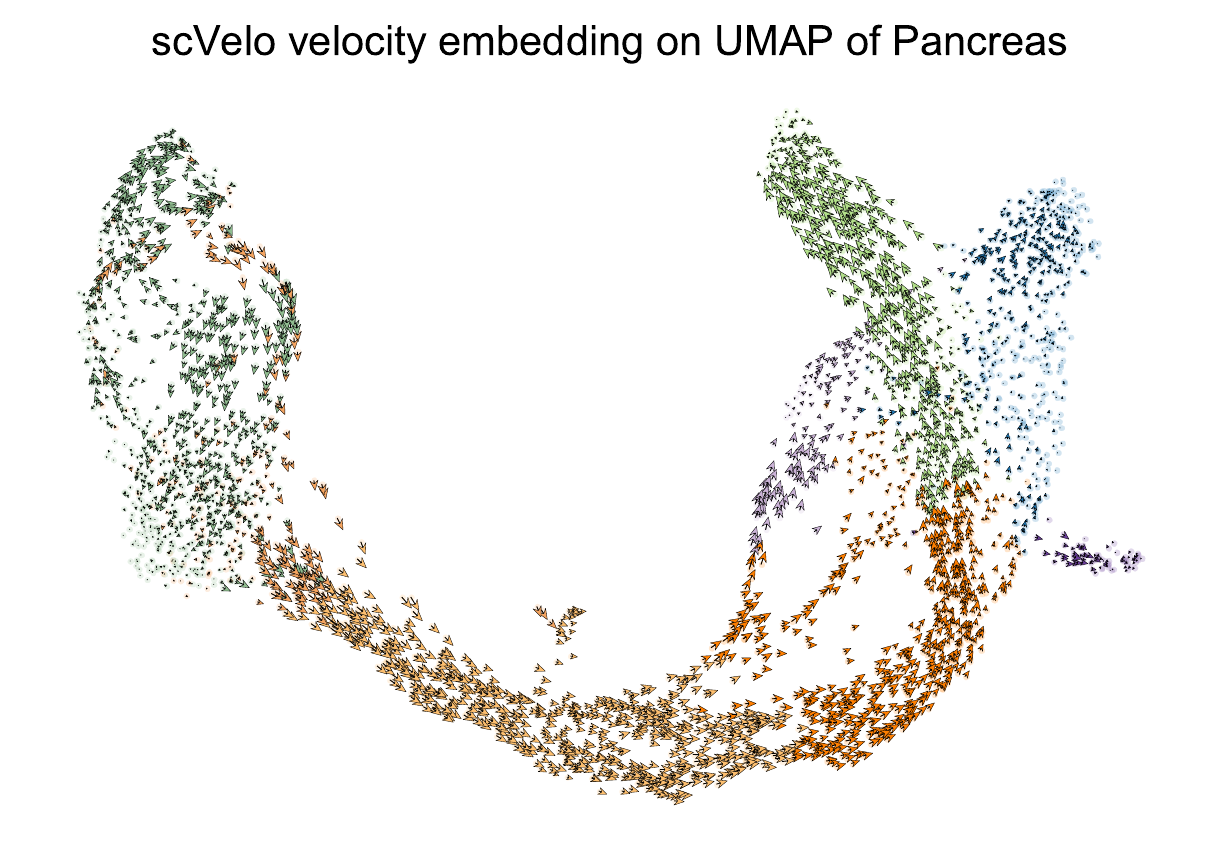}
  \caption{The arrow plot of the velocity embeddings on the UMAP of the pancreas data. The top figure shows the results of  DSNE and  the bottom figure shows the results of scVeloEmbedding. Zoom in for details. }
  \label{fig:pancreas-umap-arrow}
  \end{figure}

\begin{figure}
  \centering
  \includegraphics[width=150mm]{./fig/pancreas/scvelo_pancreas_dsne_veloviz_stream.pdf}
\includegraphics[width=150mm]{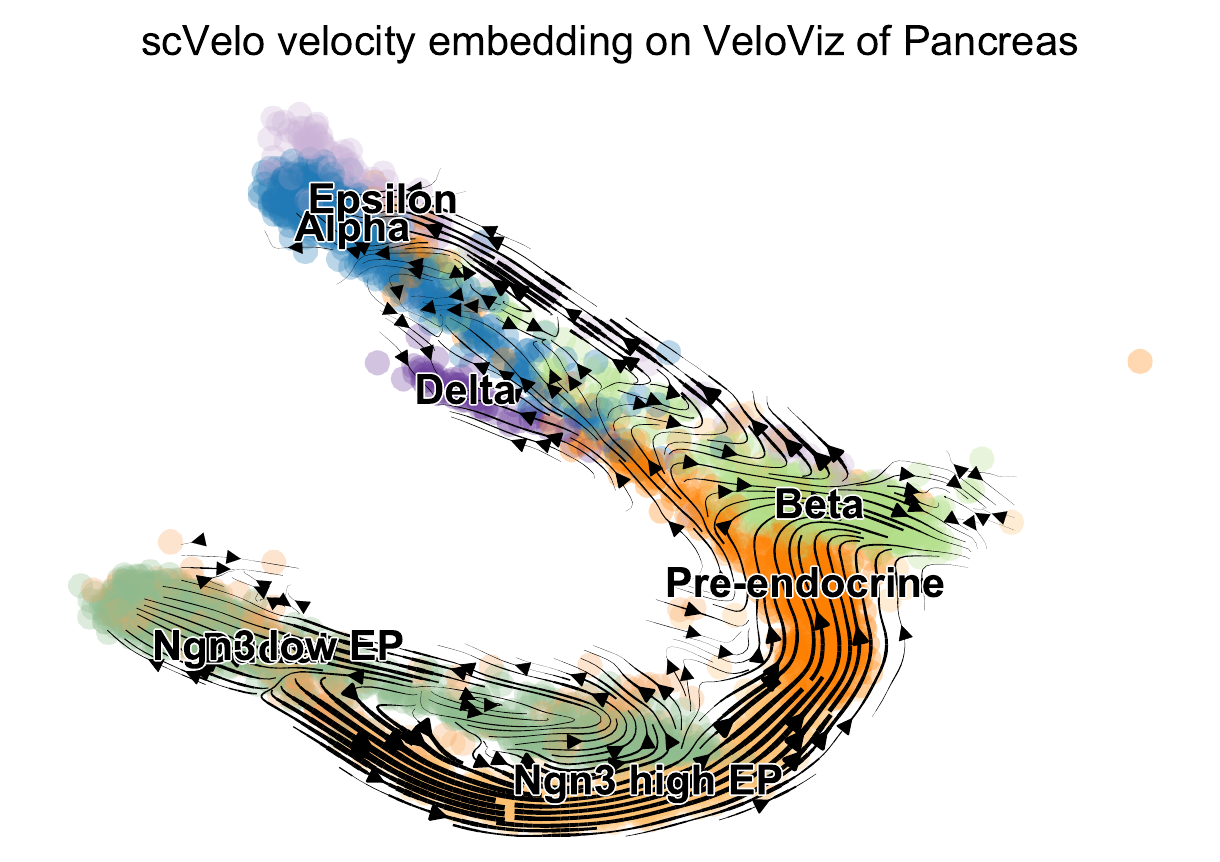}
  \caption{The  stream plot of the velocity embeddings on the VeloViz map points of the pancreas data. The top figure shows the results of DSNE and  the bottom figure shows the results of scVeloEmbedding. Zoom in for details. }
  \label{fig:pancreas-veloviz-stream}
  \end{figure}

\begin{figure}
  \centering
  \includegraphics[width=150mm]{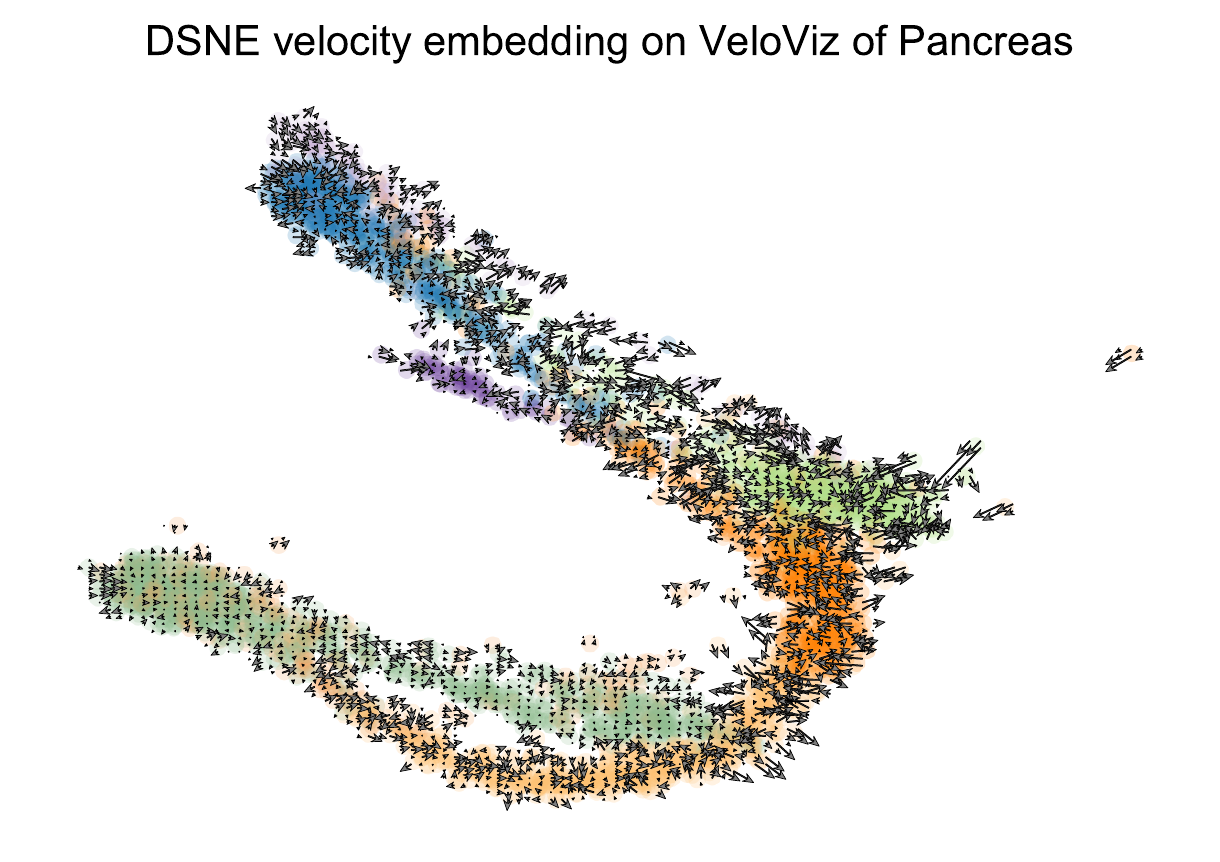}
\includegraphics[width=150mm]{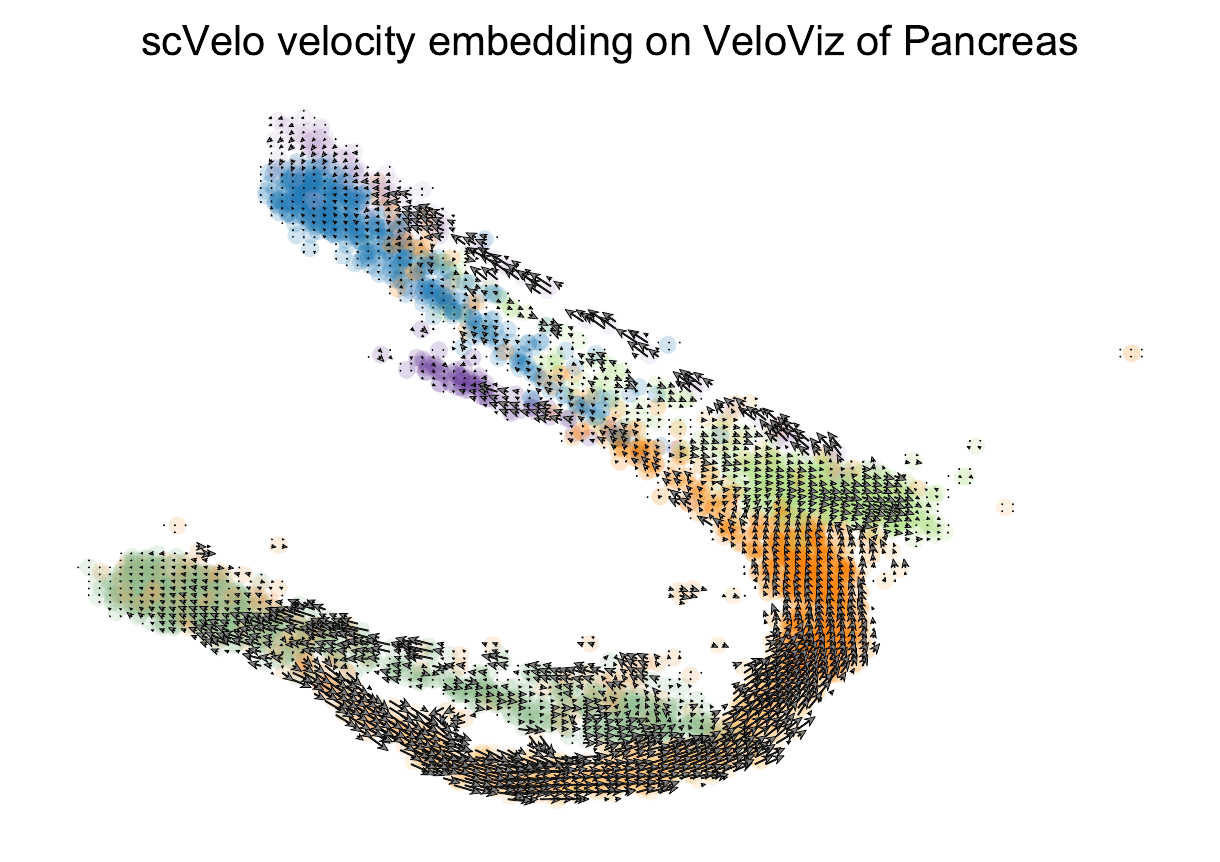}
  \caption{The grid plot of the velocity embeddings on the VeloViz map points of the pancreas data. The top figure shows the results of DSNE  and  the bottom figure shows the results of scVeloEmbedding. Zoom in for details. }
  \label{fig:pancreas-veloviz-grid}
\end{figure}

\begin{figure}
  \centering
  \includegraphics[width=150mm]{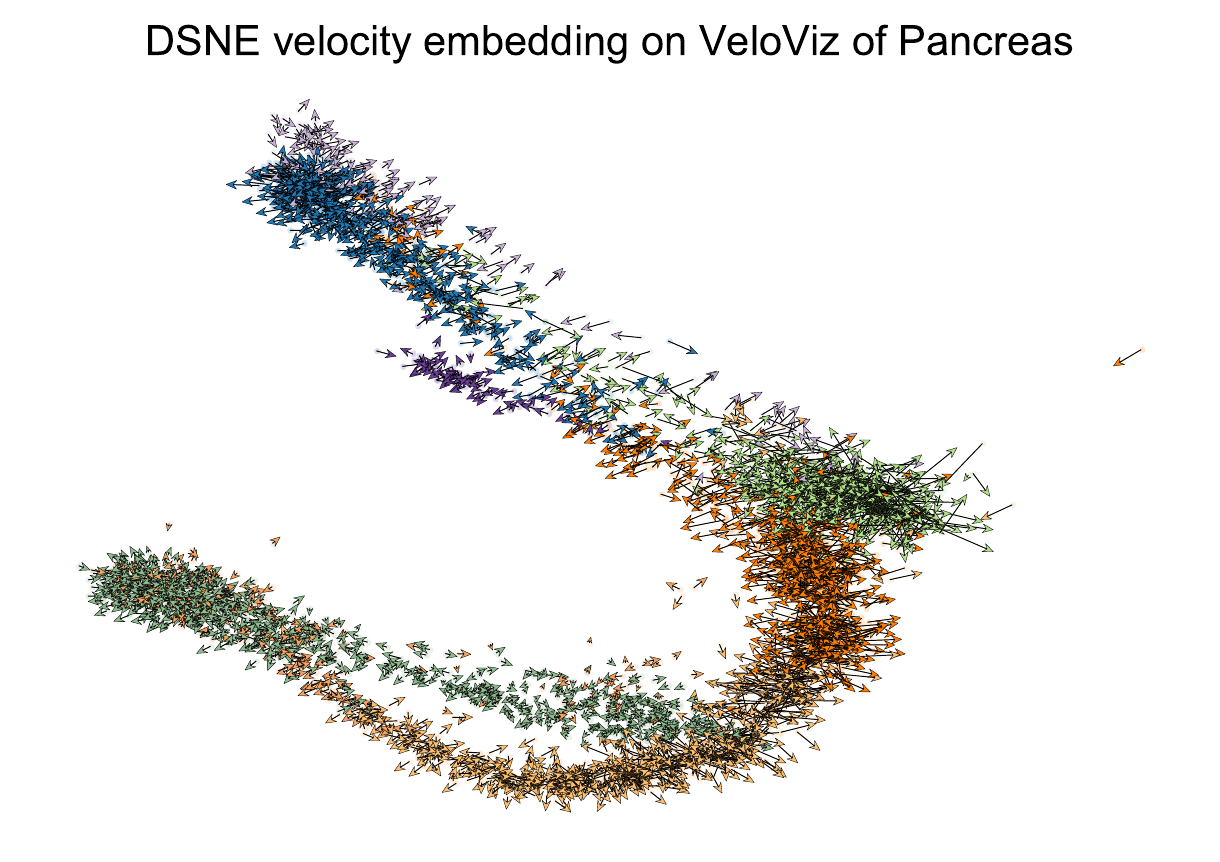}
\includegraphics[width=150mm]{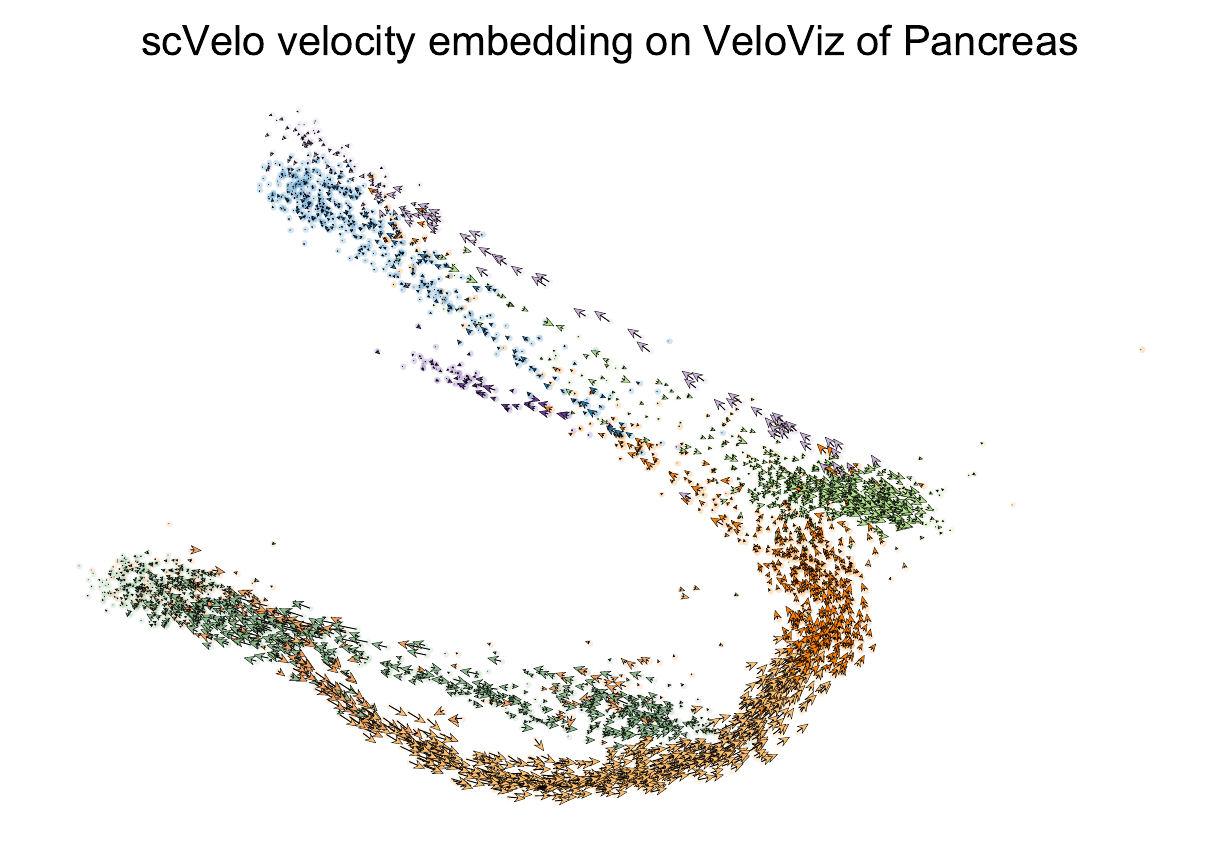}
  \caption{The arrow plot of the velocity embeddings on the VeloViz map points of the pancreas data. The top figure shows the results of  DSNE  and  the bottom figure shows the results of scVeloEmbedding. Zoom in for details. }
  \label{fig:pancreas-veloviz-arrow}
  \end{figure}

We plot the stream, grid, arrow plot in Fig~\ref{fig:pancreas-umap-stream},  Fig~\ref{fig:pancreas-umap-grid},  Fig~\ref{fig:pancreas-umap-arrow} for UMAP map points, respectively;  and Fig~\ref{fig:pancreas-veloviz-stream},  Fig~\ref{fig:pancreas-veloviz-grid},  Fig~\ref{fig:pancreas-veloviz-arrow} for VeloViz map points, respectively.

On the UMAP stream plot (Fig~\ref{fig:pancreas-umap-stream}) and grid plot ( Fig~\ref{fig:pancreas-umap-grid}), we found that scVeloEmbedding seems over smooth the velocity direction to the mean direction, while DSNE reveal more details for the local moving trend on the map, which may helpful to identify some special cells in the data. 
For the VeloViz Plots, it occurs the similar phenomenon. Note that the VeloViz organize cell clusters on the map were different from the UMAP, and which is better representation need to be checked by the biologists. 

\section{Discussion}
Currently, we leaning the embedding of the velocity with known low-dimensional embedding of map points, it is more reasonable to learn the map points and the velocity embedding of the high dimensional data points and its velocities simultaneously, this need to more dedicate design of methods, since it is  hard to adjust the map points and its velocity in the low dimension space to reduce the cost stably, this opens new research opportunity. \cite{VeloViz}  recently proposed VeloViz method  is the effort to that direction, which  gets the low dimensional embeddings with the velocity informations comes from the probability distribution which transformed from the distance of points $x_i$ with  velocity $v_i$  and $x_j$. It is helpful to organize the low dimensional points which contains the velocity information.

To recovery the velocity embedding on the low dimension map points,  it must preserve the local direction information in the low dimensional space, e.g. $ || \hat v_i - \Delta \hat x_{ij}||^2 =  \alpha || \hat w_i -  \Delta \hat y_{ij}||^2, \; j \in \text{i's neighbors}$ for some positive scalar $\alpha$. This is not specially emphasized in the dimension reduction techniques, e.g., t-SNE, UMAP, which left to the future work.  

\section{Conclusion}
In this paper, we propose DSNE to get the low dimensional velocity embeddings when given  the high dimensional data points with its velocities  and the low dimensional map points. The numerical experiments show that DSNE can faithfully  keep the direction of the velocity in the low dimensional space correspond to the velocity direction in the high dimensional space. It is helpful to visualize the cell trajectories in the biological science, which may aid to check how the cells move around its near neighbors, and the global structures may give us the sense the development relations of different cell subtypes. We hope that this method can help to recovery mystery of  the cell differentiation and embryo development. And we also expect  that you can find more usages of this method. 

\section*{Acknowledgements}
Thank to my family ( especially for my mother, Qixia Chen and father, Wenjiang Shi )  for they provides me a suitable environment for this work. Thank to all the teachers who taught me to guide  me to the road of truth.

\section*{Appendix A. Code availability }
DSNE are available as python package on~\url{https://github.com/songtingstone/dsne}. Scripts to reproduce results of the primary analyses will be made available on~ \url{https://github.com/songtingstone/dsne2021}.
The code is learned and adapted the C implementation~\url{https://github.com/danielfrg/tsne}  of  BH-SNE (\cite{BH-SNE}), special thanks to Laurens van der Maaten and Daniel Rodriguez. 

\section*{Appendix B. Derivation of the DSNE gradient and Hessian matrix. } 
DSNE use the scaled  KL divergence as the loss function 
\begin{equation}
\begin{array}{l}
C= \sum_i \sum_{j \in \text{i's neighbors}} \tilde p_{j|i} \log \frac {p_{j|i}} { q_{j|i}}
\end{array}
\end{equation}
where 
\begin{equation}
\begin{array}{l}
p_{j|i} = \frac 1 {Z_{x,i} } \exp( - 2 \beta_{x,i}( 1 - \cos_{x,ij})), \; j \in \text{  i's neighbors } \\
p_{i|i}  =  \frac 1 {Z_{x,i} } \\ 
Z_{x,i}  = 1 +  \sum_{j \in \text{  i's neighbors } } \exp( - 2 \beta_{x,i}( 1 - \cos_{x,ij})) \\
\cos_{x,ij} = \langle \hat v_i, \Delta \hat x_{ij} \rangle\\
\tilde p_{j|i} = \frac 1 {\tilde Z_{x,i}  } \exp( - 2 \beta_{x,i}( 1 - \cos_{x,ij})),  \; j \in \text{  i's neighbors } \\
\tilde p_{i|i} = 0 \\
\tilde Z_{x,i}  =  \sum_{j \in \text{  i's neighbors } } \exp( - 2 \beta_{x,i}( 1 - \cos_{x,ij})) \\
q_{j|i} = \frac 1 {Z_{y, i} } \exp( - 2 \beta_{y,i}( 1 - \cos_{y,ij})) \; j \in \text{  i's neighbors } \\
q_{i|i}  =  \frac 1 {Z_{y,i} }  \\
Z_{y,i}  = 1 +  \sum_{j \in \text{  i's neighbors } } \exp( - 2 \beta_{y,i}( 1 - \cos_{y,ij})) \\
\cos_{y,ij} = \langle \hat w_i, \Delta \hat y_{ij} \rangle\\ 
\hat w_i = \frac{w}{||w||}\\
\end{array}
\end{equation}

Note that $w_i, \beta_{y,i}$ are independent of each other, so that we can separate the loss into part of $i$, 
\begin{equation}
\begin{array}{l} 
C = \sum_i C_i\\
C_i = \sum_{j \in \text{i's neighbors}} \tilde p_{j|i} \log \frac {p_{j|i}} { q_{j|i}}
\end{array}
\end{equation}

To simplify the notation, we define $\hat q_{j|i} :=  \exp( - 2 \beta_{y,i}( 1 - \cos_{y,ij}))$, so that $Z_{y,i} = 1 + \sum_{j \in \text{  i's neighbors } } \hat q_{j|i} $  and  $ q_{j|i}  = \frac{ \hat q_{j|i} } {Z_{y,i}}$.

The gradient of the cost function $C$ with respect to $w_i$  is given by
\begin{equation}
\label{eq:gradient-C-key}
\begin{array}{ll} 
\frac {\partial C} {\partial w_i}  &= \frac {\partial C_i} {\partial w_i}  \\
	&= \frac {\partial} {\partial w_i} [ \sum_{j \in \text{i's neighbors}} \tilde p_{j|i} \log \frac {p_{j|i}} { q_{j|i}} ]  \\
	&= \frac {\partial} {\partial w_i} [ \sum_{j \in \text{i's neighbors}} - \tilde p_{j|i} \log q_{j|i} ]  \\
	&= -\sum_{j \in \text{i's neighbors}} -\tilde p_{j|i} \frac {\partial} {\partial w_i} [\log q_{j|i} ] \\ 
	&= -\sum_{j \in \text{i's neighbors}} \tilde p_{j|i}   \frac {\partial} {\partial w_i} [\log \hat q_{j|i} - \log Z_{y,i} ] \\ 
	&= -\sum_{j \in \text{i's neighbors}} \tilde p_{j|i}   \frac {\partial} {\partial w_i} [\log \hat q_{j|i}]  +  \frac {\partial} {\partial w_i} \log Z_{y,i}  \\ 
	&= -\sum_{j \in \text{i's neighbors}} \tilde p_{j|i}   \frac {\partial} {\partial w_i} [\log \hat q_{j|i}]  +   \frac 1 { Z_{y,i}} \frac {\partial} {\partial w_i}  Z_{y,i}  \\ 
	&= -\sum_{j \in \text{i's neighbors}} \tilde p_{j|i}   \frac {\partial} {\partial w_i} [\log \hat q_{j|i}]  +   \frac 1 { Z_{y,i}} \frac {\partial} {\partial w_i}  [1 + \sum_{j \in \text{  i's neighbors } } \hat q_{j|i}   ] \\ 
	&= -\sum_{j \in \text{i's neighbors}} \tilde p_{j|i}   \frac {\partial} {\partial w_i} [\log \hat q_{j|i}]  +   \frac 1 { Z_{y,i}}   \sum_{j \in \text{  i's neighbors } } \hat q_{j|i}  \frac{\partial} {\partial w_i}  [ \log \hat q_{j|i}  ] \\ 
	&=  -\sum_{j \in \text{i's neighbors}} \tilde p_{j|i}   \frac {\partial} {\partial w_i} [\log \hat q_{j|i}]  +    \sum_{j \in \text{  i's neighbors } }  q_{j|i}  \frac{\partial} {\partial w_i}  [ \log \hat q_{j|i}  ] \\ 
	&= -\sum_{j \in \text{i's neighbors}} ( \tilde p_{j|i}  - q_{j|i}) \frac{\partial} {\partial w_i}  [ \log \hat q_{j|i}  ]  
\end{array}
\end{equation}
where the sixth equality comes from the fact $\sum_{j \in \text{i's neighbors}} \tilde p_{j|i}  = 1$. 
Note that 
\begin{equation}
\begin{array}{ll} 
 \frac{\partial} {\partial w_i}  [ \log \hat q_{j|i}  ]   &=  \frac{\partial} {\partial w_i} [- 2 \beta_{y,i}( 1 - \cos_{y,ij}) ]  \\
 	&=  2 \beta_{y,i} \frac{\partial} {\partial w_i} [  \cos_{y,ij} ]  \\
	&=  2 \beta_{y,i} \frac{\partial} {\partial w_i} \langle  \frac {w_i} {||w_i||}, \Delta \hat y_{ij} \rangle  \\
	&=  2 \beta_{y,i}[ \frac { \Delta \hat y_{ij} }{||w_i||} -   \langle  w_i,  \Delta \hat y_{ij} \rangle \frac {w_i} {||w_i||^3}  ]  \\
	&=  \frac {2 \beta_{y,i} } {||w_i||} ( \Delta \hat y_{ij}  - cos_{y, ij} \hat w_i)
\end{array}
\end{equation}
The gradient of the cost function $C$ with respect to $w_i$  is given by
\begin{equation}
\begin{array}{ll} 
\frac {\partial C} {\partial w_i}  &= -\sum_{j \in \text{i's neighbors}} ( \tilde p_{j|i}  - q_{j|i}) \frac{\partial} {\partial w_i}  [ \log \hat q_{j|i}  ]   \\ 
	&=  -\sum_{j \in \text{i's neighbors}} ( \tilde p_{j|i}  - q_{j|i})  \frac {2 \beta_{y,i} } {||w_i||} ( \Delta \hat y_{ij}  - cos_{y, ij} \hat w_i) \\
	&=  \sum_{j \in \text{i's neighbors}} ( \tilde p_{j|i}  - q_{j|i})  \frac {2 \beta_{y,i} } {||w_i||} (- \Delta \hat y_{ij}  + cos_{y, ij} \hat w_i)
\end{array}
\end{equation}

Similarly, the gradient of the cost function $C$ with respect to $\beta_{y,i}$  is given by
\begin{equation}
\begin{array}{ll} 
\frac {\partial C} {\partial \beta_{y,i} }  &= -\sum_{j \in \text{i's neighbors}} ( \tilde p_{j|i}  - q_{j|i}) \frac{\partial} {\partial \beta_{y,i}}  [ \log \hat q_{j|i}  ]   \\ 
	&=  -\sum_{j \in \text{i's neighbors}} ( \tilde p_{j|i}  - q_{j|i})  [- 2 ( 1 - \cos_{y,ij}) ]\\
	&=  \sum_{j \in \text{i's neighbors}} ( \tilde p_{j|i}  - q_{j|i})  2 ( 1 - \cos_{y,ij}) 
\end{array}
\end{equation}

Next we derive the second order gradient of the DSNE cost function. It basically use the same trick as in equation(\ref{eq:gradient-C-key}), but with a little more complex calculations. 

For the second order gradient of the cost function with respect to $w_i$, we calculate $\frac{\partial } {\partial w_i} [\frac{\partial C } {\partial w_i} (1) ] $ first, where we use  $x(1)$ to denote the first element of vector $x$. Note that $\frac{\partial C } {\partial w_i} (1) =  \sum_{j \in \text{i's neighbors}} ( \tilde p_{j|i}  - q_{j|i})  \frac {2 \beta_{y,i} } {||w_i||} (- \Delta \hat y_{ij}(1)   + cos_{y, ij} \hat w_i(1)) $. So we have 
 \begin{equation}
 \label{eq:G-w-1}
\begin{array}{ll} 
\frac{\partial } {\partial w_i} [\frac{\partial C } {\partial w_i} (1) ]    &= \frac{\partial } {\partial w_i} [ \sum_{j \in \text{i's neighbors}} ( \tilde p_{j|i}  - q_{j|i})  \frac {2 \beta_{y,i} } {||w_i||} (- \Delta \hat y_{ij}(1)   + cos_{y, ij} \hat w_i(1)) ] \\
	&=  \frac{\partial } {\partial w_i} [ \sum_{j \in \text{i's neighbors}} ( \tilde p_{j|i}  - q_{j|i})  \frac {2 \beta_{y,i} } {||w_i||} (- \Delta \hat y_{ij}(1)   + cos_{y, ij} \hat w_i(1)) ] \\
	&=   \frac{\partial } {\partial w_i} [ \sum_{j \in \text{i's neighbors}} ( \tilde p_{j|i}  - q_{j|i})  2 \beta_{y,i}  (- \frac { \Delta \hat y_{ij}(1) } {||w_i||}   +  \langle  w_i, \Delta \hat y_{ij} \rangle   \frac {w_i(1)} {||w_i||^3} ) ] \\
	&= \sum_{j \in \text{i's neighbors}} ( \tilde p_{j|i}  - q_{j|i})  2 \beta_{y,i}  \frac{\partial } {\partial w_i} [- \frac { \Delta \hat y_{ij}(1) } {||w_i||}   +  \langle  w_i, \Delta \hat y_{ij} \rangle   \frac {w_i(1)} {||w_i||^3} ] \\
	&\quad -  \sum_{j \in \text{i's neighbors}}  2 \beta_{y,i}  (- \frac { \Delta \hat y_{ij}(1) } {||w_i||}   +  \langle  w_i, \Delta \hat y_{ij} \rangle   \frac {w_i(1)} {||w_i||^3} )   \frac{\partial } {\partial w_i} [q_{j|i}]\\
\end{array}
\end{equation}

 \begin{equation}
 \label{eq:G-w-2}
\begin{array}{ll} 
&\quad \frac{\partial } {\partial w_i} [- \frac { \Delta \hat y_{ij}(1) } {||w_i||}   +  \langle  w_i, \Delta \hat y_{ij} \rangle   \frac {w_i(1)} {||w_i||^3} ] \\
&= \Delta \hat y_{ij}(1) \frac { w_i } {||w_i||^3} + \Delta \hat y_{ij}   \frac {w_i(1)} {||w_i||^3}  +  \langle  w_i, \Delta \hat y_{ij} \rangle  \frac {1} {||w_i||^3}  e_1 - 3  \langle  w_i, \Delta \hat y_{ij} \rangle w_i(1) \frac {w_i} {||w_i||^5}     \\
&= \frac{1}{||w_i||^2} ( \Delta \hat y_{ij}(1) \hat w_i + \Delta \hat y_{ij} \hat w_i(1) + cos_{y,ij} e_1 -3 cos_{y,ij} \hat w_i(1) \hat w_i  )
\end{array}
\end{equation}
where $e_1 = [1, 0, \ldots, 0] \in \mathbb R^d$. 
\begin{equation}
\label{eq:G-w-3}
\begin{array}{ll} 
 \frac{\partial } {\partial w_i} q_{j|i}  &=  \frac{\partial } {\partial w_i}  [\frac {\hat  q_{j|i} } {Z_{y,i}} ]\\
	&=Z_{y,i}^{-1} \frac{\partial } {\partial w_i}  [\hat  q_{j|i} ] - \hat  q_{j|i} Z_{y,i}^{-2}  \frac{\partial } {\partial w_i} Z_{y,i}  \\
	&= q_{j|i} \frac{\partial } {\partial w_i}  [\log \hat  q_{j|i} ] - q_{j|i} \sum_{k   \in \text{i's neighbors}} q_{k|i}   \frac{\partial } {\partial w_i}  [\log \hat  q_{k|i} ]  \\
	&=  q_{j|i}   \frac {2 \beta_{y,i} } {||w_i||} ( \Delta \hat y_{ij}  - cos_{y, ij} \hat w_i)  - q_{j|i} \sum_{k   \in \text{i's neighbors}} q_{k|i}   \frac {2 \beta_{y,i} } {||w_i||} ( \Delta \hat y_{ik}  - cos_{y, ik} \hat w_i) \\
	&=  q_{j|i}   \frac {2 \beta_{y,i} } {||w_i||}  [ (  \Delta \hat y_{ij}  - \sum_{k   \in \text{i's neighbors}}    \Delta \hat y_{ik}  )  -  \hat w_i ( cos_{y, ij} - \sum_{k   \in \text{i's neighbors}} q_{k|i}  cos_{y, ik}) ] 
\end{array}
\end{equation}
Combine equation (\ref{eq:G-w-1}, \ref{eq:G-w-2}, \ref{eq:G-w-3}) into one, we get 

 \begin{equation}
 \label{eq:G-w-first-element}
\begin{array}{ll} 
\frac{\partial } {\partial w_i} [\frac{\partial C } {\partial w_i} (1) ]    &= \sum_{j \in \text{i's neighbors}} ( \tilde p_{j|i}  - q_{j|i})  2 \beta_{y,i} \\
	&\quad *\frac{1}{||w_i||^2} ( \Delta \hat y_{ij}(1) \hat w_i + \Delta \hat y_{ij} \hat w_i(1) + cos_{y,ij} e_1 -3 cos_{y,ij} \hat w_i(1) \hat w_i  ) \\
	&\quad -  \sum_{j \in \text{i's neighbors}} 2 \beta_{y,i}   (- \frac { \Delta \hat y_{ij}(1) } {||w_i||}   +  \langle  w_i, \Delta \hat y_{ij} \rangle   \frac {w_i(1)} {||w_i||^3} ) \\
	&\quad * q_{j|i}   \frac {2 \beta_{y,i} } {||w_i||}  [ (  \Delta \hat y_{ij}  - \sum_{k   \in \text{i's neighbors}}    \Delta \hat y_{ik}  )  -  \hat w_i ( cos_{y, ij} - \sum_{k   \in \text{i's neighbors}} q_{k|i}  cos_{y, ik}) ] 
\\
\end{array}
\end{equation}
From the above equation, it's easy to see that the second order gradient of cost $C$ with respect to $w_i$ is given by
 \begin{equation}
 \label{eq:G-w}
\begin{array}{ll} 
\frac{\partial^2 C } {\partial w_i\partial w_i^T}   &= \frac { 2 \beta_{y,i} } { ||w_i||^2 } \sum_{j \in \text{i's neighbors}}  (\tilde p_{j|i} - q_{j|i})  [ \Delta \hat y_{ij}  \hat w_i^T + \hat w_i \Delta  \hat y_{ij}^T + cos_{y,ij} I - 3 cos_{y,ij} \hat w_i \hat w_i^T ] \\
	& \quad - \frac{ 4\beta_{y,i}^2} {||w||^2} \sum_{j \in \text{i's neighbors}}  q_{j|i}  [ -\Delta  \hat y_{ij} + cos_{y, ij} \hat w_i] [ ( \Delta  \hat y_{ij} - \mathbb E \Delta  \hat y_{i}) - \hat w_i ( cos_{y, ij} - \mathbb E cos_{y, i} ) ]^T\\
\end{array}
\end{equation}
where 
$\mathbb E \Delta  \hat y_{i} =  \sum_{k   \in \text{i's neighbors}}  q_{k|i}    \Delta \hat y_{ik}$,  $ \mathbb E cos_{y, i} =  \sum_{k   \in \text{i's neighbors}}   q_{k|i} \Delta \hat y_{ik}$ and $I \in \mathbb R^{d \times d} $ is the identity matrix. 

Similarly, the second order gradient of cost $C$ with respect to $\beta_{y,i}$ is given by 

 \begin{equation}
 \label{eq:G-beta-1}
\begin{array}{ll} 
\frac{\partial^2 C } {\partial^2  \beta_{y,i}}   &= \frac{\partial } {\partial \beta_{y,i}} [ \sum_{j \in \text{i's neighbors}} ( \tilde p_{j|i}  - q_{j|i})  2 ( 1 - \cos_{y,ij}) 
]  \\
	&=  -   \sum_{j \in \text{i's neighbors}}    2 ( 1 - \cos_{y,ij})  \frac{\partial } {\partial \beta_{y,i}}  q_{j|i} 
\end{array}
\end{equation}

 \begin{equation}
 \label{eq:G-beta-2}
\begin{array}{ll} 
\frac{\partial } {\partial \beta_{y,i} } q_{j|i}  &=  \frac{\partial } {\partial  \beta_{y,i}}  [\frac {\hat  q_{j|i} } {Z_{y,i}} ]\\
	&=Z_{y,i}^{-1} \frac{\partial } {\partial  \beta_{y,i} }  [\hat  q_{j|i} ] - \hat  q_{j|i} Z_{y,i}^{-2}  \frac{\partial } {\partial  \beta_{y,i}} Z_{y,i}  \\
	&= q_{j|i} \frac{\partial } {\partial  \beta_{y,i} }  [\log \hat  q_{j|i} ] - q_{j|i} \sum_{k   \in \text{i's neighbors}} q_{k|i}   \frac{\partial } {\partial  \beta_{y,i} }  [\log \hat  q_{k|i} ]  \\
	&=  q_{j|i}  [ -2 ( 1 - \cos_{y,ij}) ] - q_{j|i} \sum_{k   \in \text{i's neighbors}} q_{k|i}  [ -2 ( 1 - \cos_{y,ik}) ]  \\
	&=  -q_{j|i}    [ 2 ( 1 - \cos_{y,ij}) - \sum_{k   \in \text{i's neighbors}} q_{k|i} 2 ( 1 - \cos_{y,ik})] 
\end{array}
\end{equation}
Combine equation (\ref{eq:G-beta-1}, \ref{eq:G-beta-2}) into one, we get the second order gradient of cost $C$ with respect to $\beta_{y,i}$, 
 \begin{equation}
 \label{eq:G-beta}
\begin{array}{ll} 
\frac{\partial^2 C } {\partial^2  \beta_{y,i}}   &=  \sum_{j \in \text{i's neighbors}}    2 ( 1 - \cos_{y,ij})  q_{j|i}    [ 2 ( 1 - \cos_{y,ij}) - \sum_{k   \in \text{i's neighbors}} q_{k|i} 2 ( 1 - \cos_{y,ik})]  \\
	&= 4  [ \sum_{j \in \text{i's neighbors}}   q_{j|i} ( 1 - \cos_{y,ij})^2  - ( \sum_{j \in \text{i's neighbors}}   q_{j|i}  ( 1 - \cos_{y,ij}) )^2   ] 
\end{array}
\end{equation}

\bibliographystyle{apalike}  
\bibliography{dsne}  

\begin{thebibliography}{}

\bibitem[Atta and Fan, 2021]{VeloViz}
Atta, L. and Fan, J. (2021).
\newblock VeloViz: RNA-velocity informed 2d embeddings for visualizing cellular
  trajectories.
\newblock {\em bioRxiv}.

\bibitem[Bergen et~al., 2020]{dynamical-RNA-velocity}
Bergen, V., Lange, M., Peidli, S., Wolf, F.~A., and Theis, F.~J. (2020).
\newblock Generalizing RNA velocity to transient cell states through dynamical
  modeling.
\newblock {\em Nature Biotechnology}, 38(12):1408--1414.

\bibitem[Hinton and Roweis, 2003]{SNE}
Hinton, G. and Roweis, S. (2003).
\newblock Stochastic neighbor embedding.
\newblock {\em Advances in Neural Information Processing Systems},
  15(4):833--840.

\bibitem[Jacobs, 1988]{Jacobs1988}
Jacobs, R.~A. (1988).
\newblock Increased rates of convergence through learning rate adaptation.
\newblock {\em Neural Networks}, 1(4):295--307.

\bibitem[La~Manno et~al., 2018]{RNA-velocity}
La~Manno, G., Soldatov, R., Zeisel, A., Braun, E., Hochgerner, H., Petukhov,
  V., Lidschreiber, K., Kastriti, M.~E., L{\"o}nnerberg, P., Furlan, A., Fan,
  J., Borm, L.~E., Liu, Z., van Bruggen, D., Guo, J., He, X., Barker, R.,
  Sundstr{\"o}m, E., Castelo-Branco, G., Cramer, P., Adameyko, I., Linnarsson,
  S., and Kharchenko, P.~V. (2018).
\newblock RNA velocity of single cells.
\newblock {\em Nature}, 560(7719):494--498.

\bibitem[Laurens et~al., 2008]{t-SNE}
Laurens, Maaten, V.~D., and Hinton, G. (2008).
\newblock Visualizing data using t-SNE.
\newblock {\em Journal of Machine Learning Research}, 9(2605):2579--2605.

\bibitem[Mcinnes and Healy, 2018]{UMAP}
Mcinnes, L. and Healy, J. (2018).
\newblock UMAP: Uniform manifold approximation and projection for dimension
  reduction.
\newblock {\em Journal of Open Source Software}, 3(29):861.

\bibitem[van~der Maaten, 2013]{BH-SNE}
van~der Maaten, L. (2013).
\newblock Barnes-Hut-SNE.

\bibitem[Yianilos, 1993]{vptree}
Yianilos, P.~N. (1993).
\newblock Data structures and algorithms for nearest neighbor search in general
  metric spaces.
\newblock pages 311--321.

\end{thebibliography}


\end{document}